%% file: main.tex
\documentclass[11pt, a4paper, goog]{google}

\usepackage[authoryear, sort&compress, round]{natbib}
\usepackage{multirow}
\usepackage{graphicx}
\usepackage[export]{adjustbox}
\usepackage{hyperref}
\usepackage{xcolor}
\usepackage{minted}
\usepackage{listings}

\lstdefinestyle{pseudopython}{
language=Python,
basicstyle=\ttfamily\footnotesize,
keywordstyle=\color{blue!65!black}\bfseries,
commentstyle=\color{gray}\itshape,
stringstyle=\color{purple!65!black},
backgroundcolor=\color{gray!8},
xleftmargin=6pt, xrightmargin=6pt,
framexleftmargin=6pt, framexrightmargin=6pt,
aboveskip=6pt, belowskip=0pt,
showstringspaces=false,
keepspaces=true,
columns=fullflexible,
breaklines=true,
tabsize=4,
}

\uselogo{}

\newcommand{\sname}{ScientistTwo}
\newcommand{\stdv}[2][\tiny]{#1\text{$\pm$}#2}
\definecolor{Gray}{gray}{0.9}

\title{\sname: Pioneering the Human Knowledge Frontier with Autonomous AI}

\correspondingauthor{jaehyunnam@google.com, jinsungyoon@google.com}

\reportnumber{} 

\author[1]{Jaehyun Nam}
\author[1]{Jinsung Yoon}
\author[1]{Yanzhou Pan}
\author[2]{Yubo Wang}
\author[1]{Rui Meng}
\author[1]{Parthasarathy Ranganathan}
\author[1]{Tomas Pfister}

\affil[1]{Google Cloud AI Research}
\affil[2]{University of Waterloo}

\input{sections/0_abstract}

\begin{document}

\maketitle

\input{sections/1_introduction}
\input{sections/2_related_works}
\input{sections/3_new_method}
\input{sections/4_experiment}
\input{sections/5_discussion}
\input{sections/6_conclusion}

\newpage
\bibliography{main}

\newpage
\appendix
\input{sections/appendix}

\end{document}

%% file: sections/0_abstract.tex
\begin{abstract}
Scientific discovery is defined by the ability to identify the boundaries of existing knowledge and venture into unexplored territory. The ultimate vision for AI in science is problem-driven autonomous research: given a fundamental challenge by a human expert, the AI independently navigates the scientific landscape, uncovers theoretical and empirical bottlenecks, and systematically expands the frontier of knowledge. In this paper, we introduce \emph{\sname}, a fully autonomous multi-agent framework designed to realize this vision. Specifically, \sname~takes an initial problem as input, establishes state-of-the-art baselines, formulates novel hypotheses, and coordinates specialized agents to orchestrate an end-to-end discovery cycle without human intervention. Moreover, the framework rigorously conducts experiments using diverse datasets and metrics, refines methodologies through automated ablation studies, and validates research findings via a closed-loop simulated peer-review rebuttal engine. To evaluate \sname's capabilities against the highest standards of human scientific achievement, we benchmark it across papers accepted at top-tier conferences such as ICLR, ICML, and NeurIPS. As a result, \sname~autonomously generates expert-level, publishable papers and fully verified, executable codebases.
Its solutions consistently outperform human state-of-the-art models, and achieve higher average review ratings than human-authored papers under automated AI review agents. These results show that \sname~is not merely an assistive tool but an autonomous scientific pioneer capable of pushing the frontiers of human discovery. Project website: \url{https://scientist-two.github.io/}
\end{abstract}

%% file: sections/1_introduction.tex
\input{figures/problem_setup}
\section{Introduction}

\input{figures/qualitative_result}

Scientific discovery has long been the hallmark of human ingenuity, defined by the ability to identify the boundaries of current knowledge and venture into the unknown. With the rapid advancement of foundation models \citep{team2023gemini, liu2024deepseek, singh2025openai}, artificial intelligence (AI) is transitioning from a passive conversational assistant to an active participant in the scientific process \citep{xu2025comprehensive, meng2026scientistone,jansen2025codescientist}. The ultimate ambition of this paradigm is purely \emph{problem-driven autonomous discovery}: a human researcher simply specifies a scientific challenge, and an autonomous AI independently navigates the landscape of human knowledge, diagnoses theoretical and empirical bottlenecks, formulates novel hypotheses, and executes the end-to-end research lifecycle on its own to pioneer the human knowledge frontier.

Despite recent progress in the field of autonomous research agents \citep{tang2026ai,weng2025deepscientist,yamada2025ai}, a substantial gap remains between automated assistant systems and rigorous empirical scientific standards. Existing systems primarily focus on optimizing single-scalar metrics on isolated benchmarks, lacking the multi-dimensional reasoning capabilities required to tackle complex scientific problems \citep{chen2026mars,jin2026toward}. More importantly, current agents lack the closed-loop empirical rigor of human scientists who continuously iterate based on evidence. They cannot systematically conduct ablation studies to isolate causal mechanisms for improvement, nor can they engage in dynamic peer-review processes essential for validating ideas and addressing methodological critiques through targeted supplementary experiments.

To overcome these fundamental challenges, we introduce \emph{\sname}, an expert-level autonomous multi-agent framework designed to pioneer the frontier of human knowledge (see Figure~\ref{fig:problem_setup}). Specifically, \sname~operates on a foundational division of labor: the human researcher defines the target scientific challenge, while the AI autonomously orchestrates the path to advance it. To rigorously demonstrate this capability against the highest standard of scientific rigor, we subject \sname~to the real-world testing ground: given competitive peer-reviewed human research from top-tier AI venues (e.g., ICLR, ICML, NeurIPS), \sname~must independently discover meaningful advancements beyond the established state-of-the-art, while simultaneously demonstrating that its autonomous scientific discovery process is measurable, reproducible, and transparent.

\input{figures/overview}
To achieve expert-level rigor across diverse disciplines demanded by top-tier science, \sname~coordinates a collaborative ecosystem of specialized agents (see Figure~\ref{fig:overview}):
\begin{itemize}
    \item \emph{Holistic Benchmark Reasoning \& Efficient Screening}: Rather than overfitting to a single metric, \sname~evaluates proposed ideas across comprehensive, multi-dataset benchmarks. To balance computation efficiency, it employs a subset-first evaluation strategy, rapidly filtering ideas on representative benchmark slices before allocating compute to full-scale experiments.
    \item \emph{Ablation-Driven Hypothesis Refinement}: Emulating the empirical rigor of expert researchers, \sname~autonomously designs and executes ablation studies to isolate individual component contributions, dynamically pruning ineffective components and refining its core hypothesis.
    \item \emph{Dynamic Peer-Review \& Rebuttal Loops}: To ensure publication-grade validity, generated drafts are critiqued by a simulated Peer-Review Agent. Rather than treating review comments as passive editorial guidance, a dedicated Rebuttal Agent actively conceives, codes, and executes targeted supplementary experiments to address critique. A Meta-Review Agent oversees this cycle, triggering recursive refinement loops until rigorous acceptance criteria are satisfied.
\end{itemize}

We evaluate \sname~across 107 diverse research challenges encompassing diverse areas such as optimization, reinforcement learning, large language models (LLMs), and theory (see Figure~\ref{fig:problem_setup}). Specifically, \sname~successfully advances 80.4\% of the target problems, delivering an average relative performance gain of 25.2\% over the original human state-of-the-art baselines. Furthermore, the autonomously generated manuscripts and executable codebases consistently achieve high acceptance rates across independent automated peer-review evaluations with high scores. These results demonstrate that \sname~moves beyond passive empirical problem-solving, serving as an autonomous pioneer capable of systematically expanding the frontiers of scientific knowledge.

%% file: figures/problem_setup.tex
\begin{figure*}[t]
\centering
\includegraphics[width=\linewidth]{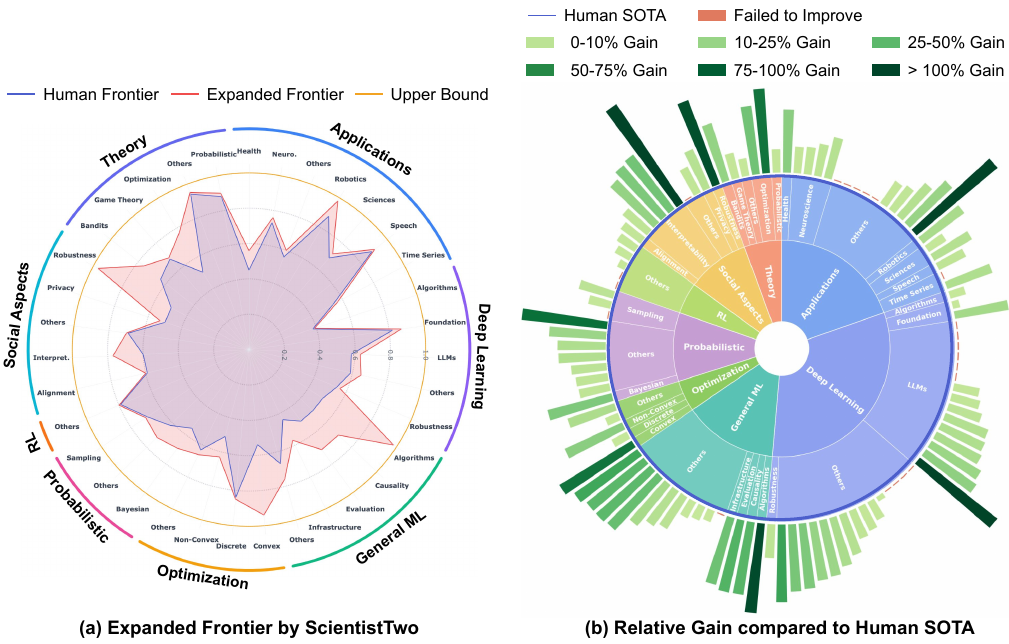}
\caption{\textbf{Teaser.} \sname~pushes the frontier of human knowledge across diverse research domains (e.g., LLMs, robotics, neuroscience, speech, robustness, reinforcement learning, game theory, privacy, optimization, and time series) by generating publication-quality papers and fully verified codebases, with the resulting methodologies consistently outperforming human state-of-the-art baselines. Specifically, \sname~improves 86 out of 107 papers (an 80.4\% success rate) with an average relative improvement of 25.2\% over human state-of-the-art methods. Furthermore, papers generated by \sname~surpass the average scores of accepted papers at ICLR 2026 and NeurIPS 2025 under the Stanford Agentic Reviewer, demonstrating its capability to produce manuscripts that reach the empirical acceptance standards of top-tier AI venues.}
\label{fig:problem_setup}
\end{figure*}

%% file: figures/qualitative_result.tex
\begin{figure}[t]
  \centering
  \setlength{\tabcolsep}{0pt} 
  \renewcommand{\arraystretch}{0}
  \resizebox{\textwidth}{!}{
  \begin{tabular}{|c|c|c|c|c|}
  \hline
    \includegraphics[width=0.2\textwidth]{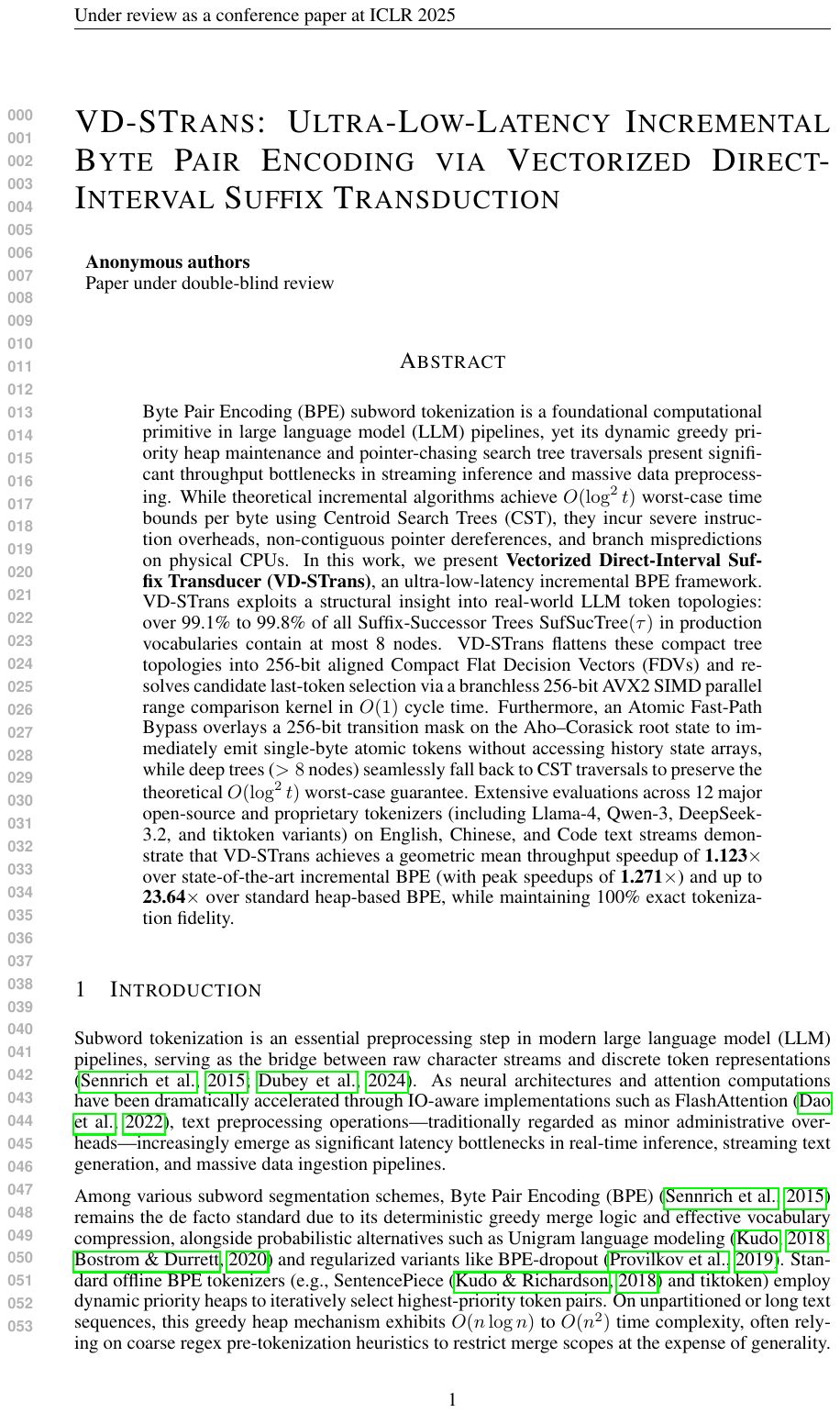} &
    \includegraphics[width=0.2\textwidth]{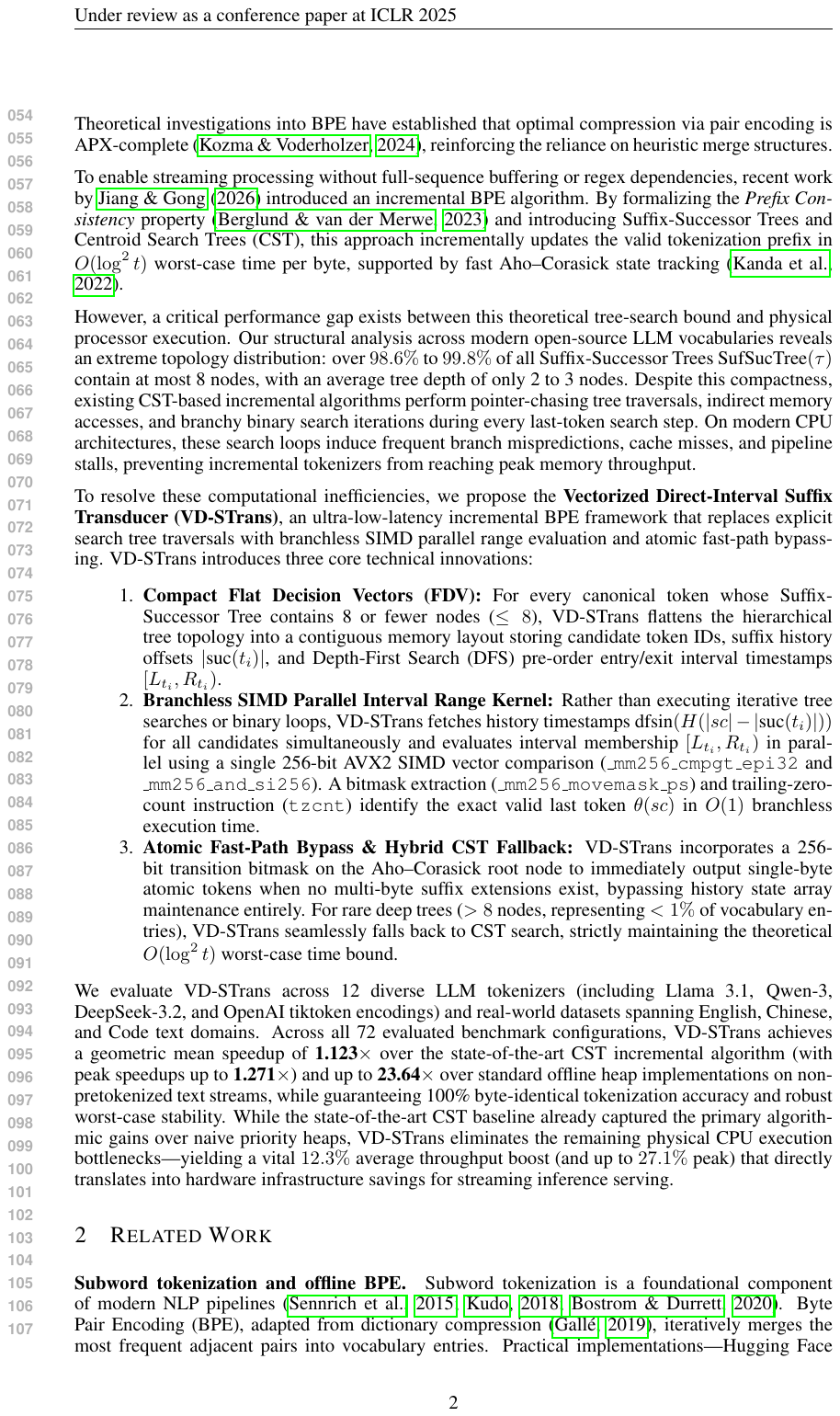} &
    \includegraphics[width=0.2\textwidth]{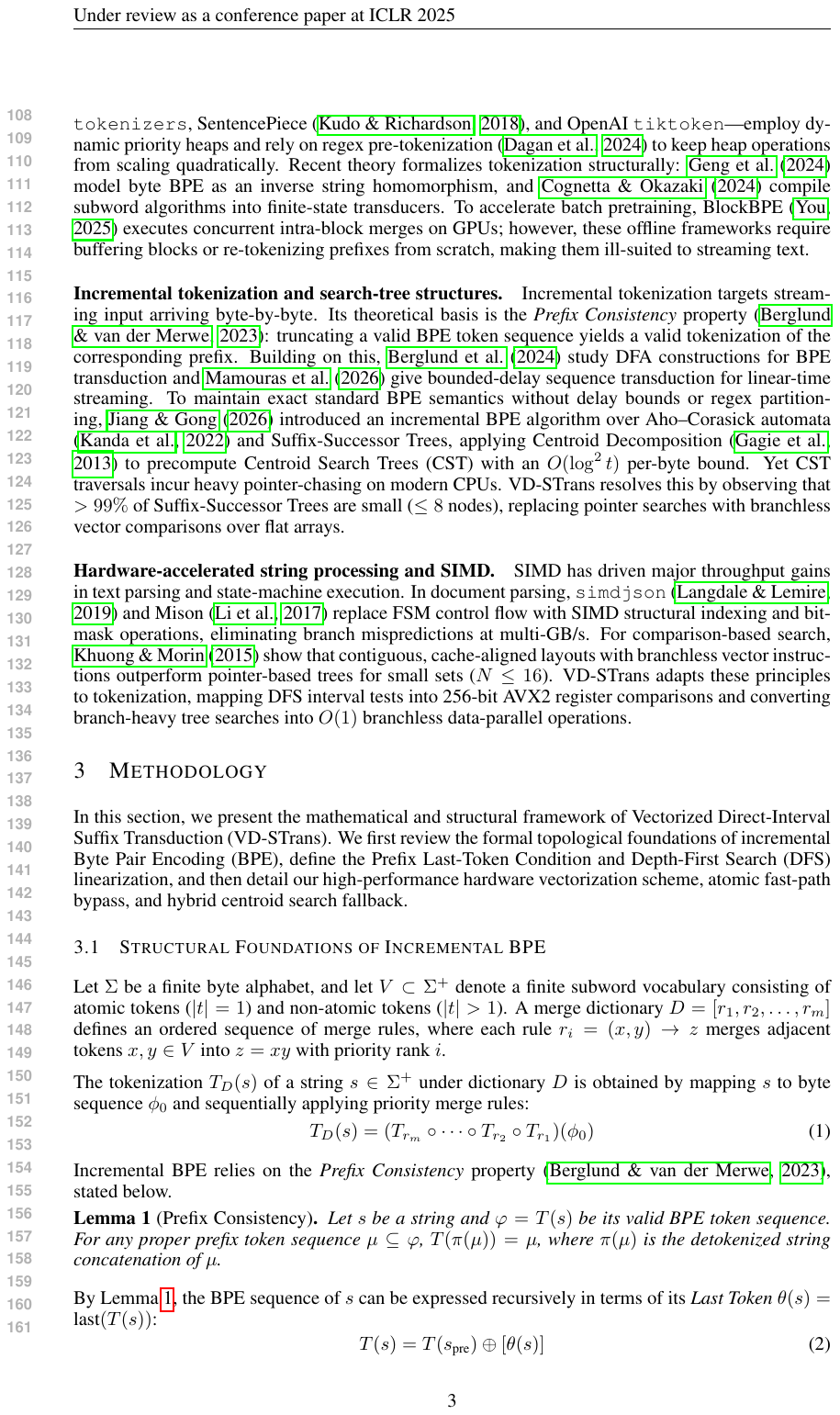} &
    \includegraphics[width=0.2\textwidth]{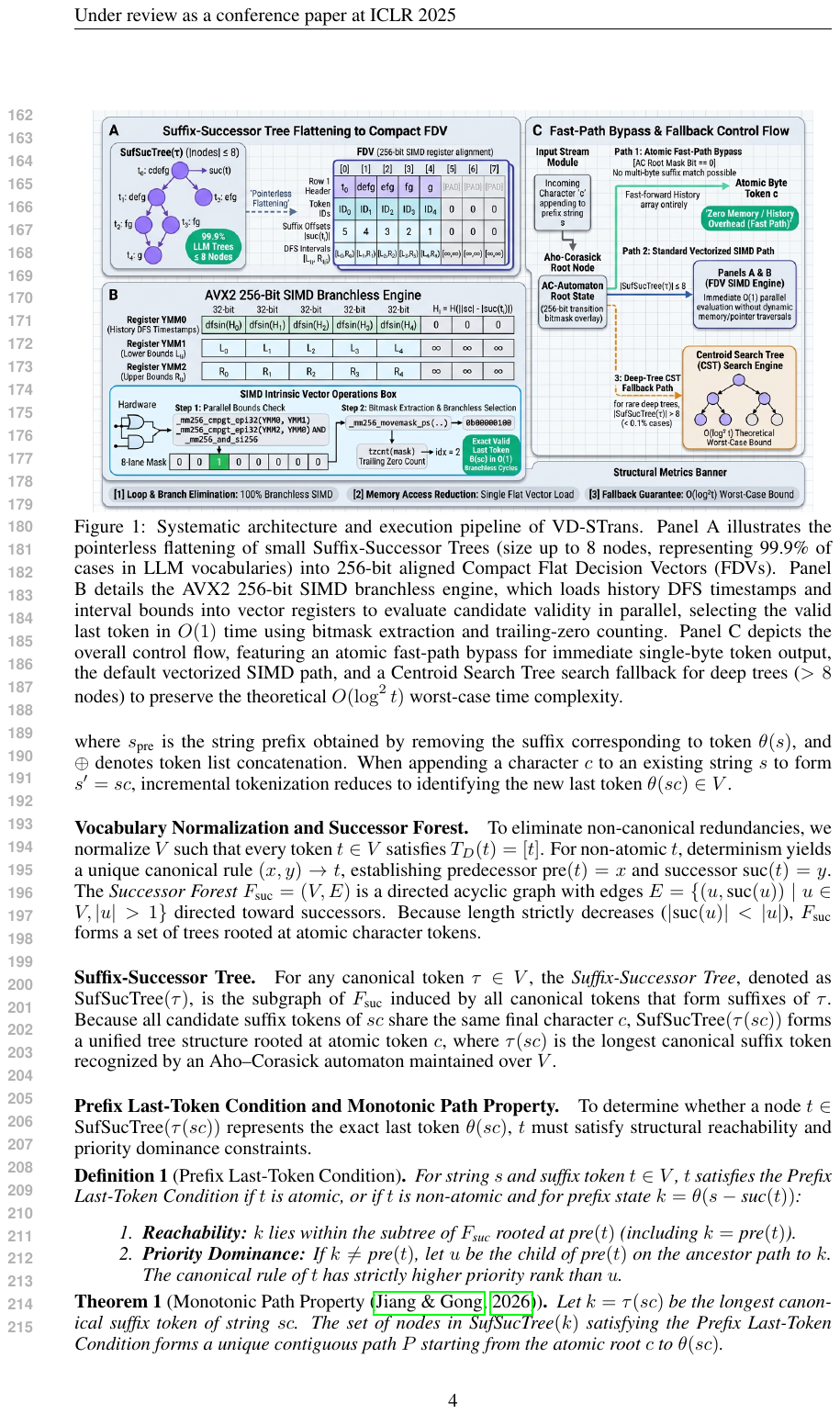} &
    \includegraphics[width=0.2\textwidth]{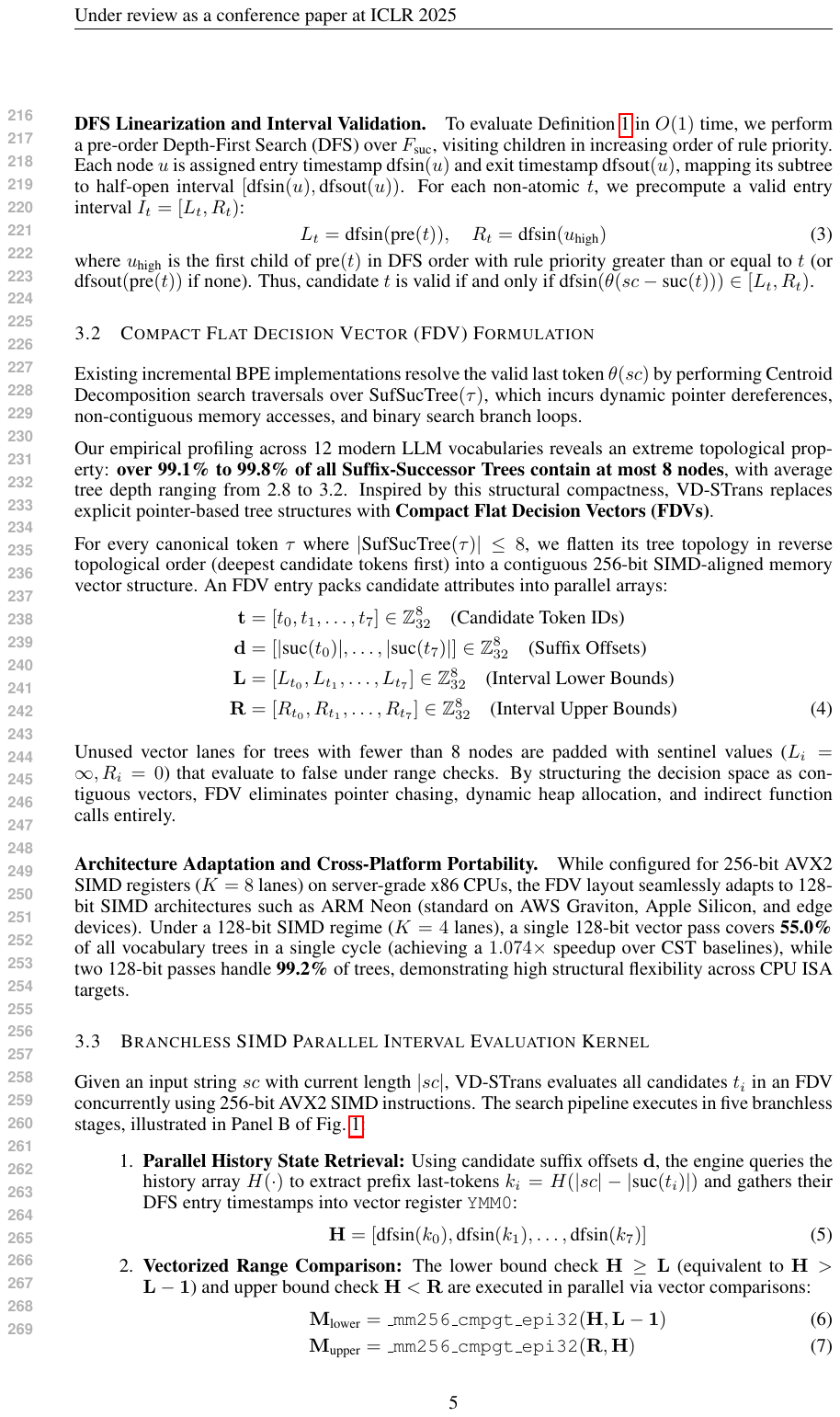} \\
    \hline
    \includegraphics[width=0.2\textwidth]{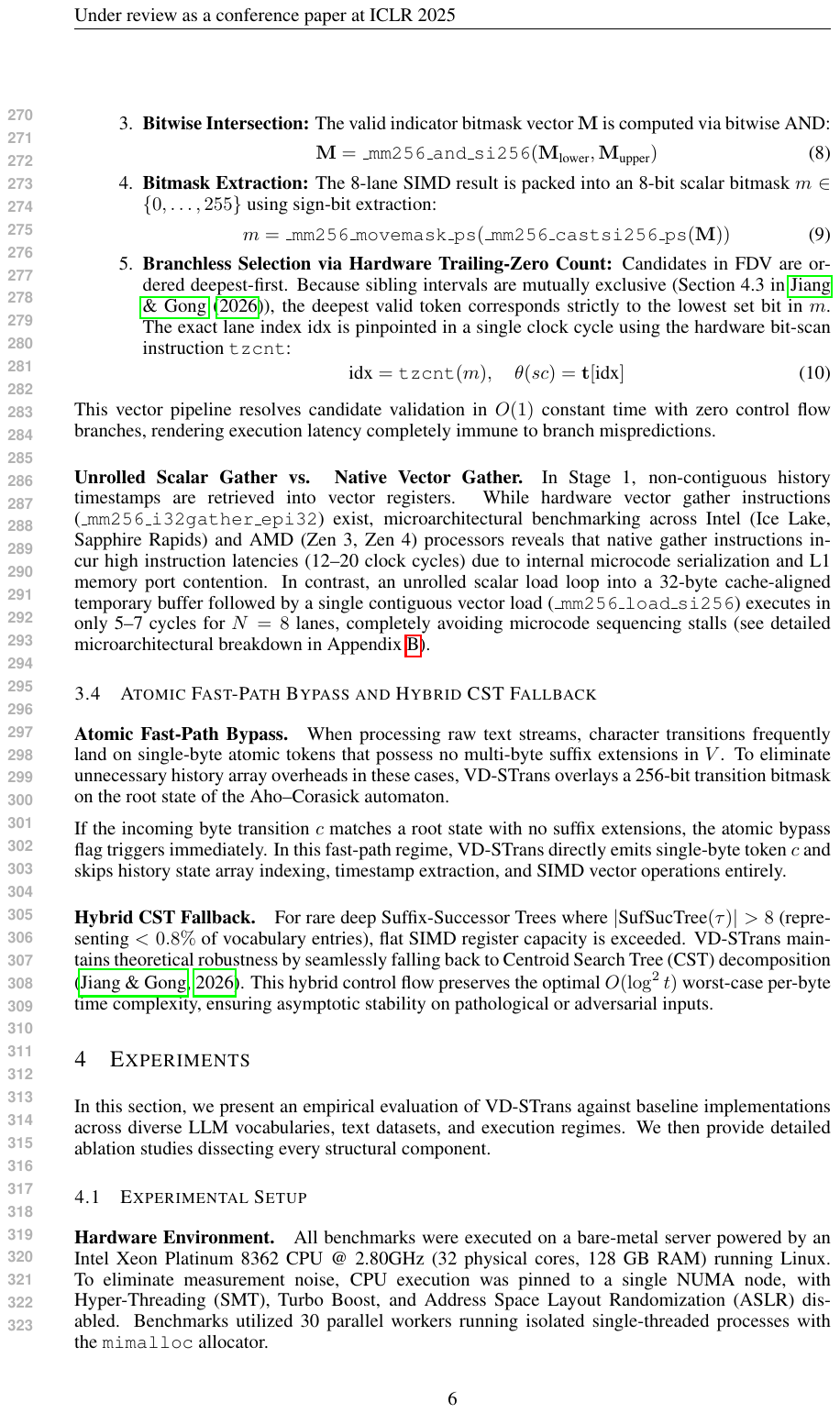} &
    \includegraphics[width=0.2\textwidth]{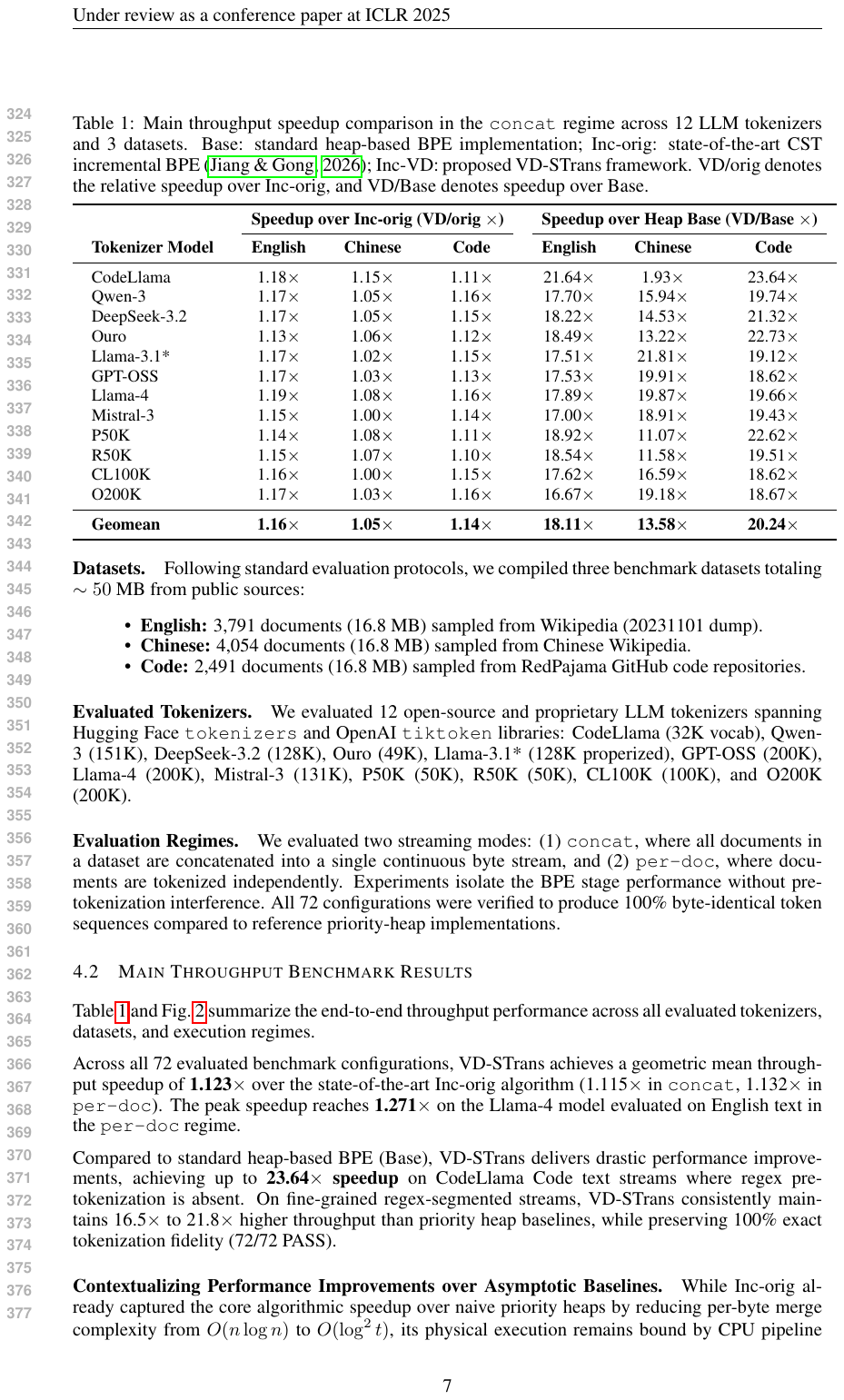} &
    \includegraphics[width=0.2\textwidth]{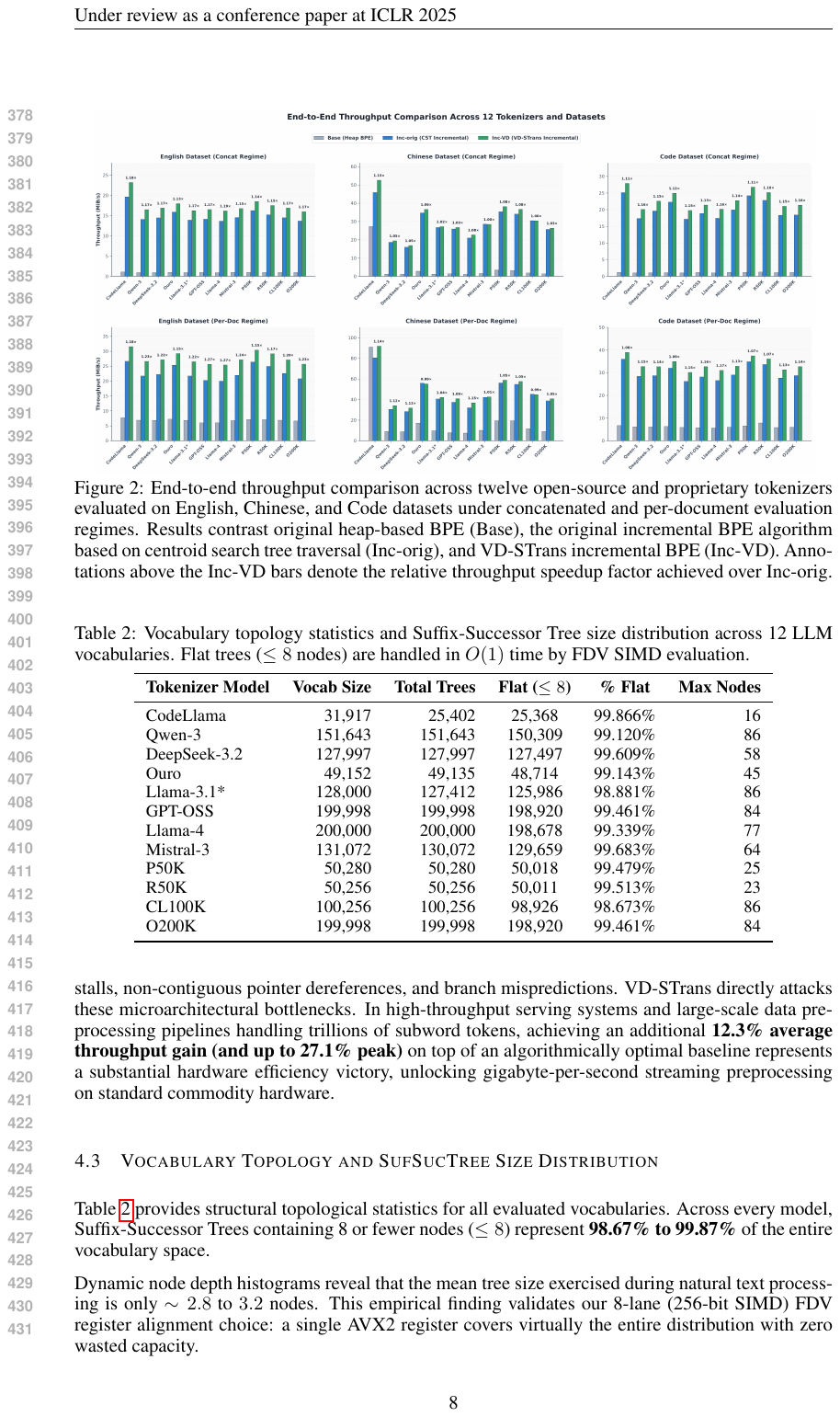} &
    \includegraphics[width=0.2\textwidth]{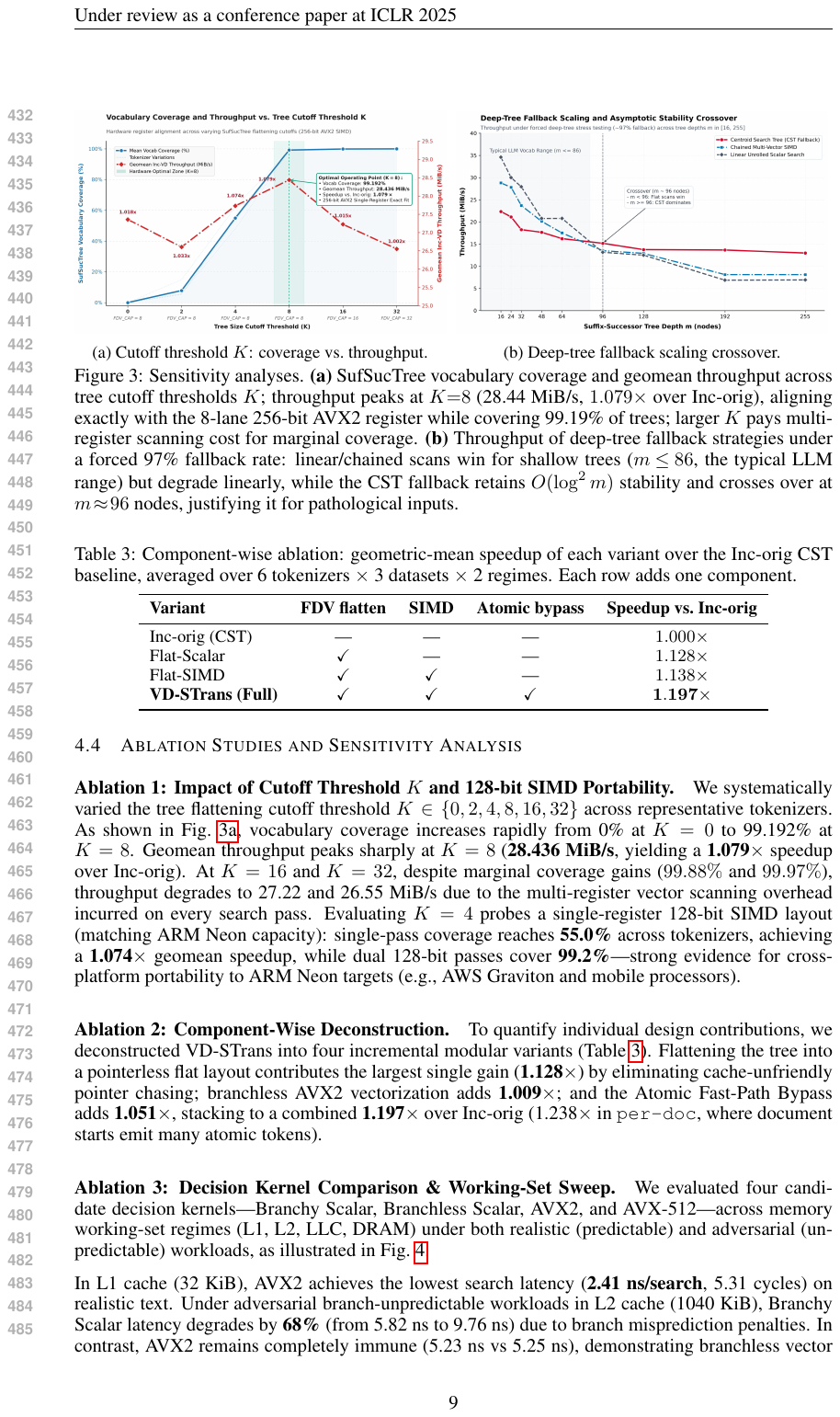} &
    \includegraphics[width=0.2\textwidth]{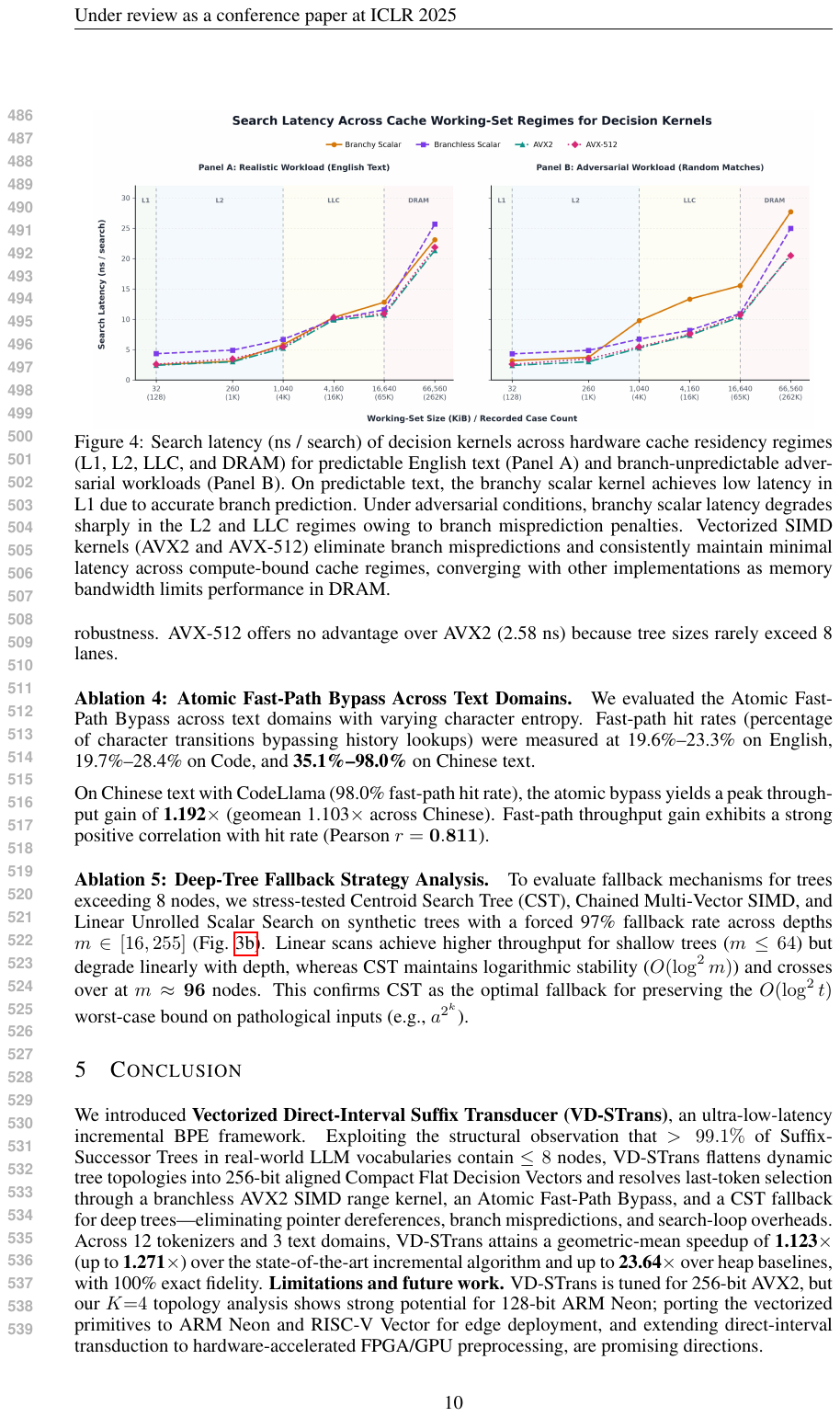}\\
    \hline
  \end{tabular}}
  \caption{\textbf{Case study.} \sname~independently establishes a novel approach that achieves a 10.9\% relative improvement over the human-designed state-of-the-art baseline~\citep{jiang2026incremental}. Moreover, evaluated under the ICLR peer-review procedures, this AI-authored paper is accepted, receiving impressive scores of 8.0 from ScholarPeer and 6.5 from the Stanford Agentic Reviewer.}
  \label{fig:qualitative_results}
\end{figure}

%% file: figures/overview.tex
\begin{figure*}[t]
\centering
\includegraphics[width=\linewidth]{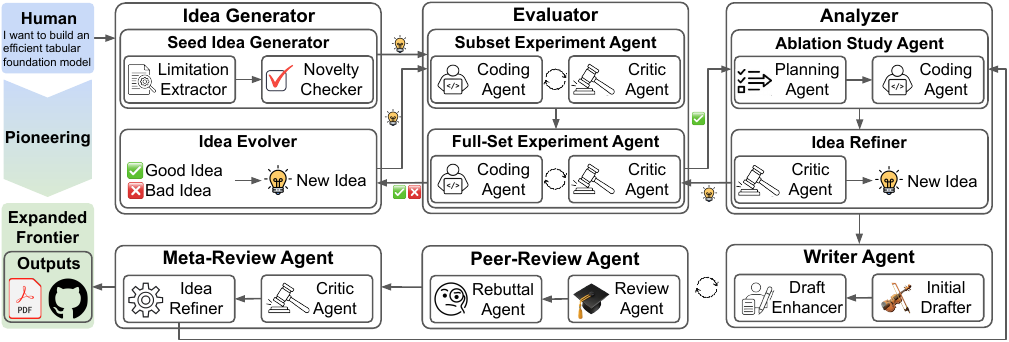}
\caption{\textbf{Overview.} \sname~leverages previous experiments, ablation studies, and reviewer to generate and validate novel ideas that advance human state-of-the-art baselines, verifying each idea across multiple datasets and evaluation metrics. Additionally, it simulates the peer-review process to iteratively refine drafts into publication-ready papers.}
\label{fig:overview}
\end{figure*}

%% file: sections/2_related_works.tex
\section{Related Work}

\noindent\textbf{AI Agents.} The rapid advancement of LLMs has sparked significant interest in autonomous AI agents capable of planning and using tools. Early general-purpose agents such as ReAct \citep{yao2022react}, HuggingGPT \citep{shen2023hugginggpt}, and AutoGen \citep{wu2024autogen} leverage external tools to break down and execute complex, open-ended tasks. In the field of software development, systems such as Voyager \citep{wang2023voyager}, AlphaCode \citep{li2022competition}, SWE-agent \citep{yang2024swe}, and Claude Code \citep{liu2026dive} use interfaces that recognize execution feedback and the environment to iteratively debug and solve coding problems. Recently, this paradigm has expanded into the fields of automated machine learning (ML) engineering and data science. Frameworks such as MLAgentBench \citep{huang2023mlagentbench}, OpenHands \citep{wang2025openhands}, AIDE \citep{jiang2025aide}, MLE-STAR \citep{nam2026mle}, and MARS \citep{chen2026mars} automate end-to-end ML workflows by executing and improving modeling pipelines, while data science agents such as DA-Agent \citep{huang2024code}, Data Interpreter \citep{hong2025data}, and DS-STAR \citep{nam2026ds} tackle heterogeneous data challenges through structured planning and recursive verification. In this paper, we introduce \sname, a specialized AI agent for autonomous research, from idea generation and implementation to the writing of high-quality publishable academic papers.

\noindent\textbf{Autonomous Research Agents.} Autonomous research agents have rapidly evolved from ML problems into sophisticated, multi-stage pipelines that coordinate the entire scientific workflow—from literature-based analysis and hypothesis generation to execution and drafting of research papers. Early end-to-end pipelines, such as AI Scientist \citep{lu2024ai}, established this automation paradigm but suffered from execution instability and issues with hallucinated writing; subsequent versions, such as AI Scientist-v2 \citep{yamada2025ai}, mitigated these problems through a best-first-tree search for experimental branches and review-based reporting. At the same time, modular systems have focused on resolving specific bottlenecks; for example, PaperOrchestra \citep{song2026paperorchestra} compiles unconstrained experiment logs and ideas into LaTeX manuscripts, while ScholarPeer \citep{goyal2026scholarpeer} deploys a multi-agent system that acts as peer-reviewers. Other frameworks introduce human-in-the-loop gates \citep{schmidgall2025agent}, apply evolutionary optimization to algorithmic discovery \citep{novikov2025alphaevolve, lyu2026evoscientist}, or focus on single-metric optimization for a target codebase \citep{li2026autosota, liu2026autoresearchclaw, tang2026ai, weng2025deepscientist}; however, they are generally limited to optimizing a single scalar metric on a single dataset. Crucially, even verifiability-centric systems like ScientistOne \citep{meng2026scientistone} lack the ability to perform an active analytical expansion. Our \sname~addresses these fundamental gaps by conducting systematic experiments on various benchmark datasets and incorporating a dynamic peer-review loop, automating the iterative process of performing complementary ablation experiments, refining ideas, and revising manuscripts, thus producing expert-level papers that meet the rigorous standards of top academic venues.

%% file: sections/3_new_method.tex
\input{tables/pseudo_code}
\input{tables/overview}
\section{\sname: An Expert-Level Autonomous Research Agent}\label{sec:method}

\sname~is an expert-level autonomous research agent that effectively orchestrates specialized AI agents throughout the scientific discovery pipeline. First, \sname~identifies the limitations of the human state-of-the-art method regarding a given scientific problem and generates ideas to address them, filtering for those with high novelty ($\S$~\ref{subsec:seed_idea}). Next, \sname~implements the selected ideas; to optimize computational efficiency, it validates the ideas on a benchmark subset before scaling to the full dataset ($\S$~\ref{subsec:eval_idea}). The agent then refines the ideas based on previous experimental results ($\S$~\ref{subsec:refine_idea}) and conducts ablation studies to analyze the sources of performance gain ($\S$~\ref{subsec:ablating_idea}). During manuscript drafting ($\S$~\ref{subsec:drafting_paper}), \sname~dynamically enhances paper quality through an iterative review process, using a Review Agent to provide critical feedback and a Rebuttal Agent to conduct supplementary experiments. Finally, in the meta-review phase ($\S$~\ref{subsec:meta-reviewing}), a Meta-Review Agent evaluates the manuscript's overall quality and determines whether it meets the submission standards. If further improvements are required, \sname~incorporates feedback, updates idea and ablation analyses, and iterates through the pipeline until the manuscript is approved for submission. For clarity, we abstract all stages (see overview in Table~\ref{tab:overview}) using the Python-style pseudocode in Listing~\ref{lst:stage_pseudocode}.

\noindent\textbf{Problem Setup: Problem-Driven Autonomous Discovery.} Scientific advancement in AI is inherently cumulative: modern progress relies on identifying the limitations of state-of-the-art work and developing methodologies to surpass them. To mirror this realistic research workflow, our goal is to build an autonomous research agent $\mathcal{A}$ capable of executing continuous research iterations. Given a scientific problem $\mathcal{G}$, the framework generates a paper $\mathcal{P}^+$ paired with a reproducible codebase $\mathcal{C}^+$, which represents new frontier knowledge on the given problem. Formally, we define this transformation as:
\begin{equation}
    (\mathcal{P}^+, \mathcal{C}^+) = \mathcal{A}(\mathcal{G}), \quad \text{where} \quad \mathcal{A} = \{\mathcal{A}_1, \mathcal{A}_2, \dots, \mathcal{A}_{N_\mathrm{a}}\}
\end{equation}
where $\mathcal{A}$ comprises $N_\mathrm{a}$ specialized AI agents, each assigned to distinct phases of the research life cycle (e.g., limitation extraction, idea generation, code modification, empirical validation, etc.). 
In addition, $\mathcal{P}^+$ should identify and resolve key methodological or empirical bottlenecks present in $\mathcal{G}$ and $\mathcal{C}^+$ should correctly implement the proposed idea while maintaining execution reproducibility and demonstrating measurable performance gains.

\subsection{Generating Novel Seed Ideas: Targeting Limitation Resolution}\label{subsec:seed_idea}
\input{figures/method_3_2}

\noindent\textbf{Finding Limitations.}
\sname~initiates the research pipeline by identifying the core limitations of the human state-of-the-art method with respect to a given scientific problem $\mathcal{G}$.
Specifically, initially, the Limitation Extractor extracts a set of limitations from $\mathcal{G}$.
Then, the Limitation Verifier verifies whether this set is sufficient to guide novel improvements to $\mathcal{G}$.
If the Limitation Verifier considers the set insufficient, the Limitation Extractor again identifies missing limitations or weaknesses and expands the collection.
This verification loop repeats until the Limitation Verifier confirms that all actionable limitations have been thoroughly extracted or maximum number of iterations is reached.

\noindent\textbf{Generating Novel Seed Ideas.} \sname~first generates an initial idea $h_0$ specifically designed to address the identified limitations, and computes its novelty score using the Novelty Checker. Then, starting with the initial set of candidates $\mathcal{H}_0 = \{h_0\}$, \sname~leverages the Idea Generator Agent to iteratively expand the candidate pool $\mathcal{H}_0$ with distinct, higher-novelty ideas.
This process continues until a collection of $N_{\mathrm{seed}}$ candidate ideas is gathered. 
Finally, the seed ideas in $\mathcal{H}_0= \{h_i\}_{i=0}^{N_\mathrm{seed}-1}$ are sorted in descending order according to their novelty scores $\{s_i\}_{i=0}^{N_{\mathrm{seed}}-1}$, ensuring that $s_i \ge s_j$ whenever $i < j$.
This allows \sname~to prioritize implementing the ideas with the highest originality.

\subsection{Evaluating Ideas: Experimenting from Subset to Full-Set}\label{subsec:eval_idea}


\noindent\textbf{Generating Baseline Results on Subset.}
To establish a reliable reference, \sname~first employs a Baseline Coding Agent to reproduce the primary experiments of $\mathcal{G}$ on the benchmark subset, producing baseline experimental results $\mathcal{E}_{\mathrm{base}}$ alongside a reproducible subset codebase $\mathcal{C}_{\mathrm{base}}$.

\noindent\textbf{Evaluating an Idea on the Subset.}
For a candidate idea $h$ (here, we select only the top-$N_0$ highest-ranked ideas based on $\{s_i\}_{i=0}^{N_0-1}$ from $\mathcal{H}_0$), \sname~leverages a Subset Coding Agent to implement $h$ by modifying $\mathcal{C}_{\mathrm{base}}$, producing the resulting logs $\mathcal{E}_{\mathrm{sub}}^h$ alongside a reproducible codebase $\mathcal{C}_\mathrm{sub}^h$.
Then, to evaluate performance, a Subset Critic Agent compares $\mathcal{E}_{\mathrm{sub}}^h$ against $\mathcal{E}_{\mathrm{base}}$ and emits a categorical decision $d^h$ with feedback $r^h$:
\begin{enumerate}[leftmargin=*, noitemsep, topsep=2pt]
    \item $\mathrm{Bad}$: If performance is substantially inferior to the baseline, $h$ is discarded ($d^h = \mathrm{Bad}$).
    \item $\mathrm{Good}$: If $h$ consistently outperforms the baseline, it is approved for scale-up ($d^h = \mathrm{Good}$).
    \item $\mathrm{Engineer}$: If $h$ shows potential but requires hyperparameter tuning or code adjustments, a Subset Engineering Agent refines $h$ and $\mathcal{C}_{\mathrm{sub}}^h$ guided by strategy $r^h$, generating corresponding results $\mathcal{E}_{\mathrm{sub}}^h$.
\end{enumerate}
This refinement loop repeats until $d^h \in \{\mathrm{Good}, \mathrm{Bad}\}$ or a budget of $N_{\mathrm{eng}}$ iterations is exhausted. If $N_{\mathrm{eng}}$ is reached without achieving $d = \mathrm{Good}$, $h$ is designated as $\mathrm{Bad}$ and pruned—mirroring practical research settings where unpromising avenues are abandoned after bounded optimization.

\input{figures/method_3_3}
\noindent\textbf{Scaling Up to the Full-Set.}
For ideas validated as $\mathrm{Good}$, \sname~scales the evaluation to the full benchmark. A Full-Set Coding Agent adapts $\mathcal{C}_{\mathrm{sub}}^h$ to run across the entire benchmark suite.
A Full-Set Critic Agent and Full-Set Engineer perform final validation and engineering against the full benchmark, producing final execution outputs $\mathcal{E}_{\mathrm{full}}^h$, updated codebase $\mathcal{C}_{\mathrm{full}}^h$, and terminal decision $d^h$.

\noindent\textbf{Unified Coder Interface.}
To streamline subsequent rounds of idea improvement, we abstract this entire idea experiment pipeline into a single high-level Idea Implementer Agent $\mathcal{A}_{\mathrm{Coder}}$. Formally,
\begin{equation}
    h,\mathcal{E}^h, \mathcal{C}^h, d^h, r^h = \mathcal{A}_{\mathrm{Coder}}(\mathcal{G}, h).
\end{equation}

\subsection{Refining Ideas: Improving Ideas Using Experimental Results}\label{subsec:refine_idea}

In the initial evaluation round ($k=0$), \sname~executes the top-$N_0$ seed ideas from $\mathcal{H}_0$ using $\mathcal{A}_{\mathrm{Coder}}$, yielding a set of execution traces $\mathcal{R}_0 = \{(h, \mathcal{E}^h, \mathcal{C}^h, d^h, r^h) \mid h \in \mathcal{H}_0\}$.
To continuously enhance idea quality, we propose an evolution strategy that uses these execution traces as feedback.

\noindent\textbf{Idea Evolution.}
In refinement round $k \ge 1$, \sname~aggregates all historic execution traces $\mathbf{R}_{<k} = \bigcup_{i=0}^{k-1} \mathcal{R}_i$ to generate a set $\mathcal{I}_k$ of $N_k$ evolved ideas.
An Idea Evolver Agent $\mathcal{A}_{\mathrm{Evolve}}$ analyzes both successful results ($d^h = \mathrm{Good}$) and diagnostic failure logs ($d^h = \mathrm{Bad}$) to propose refined hypotheses.

\noindent\textbf{Exploration and Exploitation.}
Relying exclusively on $\mathcal{A}_{\mathrm{Evolve}}$ risks trapping the optimization process in local optima centered around early seed ideas.
To ensure broad coverage of the solution space, \sname~complements the evolved set $\mathcal{I}_k$ with $N_e$ previously unevaluated seed ideas drawn from $\mathcal{H}_0$ in descending order of novelty score.
Formally, the total candidate pool $\mathcal{H}_k$ for round $k$ is $\mathcal{I}_k \cup \mathcal{H}_{0}^{(k)}$,where $\mathcal{H}_{0}^{(k)} \subset \mathcal{H}_0$ contains the next $N_e$ highest-ranked unevaluated seed ideas.

\input{figures/method_3_4}
\noindent\textbf{Experimenting Ideas.}
For each candidate $h \in \mathcal{H}_k$, \sname~invokes $\mathcal{A}_{\mathrm{Coder}}$ to evaluate the idea:
\begin{equation}
    \mathcal{R}_k = \left\{ \left(h, \mathcal{E}^h, \mathcal{C}^h, d^h, r^h\right) \;\middle|\; h, \mathcal{E}^h, \mathcal{C}^h, d^h, r^h = \mathcal{A}_{\mathrm{Coder}}(\mathcal{G}, h), \; h \in \mathcal{H}_k \right\}.
\end{equation}
This evolutionary loop iterates until $S$ successful ideas ($d^h = \mathrm{Good}$) are collected, or the maximum refinement limit $K$ is reached.
Formally, idea refinement terminates successfully when $\sum_{i=0}^k \sum_{h \in \mathcal{H}_i} \mathbb{I}(d^h = \mathrm{Good}) \ge S$, where $\mathbb{I}(\cdot)$ is the indicator function.
If round $K$ is reached with zero successful ideas ($\sum_{i=0}^K \sum_{h \in \mathcal{H}_i} \mathbb{I}(d^h = \mathrm{Good}) = 0$), \sname~terminates the entire process.

\noindent\textbf{Selecting the Best Idea.}
Upon discovering at least one successful idea, \sname~selects the optimal candidate $h_{\mathrm{best}}$ for downstream ablation analysis.
A Selector Agent $\mathcal{A}_{\mathrm{Selector}}$ compares performance metrics and execution logs across all validated ideas $\{ h \mid d^h = \mathrm{Good} \}$ evaluated on the full benchmark:
\begin{equation}
    h_{\mathrm{best}}, \mathcal{E}_{\mathrm{best}}, \mathcal{C}_{\mathrm{best}} = \mathcal{A}_{\mathrm{Selector}}\left(\mathcal{G}, \left\{ (h, \mathcal{E}^h, \mathcal{C}^h) \;\middle|\; d^h = \mathrm{Good} \right\}\right).
\end{equation}

\subsection{Ablation Studies: Analyzing Source of Gain and Refining Ideas}\label{subsec:ablating_idea}
Once the optimal candidate idea $h_{\mathrm{best}}$ is selected, \sname~conducts systematic component-level ablation studies.
This phase isolates the explicit sources of empirical gain for scientific interpretation, and leverages fine-grained ablation feedback to perform an additional round of idea refinement.

\noindent\textbf{Ablation Planning and Execution.}
To evaluate individual components, an Ablation Planner Agent automatically formulates a set of $N_p$ executable ablation plans $\{p_1, \dots, p_{N_p}\}$ tailored to $h_{\mathrm{best}}$.
For each ablation plan $p_i$, an Ablation Coding Agent modifies the validated codebase $\mathcal{C}_{\mathrm{best}}$ to execute, yielding an ablation outcome $c_i$.
The aggregated ablation results are collected as $\mathcal{E}_{\mathrm{abl}} = \{c_1, \dots, c_{N_p}\}$.

\noindent\textbf{Refining the Idea via Ablation Insights.}
Emulating expert human research practices, where removing redundant or counterproductive components often yields a better method, \sname~uses $\mathcal{E}_{\mathrm{abl}}$ to further refine $h_{\mathrm{best}}$.
An Ablation Critic Agent $\mathcal{A}_{\mathrm{AblCritic}}$ inspects the component breakdown to determine whether $h_{\mathrm{best}}$ is optimal or requires additional modification, generating $d_{\mathrm{abl}}, r_{\mathrm{abl}}$ where $d_{\mathrm{abl}} \in \{\mathrm{Good}, \mathrm{Refine}\}$.
If $d_{\mathrm{abl}} = \mathrm{Good}$, $h_{\mathrm{best}}$ is finalized and passed to the paper drafting stage.
If $d_{\mathrm{abl}} = \mathrm{Refine}$, \sname~re-engages the Full-Set Engineering Agent $\mathcal{A}_{\mathrm{FullEng}}$ to produce a refined hypothesis $h_{\mathrm{new}}$, updated results $\mathcal{E}_{\mathrm{new}}$, and revised codebase $\mathcal{C}_{\mathrm{new}}$ guided by critique $r_{\mathrm{abl}}$.

\noindent\textbf{Robust Verification and Loop.}
Because structural refinements do not guarantee improved performance, \sname~verifies whether $\mathcal{E}_{\mathrm{new}}$ strictly outperforms $\mathcal{E}_{\mathrm{best}}$.
The baseline variables are updated—$(h_{\mathrm{best}}, \mathcal{E}_{\mathrm{best}}, \mathcal{C}_{\mathrm{best}}) \leftarrow (h_{\mathrm{new}}, \mathcal{E}_{\mathrm{new}}, \mathcal{C}_{\mathrm{new}})$—if and only if $\mathcal{E}_{\mathrm{new}}$ is preferred than $\mathcal{E}_{\mathrm{best}}$ by the Result Comparison Agent.
When an update occurs, \sname~re-executes the ablation planning phase on the updated candidate.
This refinement loop repeats for a maximum of $N_{\mathrm{abl}}$ iterations or until $\mathcal{A}_{\mathrm{AblCrit}}$ emits $d_{\mathrm{abl}} = \mathrm{Good}$, ensuring a fully optimized hypothesis prior to manuscript generation.

\input{figures/method_3_567}
\subsection{Manuscript Drafting: Simulating the Peer-Review Process}\label{subsec:drafting_paper}
\noindent\textbf{Initial Drafting.}
First, an Initial Drafter Agent $\mathcal{A}_{\mathrm{Draft}}$ (incorporating PaperOrchestra; \citealp{song2026paperorchestra}) synthesizes the selected idea $h_{\mathrm{best}}$, main benchmark results $\mathcal{E}_{\mathrm{best}}$, and component ablation studies $\mathcal{E}_{\mathrm{abl}}$ into a full, conference-formatted manuscript $\mathcal{P}_{\mathrm{new}}$.

\noindent\textbf{Enhancing the Draft via Simulated Review-Rebuttal.}
To rigorously elevate manuscript quality, \sname~simulates an interactive peer-review and rebuttal process.
A Peer-Reviewer Agent $\mathcal{A}_{\mathrm{Reviewer}}$ (i.e., ScholarPeer; \citealp{goyal2026scholarpeer}) critically evaluates $\mathcal{P}_{\mathrm{new}}$, generating a detailed evaluation $\mathcal{R}_{\mathrm{new}}$ containing identified strengths, weaknesses, targeted questions, and an overall numerical score $s_\mathrm{review} \in [1, 10]$ based on the standard ICLR grading scale.
If the $s_\mathrm{review}$ is below the acceptance threshold (e.g., 8), \sname~initiates an automated rebuttal stage to address reviewer concerns.
A Rebuttal Planner Agent $\mathcal{A}_{\mathrm{RebPlan}}$ analyzes $\mathcal{R}_{\mathrm{new}}$ to formulate a set of $N_t$ supplementary experimental tasks $\{t_1, \dots, t_{N_t}\}$ designed to resolve reviewer queries.
Next, a Rebuttal Coding Agent $\mathcal{A}_{\mathrm{RebCoder}}$ implements and executes each planned task $t_i$ using codebase $\mathcal{C}_{\mathrm{best}}$, generating supplementary results $e_i$,
yielding the aggregated supplementary experiment results $\mathcal{E}_{\mathrm{reb}} = \{e_1, \dots, e_{N_t}\}$.

\noindent\textbf{Iterative Enhancement.}
A Paper Enhancer Agent $\mathcal{A}_{\mathrm{Enhancer}}$ integrates the $\mathcal{R}_{\mathrm{new}}$ and supplementary findings $\mathcal{E}_{\mathrm{reb}}$ into $\mathcal{P}_{\mathrm{new}}$, revising narrative claims and updating empirical tables and figures.
The updated manuscript is re-evaluated by $\mathcal{A}_{\mathrm{Reviewer}}$, updating $\mathcal{R}_{\mathrm{new}}$ and score $s_\mathrm{review}$.
This review-rebuttal cycle repeats until $s_\mathrm{new} \ge 8$ or the maximum budget of $N_{\mathrm{peer}}$ review iterations is reached, producing a polished, thoroughly validated final manuscript $\mathcal{P}_{\mathrm{new}}$.

\subsection{Meta-Reviewing: Final Assessment and Review-Driven Refinement}\label{subsec:meta-reviewing}
To mirror the complete lifecycle of academic publishing, \sname~integrates a Meta-Review Agent $\mathcal{A}_{\mathrm{Meta}}$. Specifically, $\mathcal{A}_{\mathrm{Meta}}$ evaluates the revised manuscript $\mathcal{P}_{\mathrm{new}}$ alongside the generated peer review $\mathcal{R}_{\mathrm{new}}$ to make a final publication assessment and generate actionable strategic feedback.

\noindent\textbf{Meta-Review Decision.}
Formally, $\mathcal{A}_{\mathrm{Meta}}$ takes the paper draft and reviewer feedback as inputs to produce a decision $d_{\mathrm{meta}}$ and meta-critique $r_{\mathrm{meta}}$, where $d_{\mathrm{meta}}\in \{\mathrm{Accept}, \mathrm{Refine}\}$.
If $d_{\mathrm{meta}} = \mathrm{Accept}$, $\mathcal{P}_{\mathrm{new}}$ is judged to meet top-tier conference standards.
\sname~finalizes the process and exports the final improved paper and its codebase as $\mathcal{P}^+ \leftarrow \mathcal{P}_{\mathrm{new}}$ and $\mathcal{C}^+ \leftarrow \mathcal{C}_{\mathrm{best}}$.

\noindent\textbf{Review-Driven Idea Refinement.}
When $d_{\mathrm{meta}} = \mathrm{Refine}$, indicating that the meta-reviewer identified a critical algorithmic or empirical weakness, \sname~initiates a deep idea refinement phase.
Using meta-critique $r_{\mathrm{meta}}$ as guidance, the Full-Set Engineering Agent $\mathcal{A}_{\mathrm{FullEng}}$ updates $h_{\mathrm{best}}$ to construct a revised hypothesis $h_{\mathrm{new}}$, updated results $\mathcal{E}_{\mathrm{new}}$, and modified codebase $\mathcal{C}_{\mathrm{new}}$.

\noindent\textbf{Verification and Re-Drafting.}
To ensure that the meta-review modification yields true scientific progression, \sname~re-evaluates $\mathcal{E}_{\mathrm{new}}$ against $\mathcal{E}_{\mathrm{best}}$ using the Result Comparison Agent.
If $\mathcal{E}_{\mathrm{new}}$ is verified as strictly superior , \sname~updates the core state $(h_{\mathrm{best}}, \mathcal{E}_{\mathrm{best}}, \mathcal{C}_{\mathrm{best}}) \leftarrow (h_{\mathrm{new}}, \mathcal{E}_{\mathrm{new}}, \mathcal{C}_{\mathrm{new}})$.
Because the fundamental idea has changed, \sname~re-executes downstream ablation planning, ablation execution, manuscript re-drafting, and simulated peer-review cycles.
Conversely, if $\mathcal{E}_{\mathrm{new}}$ fails to outperform the baseline, the refinement is discarded, and \sname~terminates the process using the previous best outputs ($\mathcal{P}^+ \leftarrow \mathcal{P}_{\mathrm{new}}, \mathcal{C}^+ \leftarrow \mathcal{C}_{\mathrm{best}}$).
This meta-refinement loop repeats for a maximum of $N_{\mathrm{meta}}$ iterations or until $d_{\mathrm{meta}} = \mathrm{Accept}$, yielding a rigorously validated final contribution.

%% file: tables/pseudo_code.tex

\begin{listing}
\caption{\textbf{Python-style pseudocode for each pipeline stage.} At each stage, \sname~evaluates candidate artifacts using a critic agent. If accepted, the artifact is
returned; if rejected, it is discarded; otherwise, it is iteratively refined based on the critic's feedback up to a maximum number of rounds.}\label{lst:stage_pseudocode}
\begin{lstlisting}[style=pseudopython]
def stage(candidate, critic, refine, max_rounds):
    for _ in range(max_rounds):
        verdict, feedback = critic(candidate)
        if verdict == "accept":
            return candidate
        elif verdict == "refine":
            candidate = refine(candidate, feedback)
        else:
            return None
    return None
\end{lstlisting}
\end{listing}

%% file: tables/overview.tex
\begin{table}[t!]
\centering
\caption{\textbf{\sname~overview.} At each stage, \sname~generates a \texttt{candidate} using a specialized AI agent and employs a corresponding \texttt{critic} agent to decide whether to accept the output or invoke a \texttt{refine}ment agent to improve and supplement it.}\label{tab:overview}
\resizebox{\textwidth}{!}{
\begin{tabular}{llll}
\toprule
\texttt{stage} & \texttt{candidate} & \texttt{critic} & \texttt{refine}\\
\midrule
\multicolumn{4}{l}{\textbf{\textit{Generating Novel Seed Ideas ($\S$~\ref{subsec:seed_idea})}}}\\
Finding limitations & Set of limitations & \textit{Can guide novel improvement?} & \textit{Add missing limitations}\\
Seed idea generation & Seed ideas & \textit{Is it novel?} & \textit{Add more novel ideas}\\
\midrule
\multicolumn{4}{l}{\textbf{\textit{Evaluating Ideas ($\S$~\ref{subsec:eval_idea})}}}\\
Reproduce baseline on subset & SOTA baseline result & - & - \\
Idea experiment on subset & Idea, Code, Results & \textit{Is it better than the reproduced baseline?} & \textit{Refine idea through engineering}\\
Idea experiment on full-set & Idea, Code, Results & \textit{Is it better than the original SOTA result?} & \textit{Refine idea through engineering}\\
\midrule
\multicolumn{4}{l}{\textbf{\textit{Refining Ideas ($\S$~\ref{subsec:refine_idea}})}}\\
Idea evolution & Evolved idea & \textit{Idea experiment} & \textit{Evolve idea from traces}\\
Select best idea & Best idea & \textit{What is the best idea from traces?} & - \\
\midrule
\multicolumn{4}{l}{\textbf{\textit{Ablation Studies ($\S$~\ref{subsec:ablating_idea}})}}\\
Ablation study & Idea, Ablation results & \textit{Is the component breakdown clean?} & \textit{Refine the method}\\
\midrule
\multicolumn{4}{l}{\textbf{\textit{Drafting \& Meta-Reviewing ($\S$~\ref{subsec:drafting_paper}, $\S$~\ref{subsec:meta-reviewing})}}}\\
Initial drafting & Manuscript & - & - \\
Peer-Review & Manuscript & \textit{Is review score good enough?} & \textit{Run rebuttal experiments}\\
Meta-Review & Manuscript, Review & \textit{Does it meet the venue bar?} & \textit{Refine idea and analyze again}\\
\bottomrule
\end{tabular}}
\end{table}

%% file: figures/method_3_2.tex
\begin{figure*}[t]
\centering
\includegraphics[width=\linewidth]{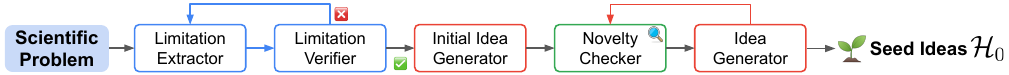}
\caption{\textbf{Generating Novel Seed Ideas.} \sname~begins its research by generating novel ideas that can address the limitations of the human state-of-the-art (see $\S$~\ref{subsec:seed_idea}).}
\label{fig:method_3_2}
\end{figure*}

%% file: figures/method_3_3.tex
\begin{figure*}[t]
\centering
\includegraphics[width=\linewidth]{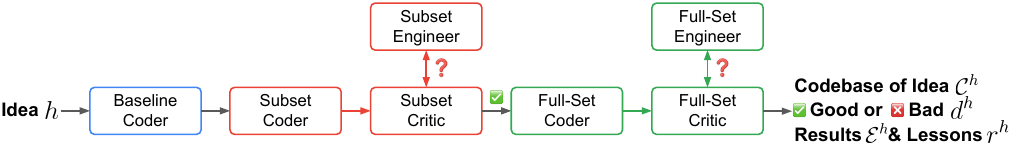}
\caption{\textbf{Evaluating Ideas.} \sname~utilizes $\mathcal{A}_\mathrm{Coder}$ to implement the generated idea through subset testing, critic evaluation, engineering refinement loops, and full-set scaling (see $\S$~\ref{subsec:eval_idea}).}
\label{fig:method_3_3}
\end{figure*}

%% file: figures/method_3_4.tex
\begin{figure*}[t]
\centering
\includegraphics[width=\linewidth]{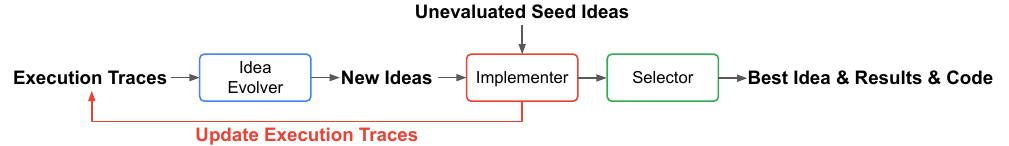}
\caption{\textbf{Refining Ideas.} \sname~improves ideas based on prior experimental results, ultimately selecting the best hypothesis for ablation studies and paper drafting (see $\S$~\ref{subsec:refine_idea}).}
\label{fig:method_3_4}
\end{figure*}

%% file: figures/method_3_567.tex
\begin{figure*}[t]
\centering
\includegraphics[width=\linewidth]{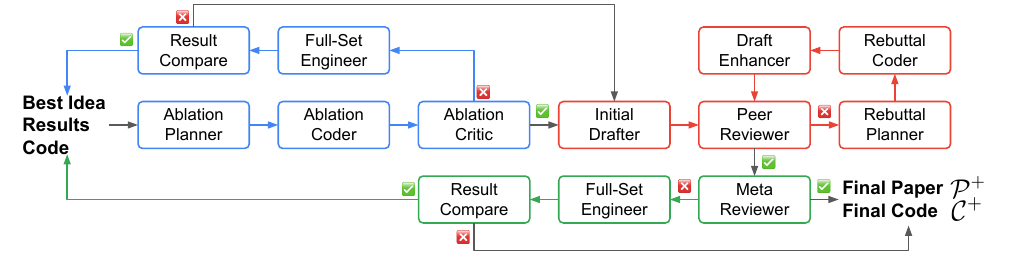}
\caption{\textbf{From Ablation Study to Drafting.} \sname~conducts ablation studies and simulates the peer-review and rebuttal processes. Beyond this, \sname~dynamically refines the idea and enhances paper quality using feedback from each stage (see $\S$~\ref{subsec:ablating_idea}, $\S$~\ref{subsec:drafting_paper}, $\S$~\ref{subsec:meta-reviewing}).}
\label{fig:method_3_567}
\end{figure*}

%% file: sections/4_experiment.tex
\section{Experiments}\label{sec:experiments}

In this section, we empirically validate the effectiveness of \sname. Specifically, in $\S$~\ref{subsec:main_result}, we present quantitative and qualitative comparisons against existing autonomous research agents. In $\S$~\ref{subsec:ablation}, we conduct component-wise ablation studies to evaluate each constituent part of \sname. In $\S$~\ref{subsec:discussion}, we provide a discussion including cost analysis and case studies.

\noindent\textbf{Common Setup.} We evaluate the effectiveness of \sname\ on 107 scientific problems drawn from top machine learning venues, including ICLR, ICML, and NeurIPS (see Appendix~\ref{subapp:benchmark} for details). All experiments are conducted using Gemini 3.6 Flash and Claude Opus 4.8, unless otherwise specified. For evaluation, we primarily report review scores from automated AI review agents: ScholarPeer \citep{goyal2026scholarpeer} and Stanford Agentic Reviewer\footnote{\url{https://paperreview.ai/}}. ScholarPeer serves as an in-distribution evaluation, as it is also used to refine the draft quality generated by \sname. Conversely, Stanford Agentic Reviewer serves as a held-out evaluator that was unseen during development by both the baselines and our method. Full implementation details and configurations are provided in Appendix~\ref{subapp:config}.

\subsection{Main Results}\label{subsec:main_result}
\input{tables/ai_scientist_comparison}
\noindent\textbf{Quantitative Comparison.} As shown in Table~\ref{tab:vs_ai_scientist}, \sname~achieves a 91.9\% acceptance rate under ScholarPeer (nearly doubling the review score from ScientistOne's 3.8 to 7.5). More importantly, while all baselines fail to achieve acceptance from the Stanford Agentic Reviewer, \sname~is the only agent that produces research where 72.1\% of generated papers meet high acceptance standards. This result shows that while current autonomous research agents cannot produce expert-level research results, \sname~possesses such capabilities.

\input{figures/qual_compare_main}
\noindent\textbf{Qualitative Comparison.}
To assess the depth and completeness of the generated manuscripts, we compare \sname~against the strongest baseline, ScientistOne. As illustrated in Figure~\ref{fig:qual_compare_main}, \sname~demonstrates substantially broader experimental coverage and analytical rigor. While ScientistOne generates only 4 figures and 2 tables—evaluating a single metric on an isolated benchmark with limited baselines—\sname~produces 9 figures and 12 tables (including the Appendix). Furthermore, \sname~reports multi-metric evaluations across diverse datasets and incorporates an extensive set of baseline comparisons, reflecting a publication-ready experimental design.

\input{tables/conference_accepted_comparison}
\noindent\textbf{Comparison with Human Researchers.} We evaluate the research quality produced by \sname~against accepted publications. Specifically, we use ScholarPeer and the Stanford Agentic Reviewer to evaluate papers accepted at NeurIPS 2025, ICLR 2026, and ICML 2026 (including spotlight presentations), whose problem specifications and codebases serve as benchmark tasks for \sname. As shown in Table~\ref{tab:vs_conference}, \sname~successfully executes research on 86 out of 107 problems, achieving an 80.4\% success rate. Furthermore, papers generated by \sname~surpass the average scores of accepted papers at ICLR 2026 and NeurIPS 2025 under both ScholarPeer and Stanford Agentic Reviewer. While \sname~does not yet achieve spotlight-level quality, these results still demonstrate that it functions as an expert-level research agent capable of producing manuscripts that meet the acceptance threshold of top-tier AI venues. Finally, we include comparisons with AI-generated papers accepted at Agent4Science 2025—the first venue dedicated to AI-generated research—to demonstrate that prior systems were incapable of meeting top conference standards.

\input{tables/autosota_comparison}
\noindent\textbf{Comparison with AutoSOTA.} While AutoSOTA~\citep{li2026autosota} operates in a distinct setting, i.e., modifying existing codebases to optimize a single scalar metric, \sname~is designed for end-to-end scientific paper generation. For completeness, we provide a comparison based on the performance gains reported over human state-of-the-art baselines. To extract these gains systematically, we parse the main tables for 10 times using Gemini 3.6 Flash and averaged them. As shown in Table~\ref{tab:vs_autosota}, \sname~shows superior performance to AutoSOTA in terms of average and median improvement across average of all tasks and NeurIPS 2025 tasks, but performance drops slightly in the ICLR 2026 tasks.
We hypothesize that this advantage stems from \sname~proposing novel methodological innovations to overcome baseline limitations, whereas AutoSOTA relies primarily on searching within the existing hyperparameter and execution space to improve the given method. Appendix~\ref{sec:autosota} examines the five ICLR 2026 papers one by one and supports this account: none of AutoSOTA's five changes introduces a new algorithmic component, and each is a configuration-level edit of at most a few lines.

\subsection{Ablation Studies}\label{subsec:ablation}

Here, we evaluate on the 49 target problems sourced from ICML 2026 Spotlight papers.

\input{figures/abl_idea_improvement}
\noindent\textbf{Effectiveness of Idea Evolution.} As shown in Figure~\ref{fig:abl_idea_improvement}(a), the relative gain over the human state-of-the-art baseline steadily increases as \sname~proceeds through iterative idea refinement. In particular, the magnitude of improvement is most pronounced during the early refinement stages.

\noindent\textbf{Balancing Exploration and Exploitation.} As shown in Figure~\ref{fig:abl_idea_improvement}(b), the majority of the top-performing ideas chosen by the Selector Agent are identified in the early stages. This indicates that the initial seed ideas generated by \sname~to overcome the limitations of human state-of-the-art methods are already well-formed and competitive. Conversely, when seed ideas yield marginal gains or fail, the Idea Evolver Agent leverages these experimental traces to evolve them into stronger hypotheses that overcome baseline shortcomings. Consequently, as refinement progresses, newly selected best ideas are substantially more likely to be drawn from evolved candidates rather than original seed ideas (as evidenced by the dominant share of the blue bars after the initial round).

\input{tables/ablation_rebuttal}
\noindent\textbf{Effectiveness of the Rebuttal Agent.} During the simulated dynamic peer-review phase, \sname~leverages the Rebuttal Agent to conduct supplementary experiments and uses the findings to enhance the draft. As shown in Table~\ref{tab:abl_rebuttal}, this rebuttal procedure effectively addresses the weaknesses flagged by ScholarPeer.
While it is not surprising that \sname~gradually achieves good scores on ScholarPeer, as we utilize ScholarPeer reviews, we found that our peer-review simulation framework is generalized across Review Agent. Specifically, incorporating ScholarPeer reviews also improves acceptance rate from Stanford Agentic Reviewer, with this effect being particularly pronounced in the first review round.
Furthermore, the first and second rows of Table~\ref{tab:abl_rebuttal} demonstrate that even without the Rebuttal Agent, \sname~produces initial drafts of significantly higher quality than the prior state-of-the-art baseline, ScientistOne. This indicates that the preceding autonomous research cycle—specifically the holistic benchmark reasoning, idea refinement, and ablation-driven hypothesis refinement—is highly effective for drafting a robust manuscript, achieving a nearly 50\% acceptance rate when evaluated by both ScholarPeer and Stanford Agentic Reviewer.

\input{tables/ablation_review_refine}
\noindent\textbf{Effectiveness of Review-Driven Idea Refinement.} In the final stage, \sname~employs the Meta-Review Agent to evaluate whether the generated paper meets the acceptance standards of a top-tier venue. If the agent indicates that further improvement is required, \sname~refines the underlying idea by incorporating the review as feedback. As shown in Table~\ref{tab:abl_meta_review}, this refinement process enables \sname~to generate more effective ideas that advance beyond the current knowledge frontier. For instance, prior to review-driven refinement, \sname~already established a new state-of-the-art method, LFR-Engram, outperforming the human baseline Engram \citep{kwon2026ai}. However, the Meta-Review Agent deemed LFR-Engram insufficient to meet expert standards. By leveraging the reviewer feedback for refinement, \sname~subsequently synthesized FCD-Engram, which consistently outperforms LFR-Engram. This case study demonstrates that review-driven refinement is essential for producing more novel, high-impact research ideas.

\input{tables/ablation_audit}
\noindent\textbf{CoE Integrity Audit.} The CoE Integrity Audit \citep{meng2026scientistone} is a post-hoc evaluation framework that verifies whether claims in a generated paper are supported by its artifacts: code, empirical outputs, and bibliography. It comprises four integrity checks: (1)~\textit{Score verification}, which evaluates codebase reproducibility by comparing reported scores against those obtained from re-executing the repository; (2)~\textit{Specification compliance}, which ensures that solution code adheres strictly to task rules without reward hacking; (3)~\textit{Reference verification}, which guarantees that the bibliography contains no hallucinated citations; and (4)~\textit{Method-code alignment}, which ensures that the paper faithfully and accurately describes the codebase implementation. Following \citet{meng2026scientistone}, AI-generated papers and their corresponding codebases must be rigorously verifiable across all four dimensions.

To guarantee these properties, we introduce three dedicated refinement agents alongside careful agent prompt design. For score verification, the Coding Agent is prompted during the experimentation phase to output self-contained, reproducible scripts and execution instructions, ensuring full reproducibility without requiring additional post-hoc refinement. For specification compliance, we introduce a validation filter that uses the Coding Agent to detect and discard rule-violating solutions immediately after experimentation. For reference verification, a search-augmented LLM identifies hallucinated citations, enabling the Writer Agent to ground and correct the bibliography using live search results. Finally, for method-code alignment, the Coding Agent audits the repository against the manuscript to produce an audit report, which the Writer Agent then uses to rectify any discrepancies in the method section. As shown in Table~\ref{tab:abl_coe}, \sname~faithfully passes all four audits; removing these refinement agents leads to sporadic audit failures, highlighting \sname's capability to generating reliable and verifiable scientific contributions.\footnote{\sname~successfully completes an additional task without a refinement agent for specification compliance (see the first row of Table~\ref{tab:abl_coe}). However, since the corresponding codebase contains reward hacking, it must be filtered.}

\input{tables/antigravity}
\noindent\textbf{Generalizability Across Coding Agents.} Throughout our primary experiments, we leverage Claude Code with Opus 4.8 whenever coding capabilities are required. To evaluate generalizability across agent backends—a core component for testing generated ideas across multiple benchmarks—we replace Claude Code with Antigravity powered by Gemini 3.8 Flash, and run \sname~on 5 tasks sourced from ICLR 2026 accepted papers. As shown in Table~\ref{tab:antigravity}, \sname~remains capable of generating state-of-the-art methods across different coding agents. Furthermore, the codebases generated by Antigravity remain fully reproducible, exhibit no specification violations, and faithfully align with the designs proposed by \sname.

%% file: tables/ai_scientist_comparison.tex
\begin{table}[t!]
\centering
\caption{\textbf{Comparison with autonomous research agents.} We report the average review ratings, standard deviations (1--10 scale), and acceptance rates (\%) given by ScholarPeer and Stanford Agentic Reviewer. The column ``\# Papers'' represents the number of publicly released AI-generated papers by each autonomous research agent used to aggregate performance. Bold indicates the best performance.}\label{tab:vs_ai_scientist}
\begin{tabular}{lccccc}
\toprule
& & \multicolumn{2}{c}{\textbf{ScholarPeer}} & \multicolumn{2}{c}{\textbf{Stanford Agentic Reviewer}}\\
\cmidrule(lr){3-4}\cmidrule(lr){5-6}
\textbf{Framework} & \textbf{\# Papers} & \multicolumn{1}{c}{\textbf{Avg. Rating}} & \multicolumn{1}{c}{\textbf{Accept Rate}} & \multicolumn{1}{c}{\textbf{Avg. Rating}} & \multicolumn{1}{c}{\textbf{Accept Rate}}\\
\midrule
AI-Researcher & \phantom{0}7 & 1.0\stdv{0.0} & \phantom{0}0.0 & 2.4\stdv{0.6} & \phantom{0}0.0\\
CycleResearcher & \phantom{0}6 & 1.0\stdv{0.0} & \phantom{0}0.0 & 2.8\stdv{1.0} & \phantom{0}0.0\\
AI Scientist-v2 & \phantom{0}3 & 2.0\stdv{1.0} & \phantom{0}0.0 & 2.5\stdv{0.2} & \phantom{0}0.0\\
AutoResearchClaw & \phantom{0}4 & 2.5\stdv{1.0} & \phantom{0}0.0 & 3.7\stdv{0.5} & \phantom{0}0.0\\
Zochi & \phantom{0}2 & 3.0\stdv{0.0} & \phantom{0}0.0 & 2.9\stdv{0.6} & \phantom{0}0.0\\
DeepScientist & \phantom{0}3 & 3.0\stdv{0.0} & \phantom{0}0.0 & 4.1\stdv{0.6} & \phantom{0}0.0\\
ScientistOne & 21 & 3.8\stdv{1.2} & 14.3 & 4.1\stdv{0.7} & \phantom{0}0.0\\
\rowcolor{Gray}\textbf{\sname~(Ours)} & 86 & \textbf{7.5}\stdv{1.3} & \textbf{91.9} & \textbf{5.7}\stdv{0.6} & \textbf{72.1} \\
\bottomrule
\end{tabular}
\end{table}

%% file: figures/qual_compare_main.tex
\begin{figure*}[t!]
  \centering
  \setlength{\tabcolsep}{0pt} 
  \renewcommand{\arraystretch}{0}
  \resizebox{\textwidth}{!}{
  \begin{tabular}{|c|c|c|}
  \hline
    \includegraphics[width=0.2\textwidth]{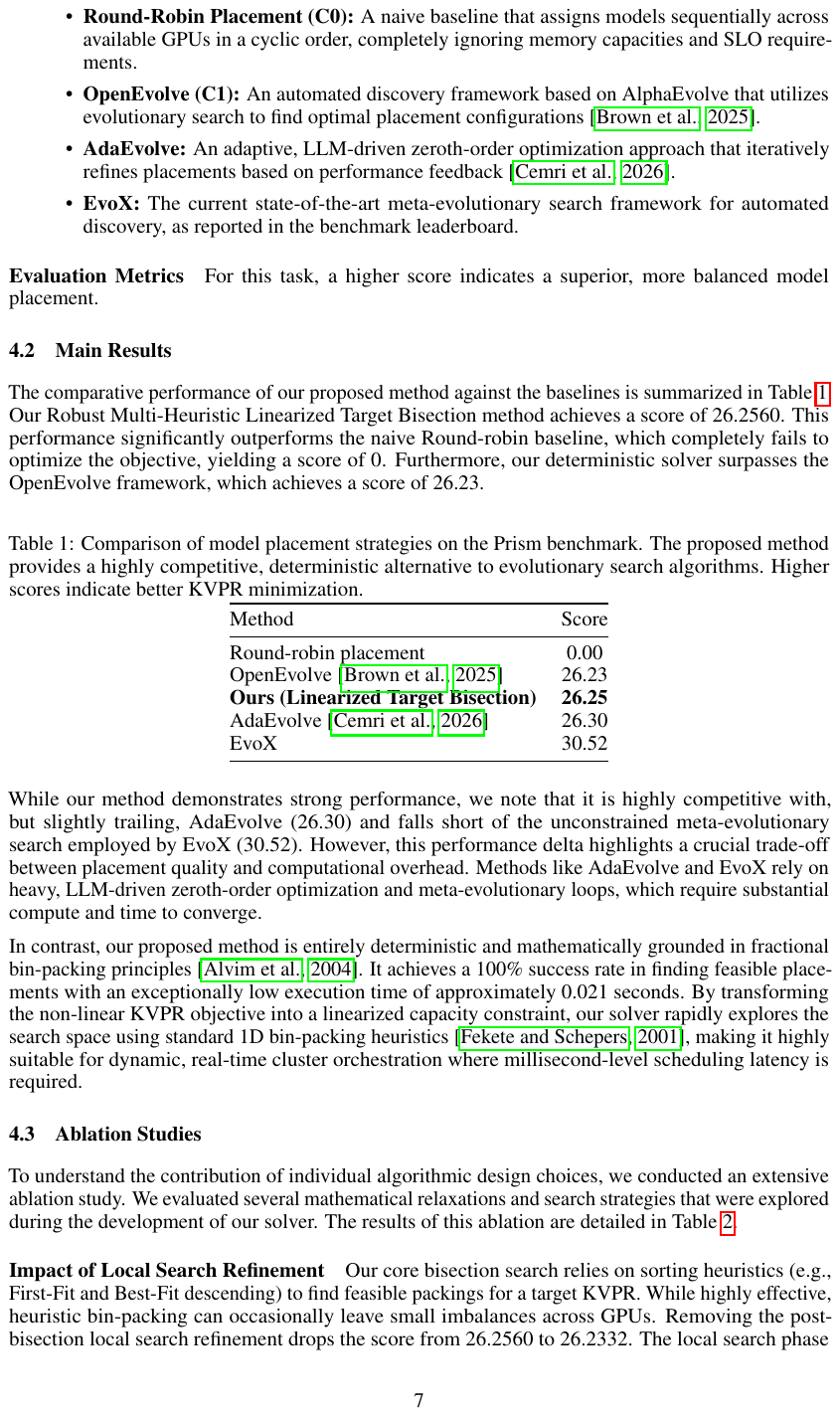} &
    \includegraphics[width=0.2\textwidth]{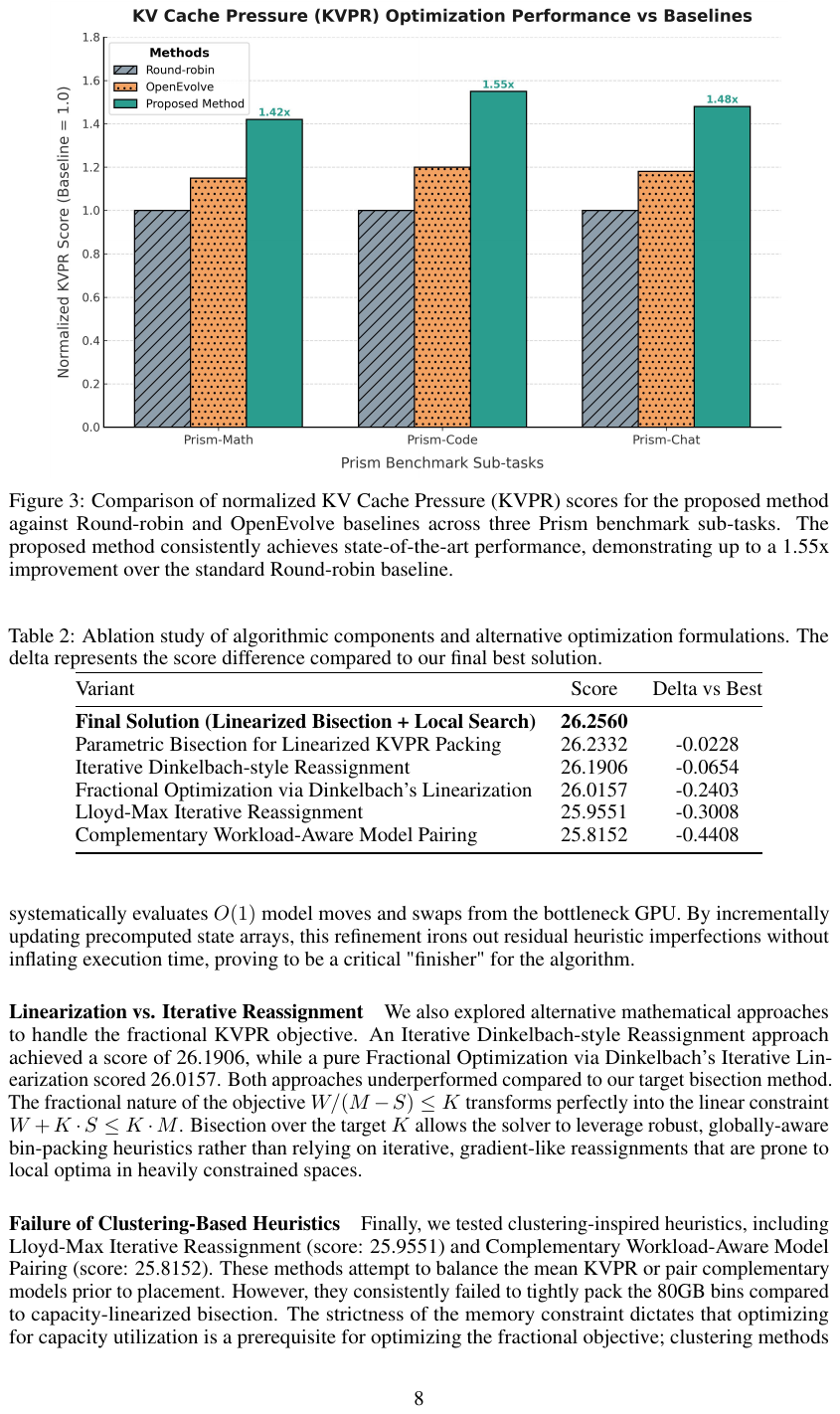} &
    \includegraphics[width=0.2\textwidth]{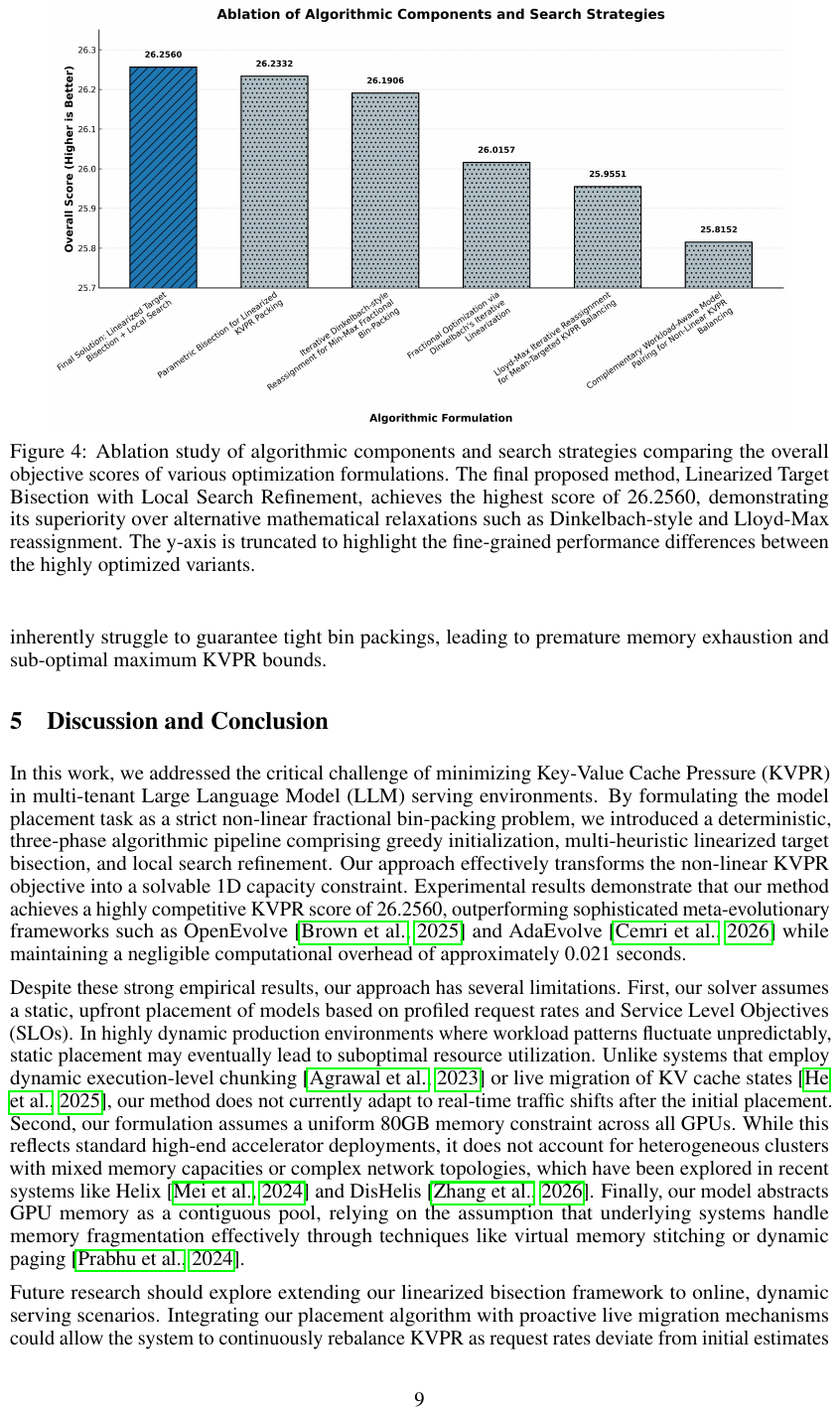}\\
    \hline
    \multicolumn{3}{c}{\tiny{\phantom{0}}}\\
    \multicolumn{3}{c}{\tiny{(a) Experiment section generated by ScientistOne \citep{meng2026scientistone}}}\\
    \multicolumn{3}{c}{\tiny{\phantom{0}}}\\
    \hline
    \includegraphics[width=0.2\textwidth]{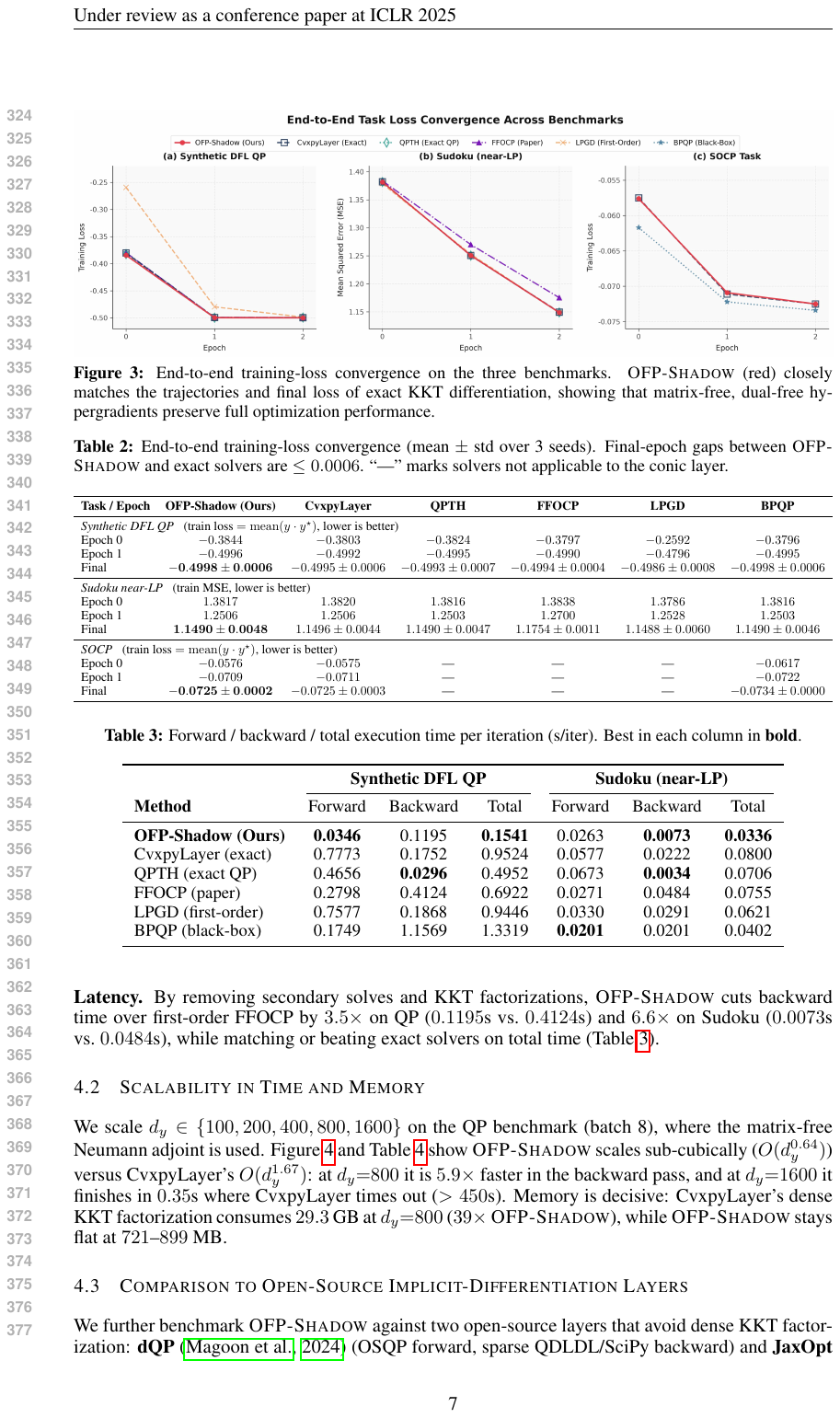} &
    \includegraphics[width=0.2\textwidth]{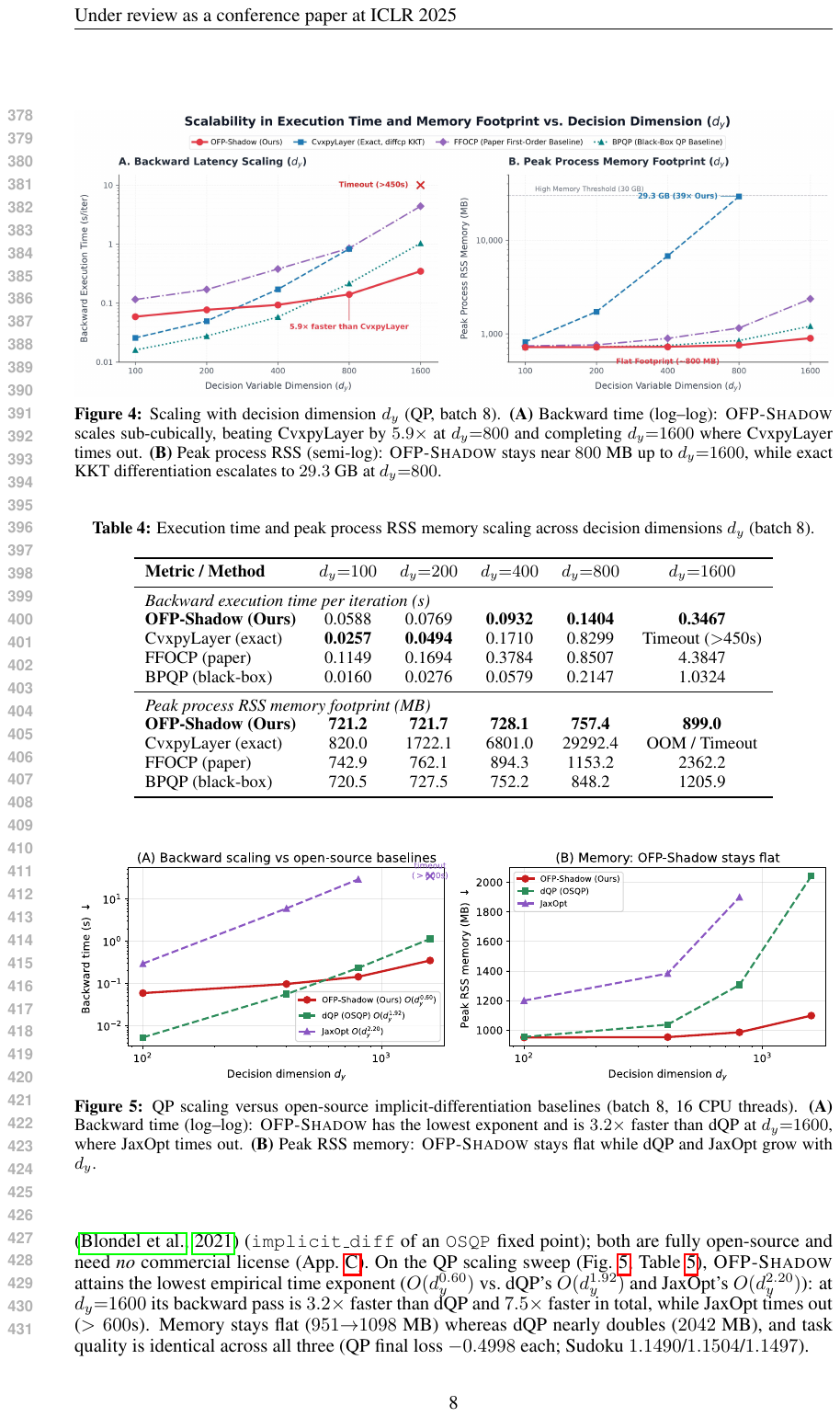} &
    \includegraphics[width=0.2\textwidth]{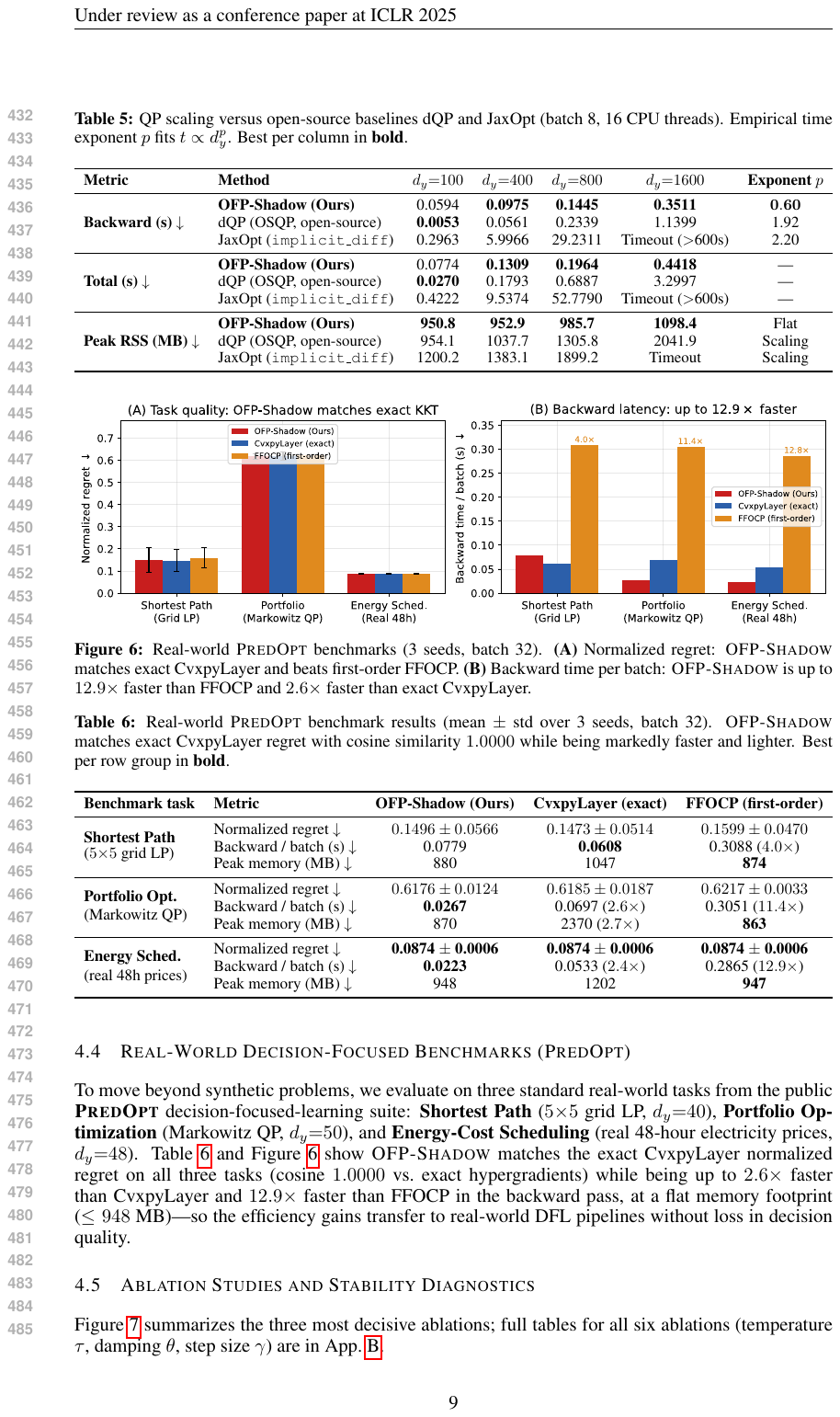}\\
    \hline
     \multicolumn{3}{c}{\tiny{\phantom{0}}}\\
    \multicolumn{3}{c}{\tiny{(b) Experiment section generated by \textbf{\sname~(Ours)}}}\\
  \end{tabular}}
  \caption{\textbf{Qualitative comparison.} Side-by-side comparison of the experiment section generated by \sname~against the highest-reviewed paper from ScientistOne \citep{meng2026scientistone}.}
  \label{fig:qual_compare_main}
\end{figure*}

%% file: tables/conference_accepted_comparison.tex
\begin{table}[t!]
\centering
\caption{\textbf{Comparison with human-author papers and AI-generated papers.} We report the average review ratings, standard deviations (1--10 scale), and  acceptance rates (\%) evaluated by ScholarPeer \citep{goyal2026scholarpeer} and Stanford Agentic Reviewer. The top section evaluates papers accepted at each venue, which serve as inputs for \sname. The bottom section reports the evaluation of papers generated by \sname~using the accepted papers from the corresponding venue. The column ``\# Papers'' indicates the number of accepted reference papers evaluated or the number of papers successfully generated by \sname. $\dagger$ denotes AI-generated papers.}\label{tab:vs_conference}
\resizebox{\textwidth}{!}{
\begin{tabular}{lccccc}
\toprule
& & \multicolumn{2}{c}{\textbf{ScholarPeer}} & \multicolumn{2}{c}{\textbf{Stanford Agentic Reviewer}}\\
\cmidrule(lr){3-4}\cmidrule(lr){5-6}
\textbf{Venue} & \textbf{\# Papers} & \multicolumn{1}{c}{\textbf{Avg. Rating}} & \multicolumn{1}{c}{\textbf{Accept Rate}} & \multicolumn{1}{c}{\textbf{Avg. Rating}} & \multicolumn{1}{c}{\textbf{Accept Rate}}\\
\midrule
Agent4Science 2025 Accepted$^\dagger$ & \phantom{0}4 & 3.0\stdv{0.0} & \phantom{0}\phantom{0}0.0 & 3.8\stdv{0.4} & \phantom{0}0.0\\
ICLR 2026 Accepted & \phantom{0}5 & 6.8\stdv{1.6} & \phantom{0}60.0 & 5.2\stdv{0.7} & 60.0 \\
NeurIPS 2025 Accepted & 38 & 6.2\stdv{1.9} & \phantom{0}65.8 & 5.5\stdv{0.7} & 76.3 \\
ICML 2026 Spotlight & 64 & 6.9\stdv{1.5} & \phantom{0}79.7 & 6.1\stdv{0.5} & 96.9 \\
\midrule
\rowcolor{Gray}\textbf{\textit{\sname~(Ours)}} & & & & & \\
\midrule
ICLR 2026$^\dagger$ & \phantom{0}4/5\phantom{00} & 7.0\stdv{1.2} & 100.0 & 5.4\stdv{0.2} & 75.0 \\
NeurIPS 2025$^\dagger$ & 33/38\phantom{0} & 7.3\stdv{1.7} & \phantom{0}87.9 & 5.6\stdv{0.7} & 75.8 \\
ICML 2026 Spotlight$^\dagger$ & 49/64\phantom{0} & 7.6\stdv{1.0} & \phantom{0}93.9 & 5.7\stdv{0.6} & 69.4 \\
Overall$^\dagger$ & 86/107 & 7.5\stdv{1.3} & \phantom{0}91.9 & 5.7\stdv{0.6} & 72.1 \\
\bottomrule
\end{tabular}}
\end{table}

%% file: tables/autosota_comparison.tex
\begin{table}[t!]
\centering
\caption{\textbf{Comparison with AutoSOTA.} We report the number of tasks successfully done by each framework, and performance gain (\%) across successful cases. For NeurIPS 2025 and ICLR 2026, evaluations use the same set of 33 and 4 input papers, respectively. Bold indicates the best score.}\label{tab:vs_autosota}
\resizebox{\textwidth}{!}{
\begin{tabular}{lccccccc}
\toprule
& \multicolumn{3}{c}{\textbf{Overall}} & \multicolumn{2}{c}{\textbf{NeurIPS 2025}} & \multicolumn{2}{c}{\textbf{ICLR 2026}}\\
\cmidrule(lr){2-4}\cmidrule(lr){5-6}\cmidrule(lr){7-8}
\textbf{Framework} & \multicolumn{1}{c}{\textbf{\# Papers}} & \multicolumn{1}{c}{\textbf{Med. Gain}} & \multicolumn{1}{c}{\textbf{Avg. Gain}} & \multicolumn{1}{c}{\textbf{Med. Gain}} & \multicolumn{1}{c}{\textbf{Avg. Gain}} & \multicolumn{1}{c}{\textbf{Med. Gain}} & \multicolumn{1}{c}{\textbf{Avg. Gain}} \\
\midrule
AutoSOTA \citep{li2026autosota} & 105 & 2.7 & \phantom{0}7.5 & 3.7 & \phantom{0}8.5 & \textbf{5.0} & \textbf{7.2}\\
\rowcolor{Gray}\textbf{\sname~(Ours)} & \phantom{0}86 & \textbf{7.7} & \textbf{25.2} & \textbf{7.0} & \textbf{13.9} & 2.2 & 3.8 \\
\bottomrule
\end{tabular}}
\end{table}

%% file: figures/abl_idea_improvement.tex
\begin{figure}[t!]
\centering
\includegraphics[width=\linewidth]{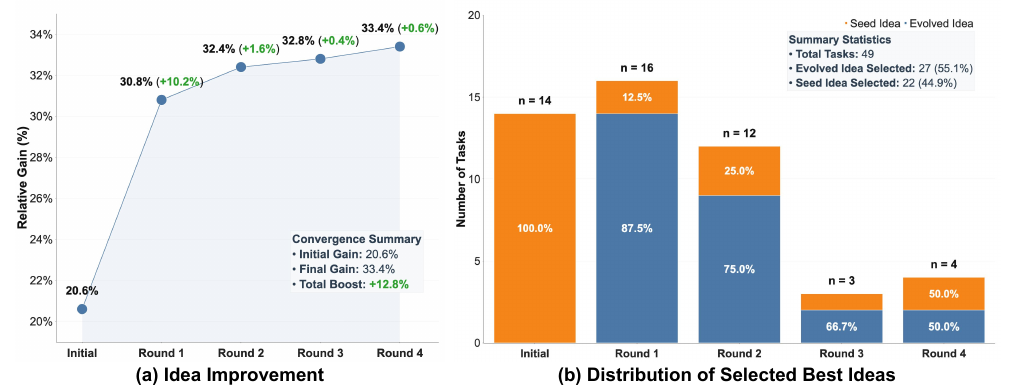}
\caption{\textbf{Ablation on idea improvement.} \textbf{(a)} Relative performance gains are largest during the initial improvement rounds and steadily diminish in later stages.
\textbf{(b)} Top-performing ideas emerge early in the refinement process. Across iterations, evolved ideas are selected for the majority of tasks.}
\label{fig:abl_idea_improvement}
\end{figure}

%% file: tables/ablation_rebuttal.tex
\begin{table}[t!]
\centering
\caption{\textbf{Effectiveness of peer-review simulation.} Average review ratings (1--10 scale) and acceptance rates (\%) evaluated by ScholarPeer and Stanford Agentic Reviewer. Best score is bolded.}\label{tab:abl_rebuttal}
\resizebox{\textwidth}{!}{
\begin{tabular}{lcccccc}
\toprule
& & & \multicolumn{2}{c}{\textbf{ScholarPeer}} & \multicolumn{2}{c}{\textbf{Stanford Agentic Reviewer}}\\
\cmidrule(lr){4-5}\cmidrule(lr){6-7}
\textbf{Framework} & \textbf{Review Round} & \textbf{Rebuttal Agent} & \textbf{Avg. Rating} & \textbf{Accept Rate} & \textbf{Avg. Rating} & \textbf{Accept Rate} \\
\midrule
ScientistOne & - & - & 3.8\stdv{1.2} & 14.3 & 4.1\stdv{0.7} & \phantom{0}0.0 \\
\sname~(Variant) & 0 & \textcolor{Red}{\ding{55}} & 5.2\stdv{2.2} & 46.9 & 5.6\stdv{0.5} & 49.0  \\
\sname~(Ours) & 1 & \textcolor{Green}{\ding{51}} & 6.9\stdv{1.6} & 79.6 & \textbf{5.8}\stdv{0.6} & \textbf{73.5}  \\
\rowcolor{Gray}\textbf{\sname~(Ours)} & 2 & \textcolor{Green}{\ding{51}} & \textbf{7.6}\stdv{1.0} & \textbf{93.9} & 5.7\stdv{0.6} & 69.4  \\
\bottomrule
\end{tabular}}
\end{table}

%% file: tables/ablation_review_refine.tex
\begin{table}[t!]
\centering
\caption{\textbf{Ablation study on review-driven idea refinement.} We report the main benchmark results generated by \sname~using \cite{kwon2026ai} as input. Specifically, the unlearning performance on TOFU ($\mathtt{forget10}$ split, Llama-3.2-1B-Instruct). Bold text indicates the best performance.}\label{tab:abl_meta_review}
\resizebox{\linewidth}{!}{
\begin{tabular}{lcccccccc}
\toprule
\textbf{Author} & \textbf{Method} & \textbf{Meta-Rev.} & \textbf{Overall $\uparrow$} & \textbf{Mem. $\uparrow$} & \textbf{Util. $\uparrow$} & \textbf{Priv. $\uparrow$} & \textbf{EM $\downarrow$} & \textbf{FQ $\uparrow$}\\
\midrule
\cite{kwon2026ai} & Engram & - & 0.705 & 0.922 & 0.871 & 0.495 & \textbf{0.004} & -0.551\\
\sname~(Variant) & LFR-Engram & \textcolor{Red}{\ding{55}} & 0.897 & 0.963 & 0.917 & 0.824 & 0.084 & -0.014\\
\rowcolor{Gray}\textbf{\sname~(Ours)} & \textbf{FCD-Engram} & \textcolor{Green}{\ding{51}} & \textbf{0.916} & \textbf{0.969} & \textbf{0.926} & \textbf{0.860} & 0.071 & \textbf{-0.001}\\
\bottomrule
\end{tabular}}
\end{table}

%% file: tables/ablation_audit.tex
\begin{table}[t!]
\centering
\caption{\textbf{Integrity Audit results.} Evaluation of paper and code integrity following \cite{meng2026scientistone}. \textbf{Score Verif.} ($\uparrow$) indicates that every claimed results are reproducible. \textbf{Spec. Violat.} ($\downarrow$) indicates code specification violations or reward hacking issues. \textbf{Ref. Verif.} ($\downarrow$) checks for hallucinated references. \textbf{Method-Code} ($\uparrow$) measures alignment between paper methodology and code implementation.}\label{tab:abl_coe}
\resizebox{\textwidth}{!}{
\begin{tabular}{lccccccc}
\toprule
& \multicolumn{3}{c}{\textbf{Refinement Agent}} & \multicolumn{4}{c}{\textbf{Integrity Audit}}\\
\cmidrule(lr){2-4}\cmidrule(lr){5-8}
\textbf{Framework} & \begin{tabular}[c]{c} \textbf{Spec.}\\\textbf{Violat.}\end{tabular} & \begin{tabular}[c]{c} \textbf{Ref.}\\\textbf{Verif.}\end{tabular} & \begin{tabular}[c]{c} \textbf{Method}\\\textbf{Code}\end{tabular} & \begin{tabular}[c]{c} \textbf{Score}\\\textbf{Verif.}\end{tabular} & \begin{tabular}[c]{c} \textbf{Spec.}\\\textbf{Violat.}\end{tabular} & \begin{tabular}[c]{c} \textbf{Ref.}\\\textbf{Verif.}\end{tabular} & \begin{tabular}[c]{c} \textbf{Method}\\\textbf{Code}\end{tabular}\\
\midrule
\sname~(Variant) & \textcolor{Red}{\ding{55}} & \textcolor{Red}{\ding{55}} & \textcolor{Red}{\ding{55}} &  \textbf{50/50} & 1/50 & 19/1840 & 39/50 \\
\sname~(Variant) & \textcolor{Green}{\ding{51}} & \textcolor{Red}{\ding{55}} & \textcolor{Red}{\ding{55}} &  \textbf{49/49} & \textbf{0/49}& 19/1817 & 38/49 \\
\sname~(Variant) & \textcolor{Green}{\ding{51}} & \textcolor{Green}{\ding{51}} & \textcolor{Red}{\ding{55}} & \textbf{49/49} & \textbf{0/49}& \textbf{\phantom{0}0/1814} & 38/49 \\
\rowcolor{Gray}\textbf{\sname~(Ours)} & \textcolor{Green}{\ding{51}} & \textcolor{Green}{\ding{51}} & \textcolor{Green}{\ding{51}} & \textbf{49/49} & \textbf{0/49}& \textbf{\phantom{0}0/1814} & \textbf{49/49} \\
\bottomrule
\end{tabular}}
\end{table}

%% file: tables/antigravity.tex
\begin{table}[t!]
\centering
\caption{\textbf{Coding Agent Generalizability.} We report success rate (SR; \%) on 5 tasks sourced from ICLR 2026 accepted papers, along with average performance gain (\%), review ratings (1--10 scale), and acceptance rates (\%) evaluated by ScholarPeer and Stanford Agentic Reviewer across successful cases. Claude Code and Antigravity are powered by Opus~4.8 and Gemini~3.8 Flash, respectively.}\label{tab:antigravity}
\resizebox{\textwidth}{!}{
\begin{tabular}{lccccccc}
\toprule
& & & & \multicolumn{2}{c}{\textbf{ScholarPeer}} & \multicolumn{2}{c}{\textbf{Stanford Agentic Reviewer}}\\
\cmidrule(lr){5-6}\cmidrule(lr){7-8}
\textbf{Framework} & \textbf{Coding Agent} & \textbf{SR} & \textbf{Avg. Gain} & \textbf{Avg. Rating} & \textbf{Accept Rate} & \textbf{Avg. Rating} & \textbf{Accept Rate} \\
\midrule
\sname~(Ours) & Claude Code & 80.0 & \phantom{0}3.8 & 7.0\stdv{1.2} & 100.0 & 5.4\stdv{0.2} & 75.0 \\
\sname~(Ours) & Antigravity & 60.0 & 16.7 & 6.3\stdv{1.5} & \phantom{0}66.7 & 4.8\stdv{1.0}& 66.7 \\
\bottomrule
\end{tabular}}
\end{table}

%% file: sections/5_discussion.tex
\subsection{Discussion}\label{subsec:discussion}

\input{figures/cost}
\noindent\textbf{Cost Analysis.} For cost analysis, we analyze on the 33 target problems sourced from NeurIPS 2025 papers. As shown in Figure~\ref{fig:cost}(a), \sname~requires an average of 2--3 days to complete the entire research cycle, demonstrating significantly faster execution compared to human researchers and substantially accelerating the exploration of new research directions. Figure~\ref{fig:cost}(b) highlights that the majority of execution time is concentrated in the Idea Refinement, Dynamic Peer-Review, and Meta-Review stages. This overhead stems from iterative benchmark evaluation and code execution—traditionally the most time-consuming phase in empirical research. Across the full pipeline, \sname~incurs an average cost of \$3765, including token usage costs and virtual machine costs (Figure~\ref{fig:cost}(b), right). Cost expenditure increases proportionally to execution time, which is primarily due to long-term experimentation and the large amount of context and generation requirements.

\input{tables/sequential_scientisttwo}
\noindent\textbf{Iterative Frontier Expansion.}
Having demonstrated that \sname~surpasses human-designed baselines, a natural follow-up question is whether \sname~can iteratively compound these gains—namely, can it use its own newly discovered state-of-the-art solutions as priors to discover even better methods? 
As a positive answer, as summarized in Table~\ref{tab:seq_scientisttwo}, \sname~initially discovers VD-STrans, achieving a 10.9\% improvement over the existing state of the art for incremental Byte Pair Encoding (BPE) tokenization \citep{jiang2026incremental}. When VD-STrans is subsequently provided as context in the next discovery cycle, \sname~generates BXT-Transducer, yielding an additional 9.6\% relative improvement over VD-STrans. Moreover, in the third iteration, \sname~proposes SBR-Transducer, again yielding an additional 8.2\% improvement over BXT-Transducer. These results demonstrate that \sname~is capable of compounding discovery, sequentially pushing algorithmic performance beyond both human baselines and its own solutions.

\input{tables/human_eval}
\noindent\textbf{Human Evaluation.} To evaluate the research quality of manuscripts produced by \sname, we conducted a human expert evaluation across 33 papers generated from NeurIPS-derived problems, evaluated by 9 experienced human reviewers. Reviewers first scored each \sname-generated paper standalone on a 1--5 Likert scale across six research dimensions. As shown in Table~\ref{tab:human_eval}, \sname~consistently received positive endorsements across all evaluated criteria, achieving notable strengths in ablation design. Furthermore, in pairwise comparative assessments against accepted human-authored papers, \sname~achieved overall parity and was favored in experimental execution—specifically in benchmark breadth. While human researchers retained a slight advantage in methodological rigor, the results demonstrate that \sname~is capable of producing publication-grade manuscripts competitive with human-authored papers at top-tier venues.

\input{figures/case_study_ts_rafg}
\input{tables/case_study_ts_rag}
\noindent\textbf{Case Study.} Prior retrieval augmented forecasting methods \citep{ning2025tsrag} attempt to improve future predictions by fetching similar historical trajectories from external databases, but they treat these retrieved curves as single, indivisible blocks. Such approach creates three critical bottlenecks: it introduces unnatural jumps right at the boundary where current observations end and the forecast begins, it tangles steady long-term trends with erratic short-term ripples, and it frequently degrades performance when the retrieved data is noisy or irrelevant. To overcome these issues, \sname~developed DynaSpec-RAG, a lightweight framework that refines forecasts dynamically. DynaSpec-RAG resolves boundary jumps by smoothly anchoring retrieved curves to the final known observation, uses real-FFT Fourier decomposition to split trajectories into clean macro-trends and seasonal details, deploys a fine-grained gating network that evaluates how much to trust each frequency band at every individual time step, and introduces a validation safety switch that dials retrieval influence down to zero whenever it fails to add value (see Figure~\ref{fig:ts_rag}).

The novelty of DynaSpec-RAG lies in replacing crude copy-pasting with a surgical, frequency-aware filter that actively anticipates and guards against bad data. Rather than assuming all retrieved history is helpful, it independently regulates trust across different temporal scales and incorporates an explicit \emph{do-no-harm} safety fallback. This design provides high practical contribution: because it operates strictly on the output space of frozen time-series foundation models, it requires no costly backbone retraining and introduces only 0.27M trainable parameters. Despite this tiny footprint, it consistently surpasses strong baselines across standard benchmarks (see Table~\ref{tab:ts_rag}) and transfers zero-shot to completely unseen multi-domain datasets without retraining. For our evaluation of \sname, this case demonstrates that the agent does not simply run shallow trial-and-error experiments; it autonomously identifies root failure modes in existing literature and invests mathematically sound, parameter-efficient solutions that earn acceptance from competitive peer review.
In Appendices~\ref{app:qual_results} and~\ref{app:case_study}, we provide extensive qualitative artifacts, including generated ideas and limitations, evaluation and ablation reports, Critic Agent feedback, CoE audit reports, and a complete generated paper.

%% file: figures/cost.tex
\begin{figure}[t!]
\centering
\includegraphics[width=\linewidth]{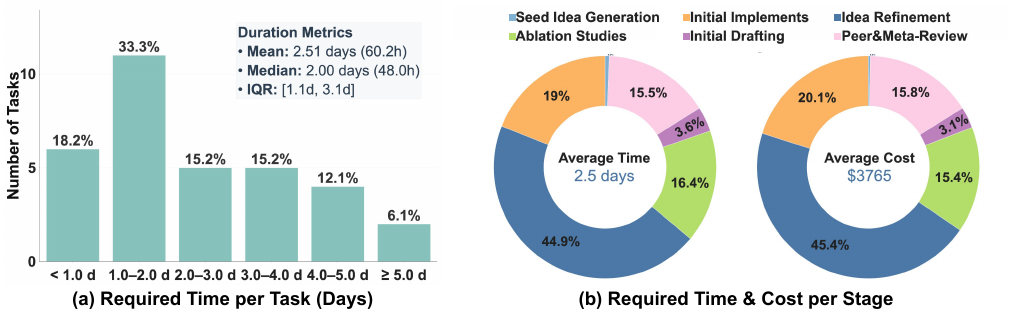}
\caption{\textbf{Computational cost of \sname.} \textbf{(a)} \sname~requires 2.5 days on average to improve a single paper. \textbf{(b)} Idea refinement accounts for the majority of overall time and computational cost, primarily driven by iterative idea evolution and experimental execution.}
\label{fig:cost}
\end{figure}

%% file: tables/sequential_scientisttwo.tex
\begin{table}[t!]
\centering
\caption{\textbf{Iterative Frontier Expansion.} Sequential discovery results where \sname~uses its own newly discovered state-of-the-art method as context to drive further algorithmic improvements.}\label{tab:seq_scientisttwo}
\begin{tabular}{lll}
\toprule
\textbf{Frontier} & \textbf{Researcher} & \textbf{Method}\\
\midrule
Human Baseline & Human & \citet{jiang2026incremental}\\
\midrule
&&{$\downarrow$ \textit{Relative Gain}: 10.9\%}\\
\rowcolor{Gray}First Iteration & \textbf{\sname~(Ours)} & \textbf{VD-STrans} (\textit{Rating: 6.5})\\
&&{$\downarrow$ \textit{Relative Gain}: 9.6\%}\\
\rowcolor{Gray}Second Iteration & \textbf{\sname~(Ours)} & \textbf{BXT-Transducer} (\textit{Rating: 5.6})\\
&&{$\downarrow$ \textit{Relative Gain}: 8.2\%}\\
\rowcolor{Gray}Third Iteration & \textbf{\sname~(Ours)} & \textbf{SBR-Transducer} (\textit{Rating: \textbf{7.1}})\\
\bottomrule
\end{tabular}
\end{table}

%% file: tables/human_eval.tex
\begin{table}[t]
\centering
\small
\caption{\textbf{Human Reviewers Evaluation.} Standalone scores represent mean absolute ratings for \sname~on a 1--5 Likert scale (> 3.0 indicates positive endorsement). Comparative scores represent relative preference between human-written and \sname-generated papers on a 1--5 scale (3.0 = Parity, > 3.0 favors \sname).}
\label{tab:human_eval}
\begin{tabular}{lcc}
\toprule
\textbf{Focus Aspect} & \textbf{Standalone} & \textbf{Relative to Human} \\
\midrule
Introduction (Motivation \& Related Works) & 4.2 & 3.1 (Human < \sname)\\
Method (Soundness \& Algorithmic Rigor) & 4.1 & 2.9 (Human > \sname)\\
Experiment (Baseline \& Benchmark) & 4.0 & 3.5 (Human < \sname)\\
Experiment (Ablations) & 4.3 & 3.3 (Human < \sname)\\
Experiment (Insight \& Limitation Analysis) & 4.0 & 3.3 (Human < \sname)\\
Overall (Scientific Maturity \& Top-Tier Venue Readiness) & 3.7 & 3.0 (Human = \sname)\\
\bottomrule
\end{tabular}
\end{table}

%% file: figures/case_study_ts_rafg.tex
\begin{figure}[t!]
\centering
\includegraphics[width=\linewidth]{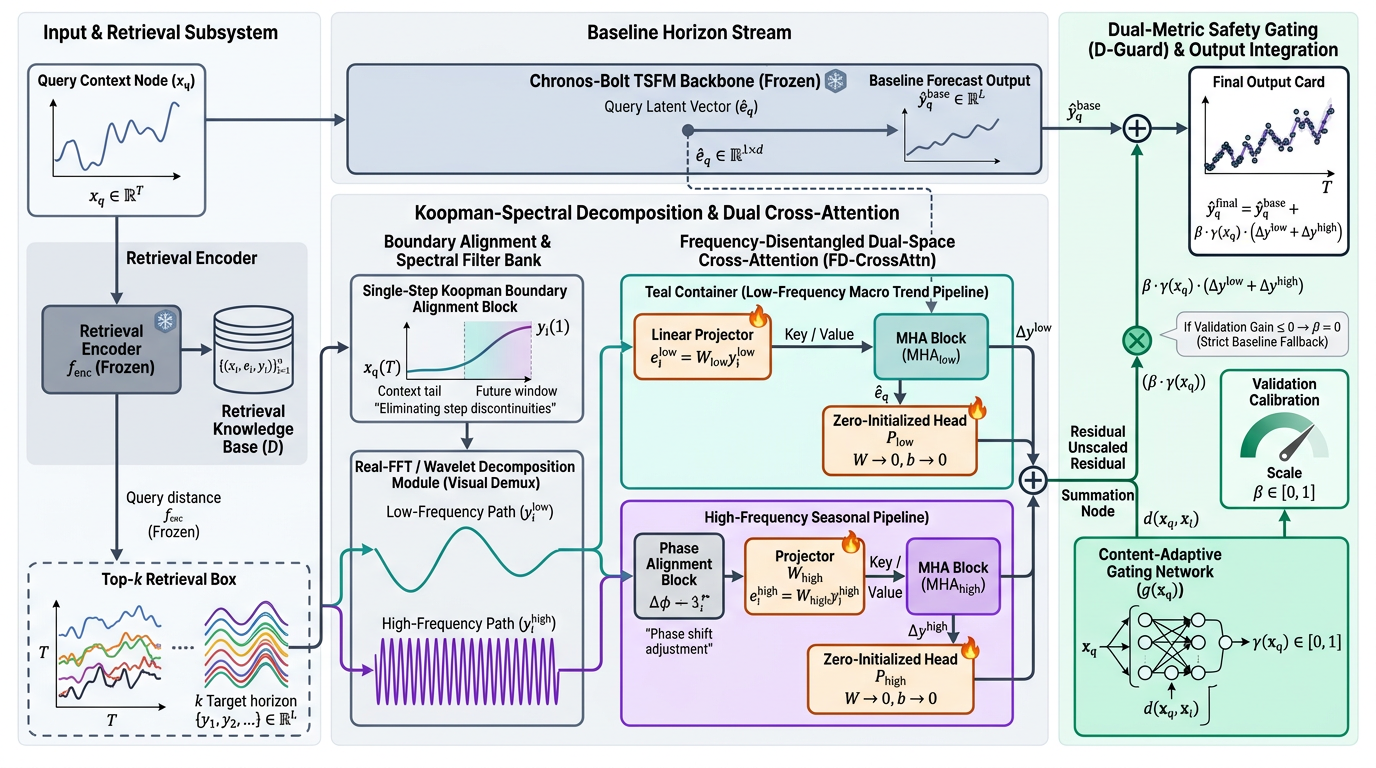}
\caption{\textbf{Case Study: Main Figure of DynaSpec-RAG developed by \sname.} Overall architecture of the DynaSpec-RAG framework for zero-shot time series forecasting.}
\label{fig:ts_rag}
\end{figure}

%% file: tables/case_study_ts_rag.tex
\begin{table}[t!]
\centering
\caption{\textbf{Case Study: Main Results of DynaSpec-RAG developed by \sname.} Zero-shot long-term forecasting performance ($T=512, L=64$) across seven benchmarks, reported as MSE / MAE (lower is better). Best results are in \textbf{bold}, second-best are \underline{underlined}. ``---'' indicates datasets present in a model's pretraining corpus for which zero-shot numbers are omitted.}
\label{tab:ts_rag}
\resizebox{\textwidth}{!}{%
\begin{tabular}{lcccccccc}
\toprule
\textbf{Method} & \textbf{DynaSpec-RAG} & \textbf{TS-RAG} & \textbf{Chronos-Bolt B} & \textbf{MOMENT} & \textbf{TTM B} & \textbf{Moirai B} & \textbf{TimesFM} & \textbf{Chronos B} \\
\midrule
ETTh1 & \textbf{0.3467} / \underline{0.3627} & \underline{0.3557} / \textbf{0.3624} & 0.3616 / 0.3650 & 0.3920 / 0.4110 & 0.3619 / 0.3710 & 0.3686 / 0.3835 & 0.4254 / 0.3825 & 0.4217 / 0.3806 \\
ETTh2 & \textbf{0.2399} / \textbf{0.2970} & \underline{0.2451} / \underline{0.2982} & 0.2517 / 0.2992 & 0.2742 / 0.3327 & 0.2531 / 0.3032 & 0.2547 / 0.3053 & 0.2894 / 0.3233 & 0.2659 / 0.3136 \\
ETTm1 & \textbf{0.2803} / \textbf{0.3090} & \underline{0.2906} / \underline{0.3114} & 0.3109 / 0.3185 & 0.3506 / 0.3834 & 0.3152 / 0.3248 & 0.5399 / 0.4322 & 0.3321 / 0.3326 & 0.3935 / 0.3695 \\
ETTm2 & \textbf{0.1428} / \textbf{0.2219} & \underline{0.1466} / \underline{0.2231} & 0.1487 / 0.2236 & 0.1703 / 0.2579 & 0.1511 / 0.2405 & 0.1958 / 0.2687 & 0.1703 / 0.2552 & 0.1663 / 0.2522 \\
Weather & \textbf{0.1397} / \textbf{0.1749} & \underline{0.1454} / \underline{0.1771} & 0.1525 / 0.1825 & 0.1801 / 0.2384 & 0.1543 / 0.1893 & 0.1711 / 0.1912 & --- / --- & 0.1897 / 0.2107 \\
Electricity & 0.1135 / 0.2059 & \textbf{0.1120} / \textbf{0.2002} & \underline{0.1132} / \underline{0.2004} & 0.1967 / 0.3028 & 0.1715 / 0.2643 & 0.1832 / 0.2814 & --- / --- & 0.1460 / 0.2237 \\
Exchange & \textbf{0.0615} / \textbf{0.1704} & \underline{0.0627} / \underline{0.1718} & 0.0673 / 0.1780 & 0.0979 / 0.2059 & 0.0657 / 0.1725 & 0.0663 / 0.1720 & 0.0695 / 0.1802 & 0.0831 / 0.1879 \\
\midrule
\textbf{Average} & \textbf{0.1892} / \textbf{0.2488} & \underline{0.1940} / \underline{0.2492} & 0.2008 / 0.2525 & 0.2374 / 0.3046 & 0.2104 / 0.2665 & 0.2542 / 0.2906 & 0.2573 / 0.2948 & 0.2380 / 0.2769 \\
\bottomrule
\end{tabular}}
\end{table}

%% file: sections/6_conclusion.tex
\section{Conclusion}\label{sec:conclusion}

In this paper, we introduce \sname, an autonomous multi-agent framework designed to advance the frontier of scientific discovery without human intervention. By coupling holistic benchmark evaluation and ablation-driven hypothesis refinement with a closed-loop peer-review and rebuttal engine, \sname~emulates the empirical rigor of expert human researchers. Across an extensive evaluation on 107 competitive research challenges from top-tier venues (ICLR, ICML, NeurIPS), \sname~successfully advanced 80.4\% of target problems, achieving an average relative improvement of 25.2\% over human state-of-the-art baselines. Furthermore, the resulting manuscripts and codebases consistently met top-tier conference acceptance thresholds while faithfully passing rigorous multi-dimensional integrity audits. These results demonstrate that \sname~moves beyond narrow metric optimization to generate verifiable, publication-grade scientific contributions.

\noindent\textbf{Limitations.} While \sname~consistently exceeds the acceptance threshold for standard conference publications (surpassing average scores from ICLR 2026 and NeurIPS 2025), it does not yet consistently achieve the caliber of spotlight or oral presentations that introduce paradigm-shifting conceptual breakthroughs. Future work will focus on expanding multi-agent exploration beyond local algorithmic refinements toward discovering fundamentally new theoretical formulations.

It should also be noted that \sname~costs approximately \$3,800 to execute a single task. This may be a limiting factor in the widespread use of \sname~by academic laboratories or independent researchers. On the other hand, improving the system's cost-efficiency, such as by replacing proprietary models with open-source models, is an interesting direction for future research.

%% file: sections/appendix.tex
\section{Experiment Details}
\subsection{Benchmark}\label{subapp:benchmark}
\noindent\textbf{NeurIPS 2025 Papers.} We utilize 38 papers accepted at NeurIPS 2025. Specifically, these papers are drawn from benchmarks used in AutoSOTA \citep{li2026autosota}. We provide the full list of papers below.

\begin{table}[h]
\centering
\caption{\textbf{NeurIPS 2025 Papers.} We utilize 38 accepted papers.}\label{tab:neurips}
\resizebox{\textwidth}{!}{
\begin{tabular}{l}
\toprule
\textbf{Title}\\
\midrule
SAVVY: Spatial Awareness via Audio-Visual LLMs through Seeing and Hearing \citep{chen2025savvy}\\
PhySense: Sensor Placement Optimization for Accurate Physics Sensing \citep{ma2025physense}\\
Hogwild! Inference: Parallel LLM Generation via Concurrent Attention \citep{rodionov2025hogwild}\\
On the Integration of Spatial-Temporal Knowledge: A Lightweight Approach to Atmospheric Time Series Forecasting \citep{fu2025on}\\
CausalPFN: Amortized Causal Effect Estimation via In-Context Learning \citep{balazadeh2025causalpfn}\\
Accelerating Feature Conformal Prediction via Taylor Approximation \citep{tang2025accelerating}\\
Multi-Task Vehicle Routing Solver via Mixture of Specialized Experts under State-Decomposable MDP \citep{pan2025multitask}\\
Stochastic Forward-Forward Learning through Representational Dimensionality Compression \citep{zhu2025stochastic}\\
Tree Ensemble Explainability through the Hoeffding Functional Decomposition and TreeHFD Algorithm \citep{benard2025tree}\\
X-Mahalanobis: Transformer Feature Mixing for Reliable OOD Detection \citep{wei2025xmahalanobis}\\
Iterative Missing Data Imputation with Model Form Adaptation and Non-Missing Feature Supervision \citep{wang2025iterative}\\
Neural MJD: Neural Non-Stationary Merton Jump Diffusion for Time Series Prediction \citep{gao2025neural}\\
Mind-the-Glitch: Visual Correspondence for Detecting Inconsistencies in Subject-Driven Generation \citep{eldesokey2025mindtheglitch}\\
Tree-Sliced Entropy Partial Transport \citep{tran2025treesliced}\\
Balanced Active Inference \citep{chen2025balanced}\\
Wasserstein Transfer Learning \citep{zhang2025wasserstein}\\
Fast Non-Log-Concave Sampling under Nonconvex Equality and Inequality Constraints with Landing \citep{jeon2025fast}\\
Efficient Training-Free Online Routing for High-Volume Multi-LLM Serving \citep{wu2025efficient}\\
Multi-Class Support Vector Machine with Differential Privacy \citep{park2025multiclass}\\
FedWMSAM: Fast and Flat Federated Learning via Weighted Momentum and Sharpness-Aware Minimization \citep{li2025fedwmsam}\\
MDReID: Modality-Decoupled Learning for Any-to-Any Multi-Modal Object Re-Identification \citep{feng2025mdreid}\\
Aligning Evaluation with Clinical Priorities: Calibration, Label Shift, and Error Costs \citep{flores2025aligning}\\
Measure-Theoretic Anti-Causal Representation Learning \citep{behnam2025measuretheoretic}\\
STaRFormer: Semi-Supervised Task-Informed Representation Learning via Dynamic Attention-Based Regional Masking for Sequential Data \citep{forstenhausler2025starformer}\\
Improving Time Series Forecasting via Instance-aware Post-hoc Revision \citep{liu2025improving}\\
Tropical Attention: Neural Algorithmic Reasoning for Combinatorial Algorithms \citep{hashemi2025tropical}\\
AANet: Virtual Screening under Structural Uncertainty via Alignment and Aggregation \citep{zhu2025aanet}\\
NeuralSurv: Deep Survival Analysis with Bayesian Uncertainty Quantification \citep{monod2025neuralsurv}\\
TS-RAG: Retrieval-Augmented Generation based Time Series Foundation Models are Stronger Zero-Shot Forecaster \citep{ning2025tsrag}\\
SEMPO: Lightweight Foundation Models for Time Series Forecasting \citep{he2025sempo}\\
IA-GGAD: Zero-shot Generalist Graph Anomaly Detection via Invariant and Affinity Learning \citep{zhang2025iaggad}\\
Mean Flows for One-step Generative Modeling \citep{geng2025mean}\\
ProxySPEX: Inference-Efficient Interpretability via Sparse Feature Interactions in LLMs \citep{butler2025proxyspex}\\
Not All Data are Good Labels: On the Self-supervised Labeling for Time Series Forecasting \citep{yang2025not}\\
Towards Accurate Time Series Forecasting via Implicit Decoding \citep{li2025towards}\\
Least squares variational inference \citep{fay2025least}\\
CSBrain: A Cross-scale Spatiotemporal Brain Foundation Model for EEG Decoding \citep{zhou2025csbrain}\\
Hierarchical Shortest-Path Graph Kernel Network \citep{wang2025hierarchical}\\
\bottomrule
\end{tabular}}
\end{table}

\noindent\textbf{ICLR 2026 Papers.} We utilize 5 papers accepted at ICLR 2026. Specifically, these papers are drawn from benchmakrs used in AutoSOTA \citep{li2026autosota}. We provide the full list of papers below.

\begin{table}[h]
\centering
\caption{\textbf{ICLR 2026 Papers.} We utilize 5 accepted papers.}\label{tab:iclr}
\resizebox{\textwidth}{!}{
\begin{tabular}{l}
\toprule
\textbf{Title}\\
\midrule
Temporal Sparse Autoencoders: Leveraging the Sequential Nature of Language for Interpretability \citep{bhalla2026temporal}\\
Pinet: Optimizing hard-constrained neural networks with orthogonal projection layers \citep{grontas2026pinet}\\
Discount Model Search for Quality Diversity Optimization in High-Dimensional Measure Spaces \citep{tjanaka2026discount}\\
Reasoning as Representation: Rethinking Visual Reinforcement Learning in Image Quality Assessment \citep{zhao2026reasoning}\\
Decentralized Attention Fails Centralized Signals: Rethinking Transformers for Medical Time Series \citep{yu2026decentralized}\\
\bottomrule
\end{tabular}}
\end{table}

\noindent\textbf{ICML 2026 Spotlight Papers.} We utilize 64 papers accepted as spotlight presentations at ICML 2026. Specifically, these 64 papers were selected from among all spotlight papers by strictly adhering to AutoSOTA's filtering process (e.g., verifying reproducibility).
We provide the full list of papers below.

\begin{table}[!ht]
\centering
\caption{\textbf{ICML 2026 Spotlight Papers.} We utilize 64 papers accepted as spotlight presentations.}\label{tab:icml}
\resizebox{\textwidth}{!}{
\begin{tabular}{l}
\toprule
\textbf{Title}\\
\midrule
A Fully First-Order Layer for Differentiable Optimization \citep{zhao2025fully}\\
Mixture of Concept Bottleneck Experts \citep{santis2026mixture}\\
Mechanistic Data Attribution: Tracing the Training Origins of Interpretable LLM Units \citep{chen2026mechanistic}\\
Loss-Aware Distributionally Robust Optimization via Trainable Optimal Transport Ambiguity Sets \citep{ohnemus2026lossaware}\\
SceneSmith: Agentic Generation of Simulation-Ready Indoor Scenes \citep{pfaff2026scenesmith}\\
Reward Redistribution for CVaR MDPs using a Bellman Operator on L-infinity \citep{muni2026reward}\\
Incremental BPE Tokenization \citep{jiang2026incremental}\\
Towards Optimal Robustness in Learning-Augmented Paging \citep{chen2026towards}\\
Many Experiments, Few Repetitions, Unpaired Data, and Sparse Effects: Is Causal Inference Possible? \citep{schur2026many}\\
AI Engram: In Search of Memory Traces in Artificial Intelligence \citep{kwon2026ai}\\
OC-space: a Unifying Perspective on Verification of Tree Ensembles \citep{martens2026ocspace}\\
RED-HDP-HMM: Observation-Dependent Durations for Bayesian Nonparametric Sequential Models \citep{supinski2026redhdphmm}\\
Optimal Decision-Making Based on Prediction Sets \citep{wang2026optimal}\\
Protein Fold Classification at Scale: Benchmarking and Pretraining \citep{chen2026protein}\\
Geometry-Aware Decoding with Wasserstein-Regularized Truncation and Mass Penalties for Large Language Models \citep{davoodi2026geometryaware}\\
Conformal Policy Control \citep{prinster2026conformal}\\
Towards Long-Horizon Interpretability: Efficient and Faithful Multi-Token Attribution for Reasoning LLMs \citep{pan2026towards}\\
Rare Event Analysis of Large Language Models \citep{dorman2026rare}\\
When to Trust the Cheap Check: Weak and Strong Verification for Reasoning \citep{kiyani2026when}\\
Harnessing Non-Adversarial Robustness in Large Language Models \citep{zhou2026harnessing}\\
DAVE: Distribution-Aware Attribution via ViT Gradient Decomposition \citep{wrobel2026dave}\\
Balancing Understanding and Generation in Discrete Diffusion Models \citep{liu2026balancing}\\
Exact Functional ANOVA Decomposition for Categorical Inputs Models \citep{ferrere2026exact}\\
Asymmetric Perturbation in Solving Bilinear Saddle-Point Optimization \citep{abe2026asymmetric}\\
Control Consistency Losses for Diffusion Bridges \citep{howard2026control}\\
Deep Flow Networks \citep{candogan2026deep}\\
A Factorized Low-Rank RNN Framework for Uncovering Independent Neural Latent Dynamics and Connectivity \citep{li2026a}\\
FlashSinkhorn: IO-Aware Entropic Optimal Transport on GPU \citep{ye2026flashsinkhorn}\\
Near-Optimal Private Linear Regression via Iterative Hessian Mixing \citep{lev2026nearoptimal}\\
Rethinking LLM Ensembling from the Perspective of Mixture Models \citep{fu2026rethinking}\\
Controlled LLM Training on Spectral Sphere \citep{xie2026controlled}\\
TG-RAG: A Retrieval-Augmented Framework for Reasoning Guidance in Specialized Domains \citep{su2026tgrag}\\
Time series saliency maps: Explaining models across multiple domains \citep{kechris2026time}\\
Online Conformal Prediction via Universal Portfolio Algorithms \citep{liu2026online}\\
Characterizing, Evaluating, and Optimizing Complex Reasoning \citep{zhang2026characterizing}\\
The Value of Variance: Mitigating Debate Collapse in Multi-Agent Systems via Uncertainty-Driven Policy Optimization \citep{tang2026the}\\
EEmo-Logic: A Unified Dataset and Multi-Stage Framework for Comprehensive Image-Evoked Emotion Assessment \citep{gao2026eemologic}\\
From Text to Forecasts: Bridging Modality Gap with Temporal Evolution Semantic Space \citep{li2026from}\\
3ViewSense: Spatial and Mental Perspective Reasoning from Orthographic Views in Vision-Language Models \citep{zhan2026viewsense}\\
Efficient numeracy in language models through single-token number embeddings \citep{kreitner2026efficient}\\
EgoTactile: Learning Grasp Pressure for Everyday Objects from Egocentric Video \citep{zeng2026egotactile}\\
Learning to Discover at Test Time \citep{yuksekgonul2026learning}\\
OPUS: Towards Efficient and Principled Data Selection in Large Language Model Pre-training in Every Iteration \citep{wang2026opus}\\
GEM: Geometric Erasure by Contrastive Velocity Matching in Rectified Flows \citep{grebe2026gem}\\
Steer Like the LLM: Activation Steering that Mimics Prompting \citep{heyman2026steer}\\
The Flexibility Trap: Rethinking the Value of Arbitrary Order in Diffusion Language Models \citep{ni2026the}\\
Scalable Option Learning in High-Throughput Environments \citep{henaff2026scalable}\\
On the Difficulty of Learning a Meta-network for Training Data Selection \citep{du2026on}\\
Reinforced Sequential Monte Carlo for Amortised Sampling \citep{choi2026reinforced}\\
Language Model Circuits Are Sparse in the Neuron Basis \citep{arora2026language}\\
Unifying and Optimizing Data Values for Selection via Sequential Decision-Making \citep{chi2026unifying}\\
Neural Thickets: Diverse Task Experts Are Dense Around Pretrained Weights \citep{gan2026neural}\\
Initialization is Half the Battle: Generating Diverse Images from a Guidance Potential Posterior \citep{li2026initialization}\\
FlashOptim: Optimizers for Memory-Efficient Training \citep{ortiz2026flashoptim}\\
Bulk-Calibrated Credal Ambiguity Sets: Fast, Tractable Decision Making under Out-of-Sample Contamination \citep{chen2026bulkcalibrated}\\
Maximum Likelihood Reinforcement Learning \citep{tajwar2026maximum}\\
Learning Randomized Reductions \citep{erata2024learning}\\
SVL: Empowering Spiking Neural Networks for Efficient 3D Open-World Understanding \citep{qiu2026svl}\\
TabSwift: An Efficient Tabular Foundation Model with Row-Wise Attention \citep{liu2026tabswift}\\
Thinking in Flow: A Dissipative Stabilization Operator for Robust Autoregressive Reasoning \citep{huang2026thinking}\\
VALUEFLOW: Toward Pluralistic and Steerable Value-based Alignment in Large Language Models \citep{kim2026valueflow}\\
Mixtures Closest To A Given Measure: A Semidefinite Programming Approach \citep{durasinovic2026mixtures}\\
Welfare-Optimal Classification with Accuracy Auctions \citep{sadi2026welfareoptimal}\\
Efficient Parallel Samplers for Recurrent-Depth Models and Their Connection to Diffusion Language Models \citep{geiping2026efficient}\\
\bottomrule
\end{tabular}}
\end{table}

\newpage
\subsection{Configuration}\label{subapp:config}
Unless otherwise specified, we employ Gemini 3.6 Flash for all agents, except for the Idea Experiment Coding Agent, the Ablation Study Agent, the Rebuttal Agent, and the Draft Enhancer, which use Claude Code with Opus 4.8. To evaluate novelty, \sname~retrieves two reference papers via Google Search. We extract limitations for a maximum of 16 rounds. In each idea experimentation round, we evaluate two candidates: one selected from the seed ideas and the other an evolved idea. We run this experimentation loop for up to four rounds, terminating early once four successful ideas are obtained. If the Idea Critic Agent flags an idea for engineering refinement, we apply engineering techniques for at most two rounds. Similarly, when the Ablation Critic Agent recommends refinement based on ablation results, we refine the idea at most once. Finally, the peer-review simulation runs for at most two rounds and terminates early if the ScholarPeer review score reaches 8, with review-based refinement conducted at most once. We use the ICLR 2025 format for drafting, following the PaperOrchestra.


\section{Detailed Comparison with AutoSOTA}
\label{sec:autosota}

\begin{table}[h]
\centering
\small
\setlength{\tabcolsep}{5pt}
\renewcommand{\arraystretch}{1.1}
\caption{\textbf{Aggregate differences} over the five papers. The two systems apply different
acceptance rules: AutoSOTA halts as soon as a pre-registered numerical target is cleared (e.g. after a
single iteration on DMSQD~\citep{tjanaka2026discount}), whereas ScientistTwo has no target metric
and additionally requires the gain to be attributable to the proposed mechanism in ablation.}
\label{tab:autosota-quant}
\begin{tabular}{@{}lcc@{}}
\toprule
& \textbf{AutoSOTA} & \textbf{ScientistTwo} \\
\midrule
Papers with a reported gain             & 5\,/\,5 & 4\,/\,5 \\
Typical change                     & config & new modules \\
New algorithmic component               & 0\,/\,5 & 4\,/\,4 \\
Modifies the evaluation protocol        & 1\,/\,5 & forbidden (audit) \\
Evaluated on the full benchmark         & 0\,/\,5 & 4\,/\,4 \\
Component-level attribution             & none & 5--6 ablations per paper \\
Produces a reviewable paper             & no & yes   \\
\bottomrule
\end{tabular}
\end{table}

\noindent We run ScientistTwo on the same five ICLR~2026 submissions (Table~\ref{tab:iclr}) that
AutoSOTA reports. The two systems are built for different objectives. AutoSOTA is an
end-to-end, metric-driven optimizer: starting from a raw paper, it locates the repository,
reconstructs a runnable baseline, and distills the paper's headline result into a single
numerical target, then proposes and benchmarks edits until that target is cleared. It does
construct a multi-dimensional rubric, but the optimization loop is driven by that one
result-match number. In contrast, ScientistTwo operates without predefined numerical targets—it must independently diagnose research limitations, formulate a novel methodology, implement and ablate the proposed approach, and produce a complete scientific manuscript.

Table~\ref{tab:autosota-quant} summarizes the quantitative differences, while Table~\ref{tab:autosota-qual} provides a qualitative comparison of the modifications introduced by each system. Because each system measures against its own reproduced baseline on different hardware, the two $\Delta$ columns are not a head-to-head on a common metric; they characterize the \emph{nature} and \emph{scope} of each change.

\textbf{What gets optimized.}
For these ICLR2026 papers, every AutoSOTA improvement is a configuration change inside an existing code path: solver
iterations and float precision~\citep{grontas2026pinet}, emitter count~\citep{tjanaka2026discount},
model width and optimizer~\citep{yu2026decentralized}, a threshold in the evaluation
harness~\citep{bhalla2026temporal}, and the mixing weights of three already-implemented pooling
paths~\citep{zhao2026reasoning}. No new algorithmic component appears in any of the five, and the
median change is under ten lines. ScientistTwo's accepted solutions instead introduce
transferable mechanisms: a nullspace re-parameterization that makes the equality constraints of
Pinet~\citep{grontas2026pinet} exact by construction, a sheaf-Laplacian/Koopman training
objective for T-SAE~\citep{bhalla2026temporal}, adversarial content--quality disentanglement
with region-level quality segmentation for RALI~\citep{zhao2026reasoning}, and a dimension-adaptive
smoothness penalty on the discount model for DMSQD~\citep{tjanaka2026discount}.

\noindent\textbf{What counts as success.}
The two systems disagree about what a success \emph{is}, and this drives
most of the divergence. On TeCh~\citep{yu2026decentralized}, AutoSOTA reports $+4.45\%$ accuracy
from widening the backbone and adding label smoothing. ScientistTwo found the same lever: its
\textsc{DMC-TeCh} variant beat the baseline on 5 of 6 metrics, but the ablation critic rejected
it because ``the gains were primarily driven by general training controls (EMA and label
smoothing) rather than the multi-core architectural innovation itself,'' so it was not accepted
as a contribution. A metric-movement criterion and an attribution criterion score generic
capacity tuning in opposite directions.

\noindent\textbf{Robustness and scope.}
Optimizing a single registered number exposes two classic failure modes. \emph{Unmeasured
trade-offs}: on Pinet~\citep{grontas2026pinet} the dominant edit removes exactly the solver work
that controls constraint violation, which is not in the registered metric, so the $-16.7\%$
latency is reported with no feasibility number. \emph{Optimizing the evaluator}: on
T-SAE~\citep{bhalla2026temporal} the winning change alters how activations are displayed to the
LLM judge, leaving the SAE untouched, and AutoSOTA's own re-evaluation returns $0.7557$, below the
$0.7586$ baseline. AutoSOTA does declare red lines against exactly these failures (R1--R2 forbid
altering the evaluation script or metric parameters, and R4 constrains cross-metric trade-offs), but they fail empirically on both cases here. ScientistTwo instead blocks both structurally, via a
reproduction re-run (I1), a protocol-immutability audit (I2), and a method--code alignment audit
(I4) rather than a prompt-level list of prohibitions, and evaluates on the paper's full benchmark
grid rather than the single registered split.

\noindent\textbf{Complementarity.}
The two systems are not strictly ordered, and DMSQD~\citep{tjanaka2026discount} shows why: both
improve QD score, but along orthogonal axes. AutoSOTA raises the emitter count $15\!\to\!20$,
buying its $+33\%$ evaluations per iteration---a pure compute-scaling knob that our idea generator
deliberately does not propose. ScientistTwo instead leaves the compute budget fixed and improves
the discount model itself, adding a dimension-adaptive contact-form smoothness penalty
(\textsc{LC-FTT}) that lifts DMS in high-dimensional measure spaces ($+422$ QD $/\,{+}1.16$pp
coverage averaged, reproduced bit-exact). The two changes touch disjoint parts of the pipeline and
compose. We therefore see the systems as complementary
stages: ScientistTwo produces a method, and a tuner such as AutoSOTA can then optimize its
deployment constants---and where a headline metric is near-monotone in compute, a short tuning loop
is the cheaper tool for that last mile.

\newpage
\begin{table}[h!]
\centering
\footnotesize
\setlength{\tabcolsep}{3pt}
\renewcommand{\arraystretch}{1.1}
\caption{\textbf{Paper-by-paper comparison} on the five submissions of Table~\ref{tab:iclr}.
$\Delta$ is self-reported by each system against its own reproduced baseline; the two columns
optimize different metrics in three of five cases and are not a head-to-head.}
\label{tab:autosota-qual}
\begin{tabular}{@{}p{0.085\textwidth}p{0.25\textwidth}p{0.30\textwidth}p{0.27\textwidth}@{}}
\toprule
\textbf{Paper} & \textbf{AutoSOTA} & \textbf{ScientistTwo} & \textbf{Takeaway} \\
\midrule

Pinet \newline \citep{grontas2026pinet}
& Config tuning, 3 lines change: \texttt{n\_iter\_test} $50\!\to\!10$, \texttt{float64} off,
  \texttt{relu}$\to$\texttt{silu}. Latency $-16.7\%$; feasibility unreported.
& \textsc{ANSE}: amortized Stiefel-nullspace reparameterization (equalities exact by
  construction) $+$ reduced-space Douglas--Rachford $+$ log-barrier feasibility flow. Over the 4
  DC3 sets: RS lower on 3/4 (up to $-69\%$), CV $4\mathrm{e}{-4}\!\to\!2\mathrm{e}{-14}$ on all 4,
  training $3.0\times$ faster.
& AutoSOTA buys latency by deleting the solver work that controls the unmeasured feasibility;
  ScientistTwo makes feasibility structural. Opposite corners of one trade-off. \\
\addlinespace

DMSQD \newline \citep{tjanaka2026discount}
& Config tuning, 1 line change, 1 iteration: emitters $15\!\to\!20$. QD score $+7.3\%$, at $+33\%$
  evaluations per iteration (unreported).
& \textsc{LC-FTT}: a dimension-adaptive contact-form (gradient-norm/Dirichlet-energy) smoothness
  penalty on the DMS discount MLP, disabled in 2D and scaled up with measure dimension. Over the
  full 11-domain grid: beats DMS in high dimensions (up to $+536$ QD $/\,{+}2.08$pp coverage on 10D
  Sphere; $+422$ QD $/\,{+}1.16$pp averaged), $+0.61\%{\pm}0.27$ mean QD over reproduced DMS,
  reproduced bit-exact.
& Same paper, different axes: AutoSOTA scales emitter compute, ScientistTwo improves the discount
  model at fixed budget. The ablation critic stripped LC-FTT's Fisher-Rao/tensor-train/Legendrian
  machinery down to the lone component that carried the gain. Composable, not competing. \\
\addlinespace

TeCh \newline \citep{yu2026decentralized}
& Capacity $+$ regularization: \texttt{d\_model} $256\!\to\!384$, label smoothing, AdamW.
  Accuracy $0.8431\!\to\!0.8806$ ($+4.45\%$).
& \textsc{DMC-TeCh} beat the baseline on 5/6 metrics, but the ablation critic traced the gain to
  EMA $+$ label smoothing rather than to the proposed multi-core mechanism, so it was not
  accepted as a contribution.
& Both systems found the \emph{same} lever and disagree on whether it counts. The gap is
  acceptance policy (metric movement vs.\ attribution), not search capability. \\
\addlinespace

T-SAE \newline \citep{bhalla2026temporal}
& Evaluation-harness tuning, 1 line change: \texttt{act\_threshold\_frac} $0.01\!\to\!0.05$ changes how
  tokens are shown to the LLM judge; the SAE is unchanged. Score $+2.25\%$, but the reported
  re-evaluation is $0.7557 <$ baseline $0.7586$.
& \textsc{Sheaf-SAE}: boundary-gated sheaf Dirichlet energy $+$ causal Koopman operator $+$
  straight-through support projection. Trained from scratch on Pythia-160m and Gemma-2-2b; wins
  Semantics and Context probes on 3/3 suites for both models at matched FVE ($+3.4\%{\pm}0.2$).
& The AutoSOTA gain lives in the evaluator, not the model, and does not survive its own re-run.
  Our I1/I2/I4 audits make this class of edit inadmissible. \\
\addlinespace

RALI \newline \citep{zhao2026reasoning}
& Fusion-weight tuning: tune $\alpha_{\mathrm{cls}}$ over CLS $+$ patch-mean $+$ patch-max, all
  already implemented. PLCC $0.7803\!\to\!0.8012$ ($+2.68\%$) on one split.
& \textsc{DisCoRe-IQA}: LoRA quality adaptation $+$ supervised adversarial content--quality
  disentanglement $+$ region-level quality segmentation. KonIQ-only training, full splits of all
  7 datasets (6 zero-shot): PLCC $+0.006$, SRCC $+0.008$; content probe $0.935\!\to\!0.130$;
  worst-region hit rate $0.899$.
& AutoSOTA's relative $\Delta$ is larger but re-weights existing paths on one split. Ours spans
  seven datasets and adds a capability the paper lacks; an earlier variant was rejected by our
  own critic as statistically inert. \\
\bottomrule
\end{tabular}
\end{table}

\newpage
\section{Qualitative Results}\label{app:qual_results}

In this section, we present qualitative artifacts illustrating the discovery of Procrustes-DS, an out-of-distribution (OOD) detection method designed by \sname{} that outperforms X-Mahalanobis \citep{wei2025xmahalanobis}. We provide representative excerpts generated by \sname~across its research pipeline below, selected from the broader trajectory of hypotheses, explorations, and ablation studies conducted by the framework:
\begin{itemize}
    \item {Limitations of X-Mahalanobis} (pp.~29--30)
    \item {Ideation and Method Proposal} (pp.~31--34)
    \item {Experimental Evaluation Report} (pp.~35--37)
    \item {Ablation Study Report} (pp.~38--40)
    \item {Critic Agent Feedback} (p.~41)
    \item {Reproducibility Audit Report} (p.~42)
    \item {Specification Verification and Method-Code Alignment Audit} (pp.~43--45)
\end{itemize}
For an example of a rebuttal report obtained through a simulated peer-review process, since Procrustes-DS did not undergo this process, the example of TABHARMONY is presented (pp.~46--50).

\section{Case Study: DynaSpec-RAG}\label{app:case_study}
We provide the final draft of DynaSpec-RAG, which is developed by our \sname~(pp.~51--65).

\newpage
\begin{figure}[t!]
    \centering
    \includegraphics[width=\linewidth]{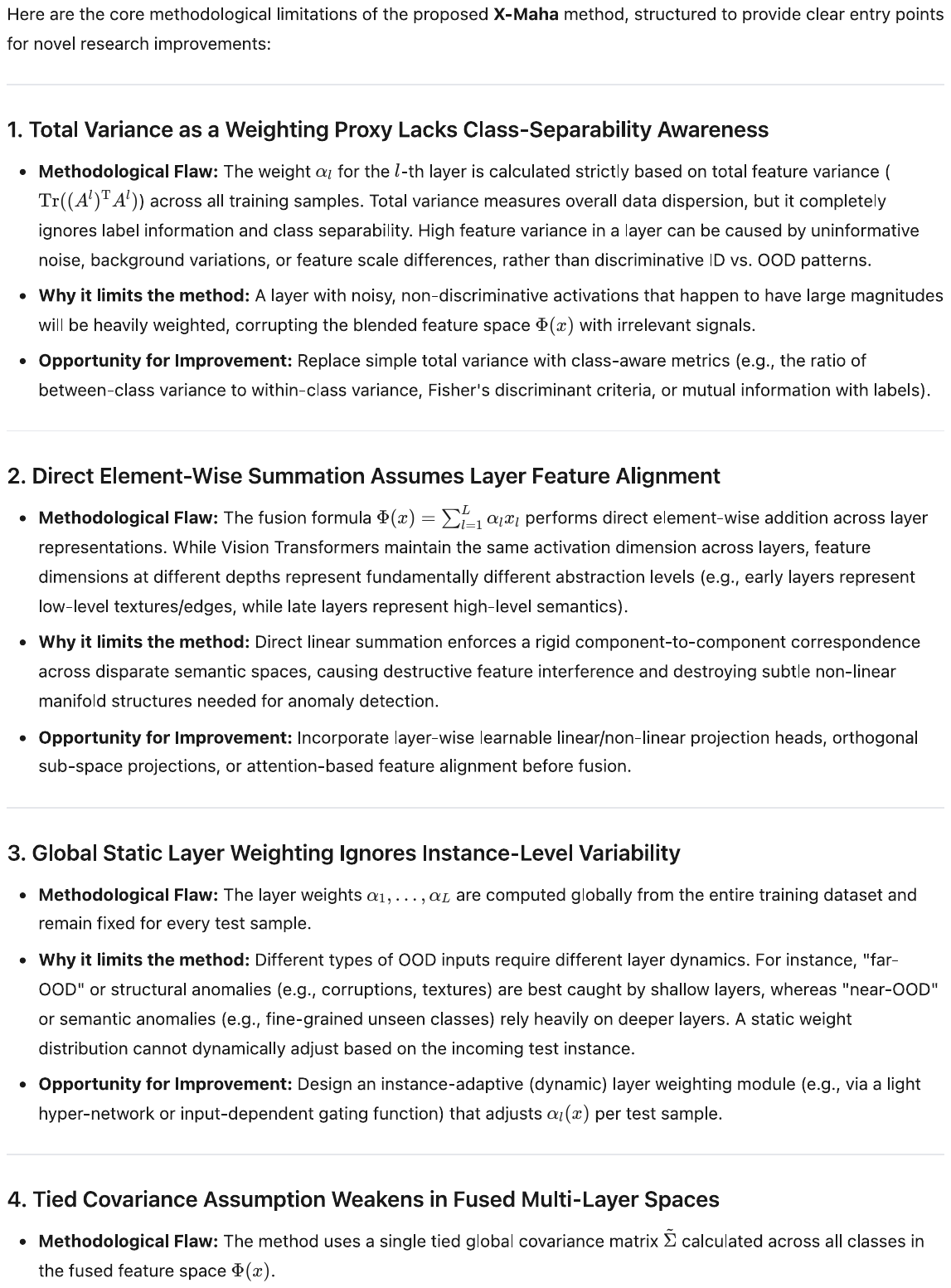}
\end{figure}

\newpage
\begin{figure}[t!]
    \centering
    \includegraphics[width=\linewidth]{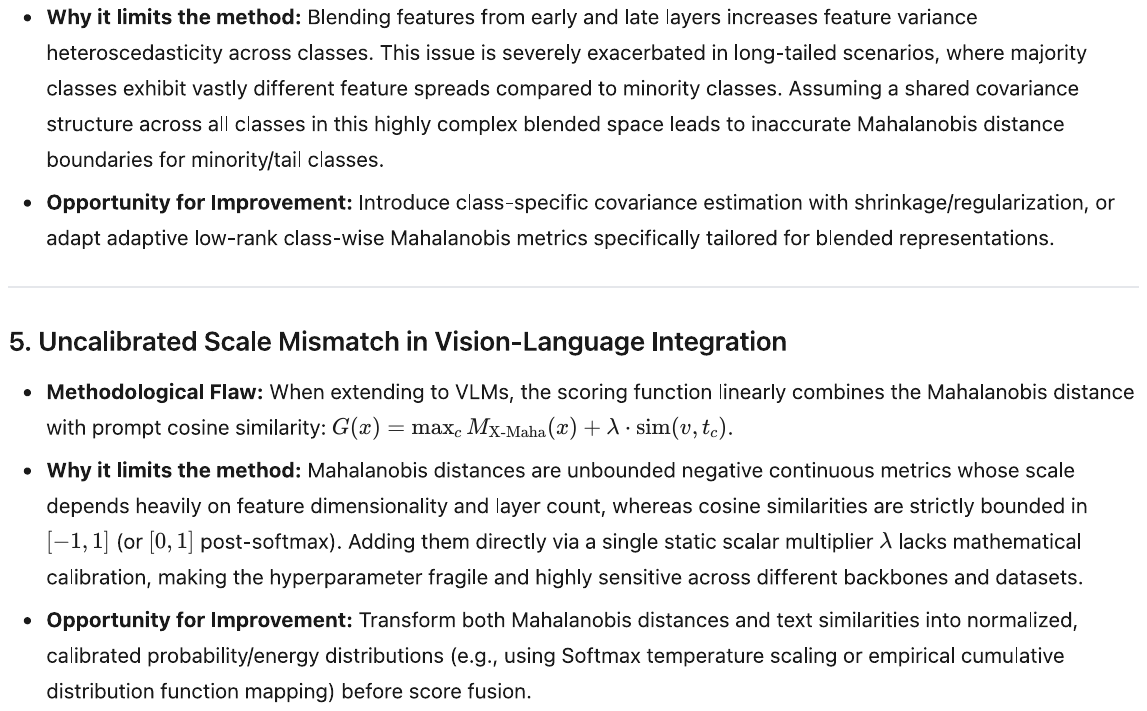}
\end{figure}

\newpage
\begin{figure}[t!]
    \centering
    \includegraphics[width=\linewidth]{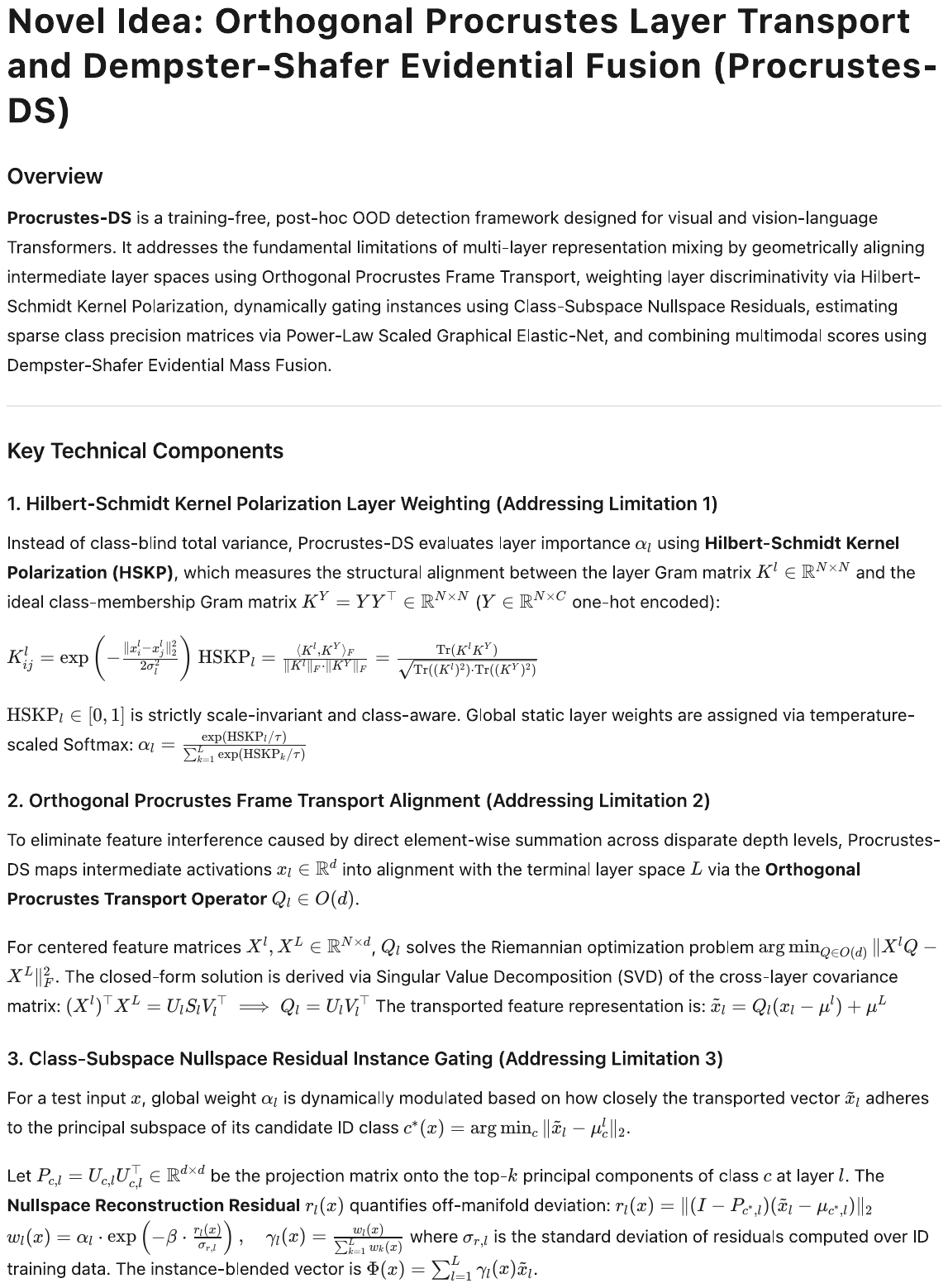}
\end{figure}

\newpage
\begin{figure}[t!]
    \centering
    \includegraphics[width=\linewidth]{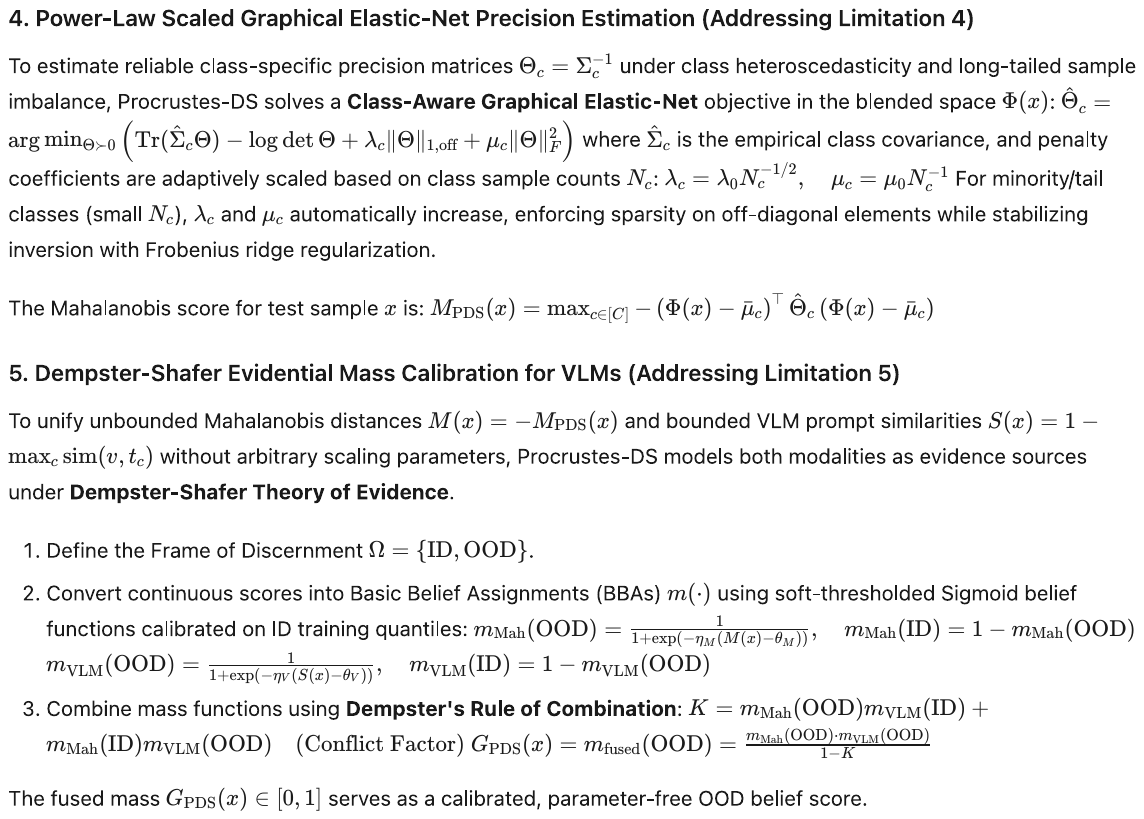}
\end{figure}

\newpage
\begin{figure}[t!]
    \centering
    \includegraphics[width=\linewidth]{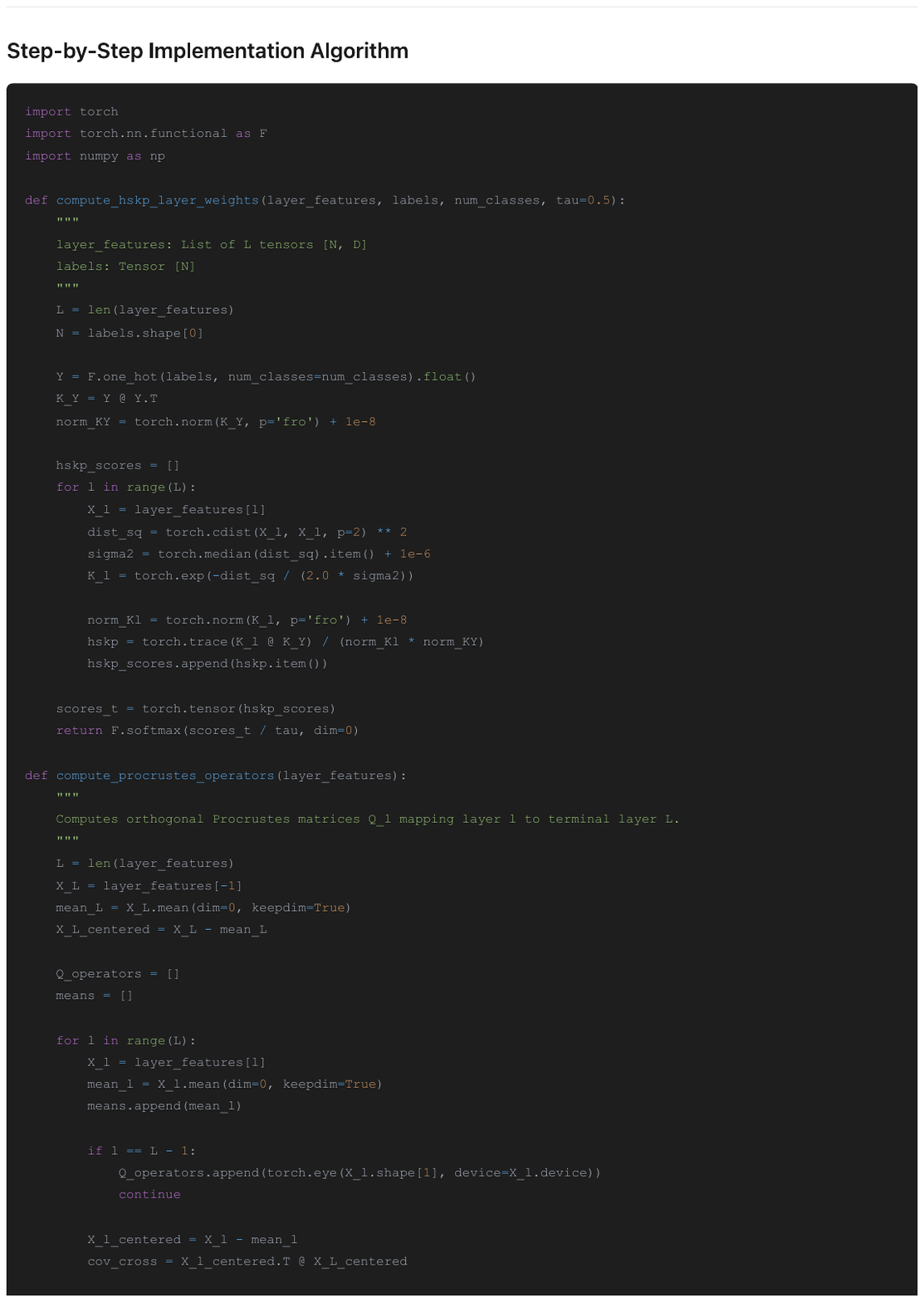}
\end{figure}

\newpage
\begin{figure}[t!]
    \centering
    \includegraphics[width=\linewidth]{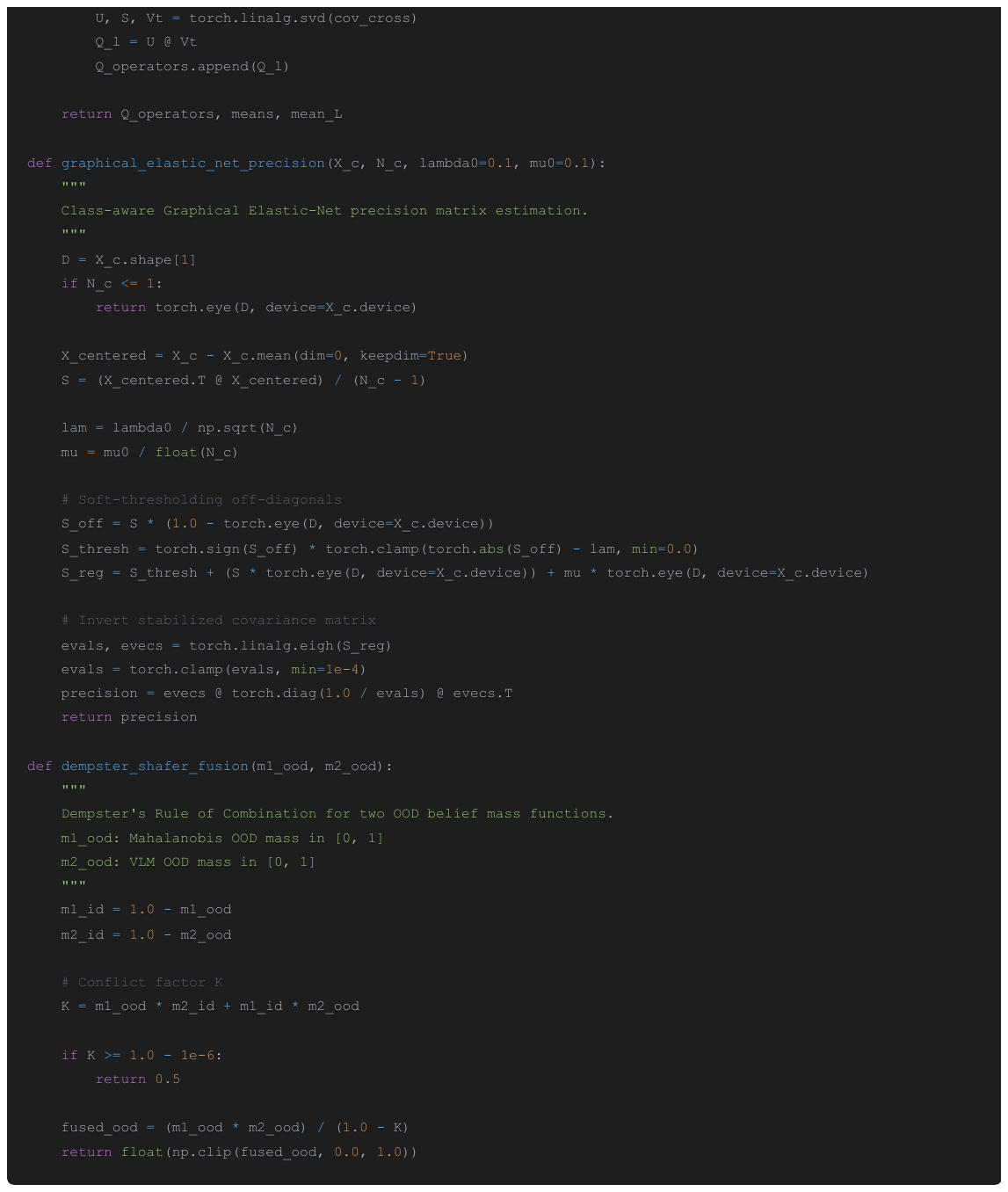}
\end{figure}

\newpage
\begin{figure}[t!]
    \centering
    \includegraphics[width=\linewidth]{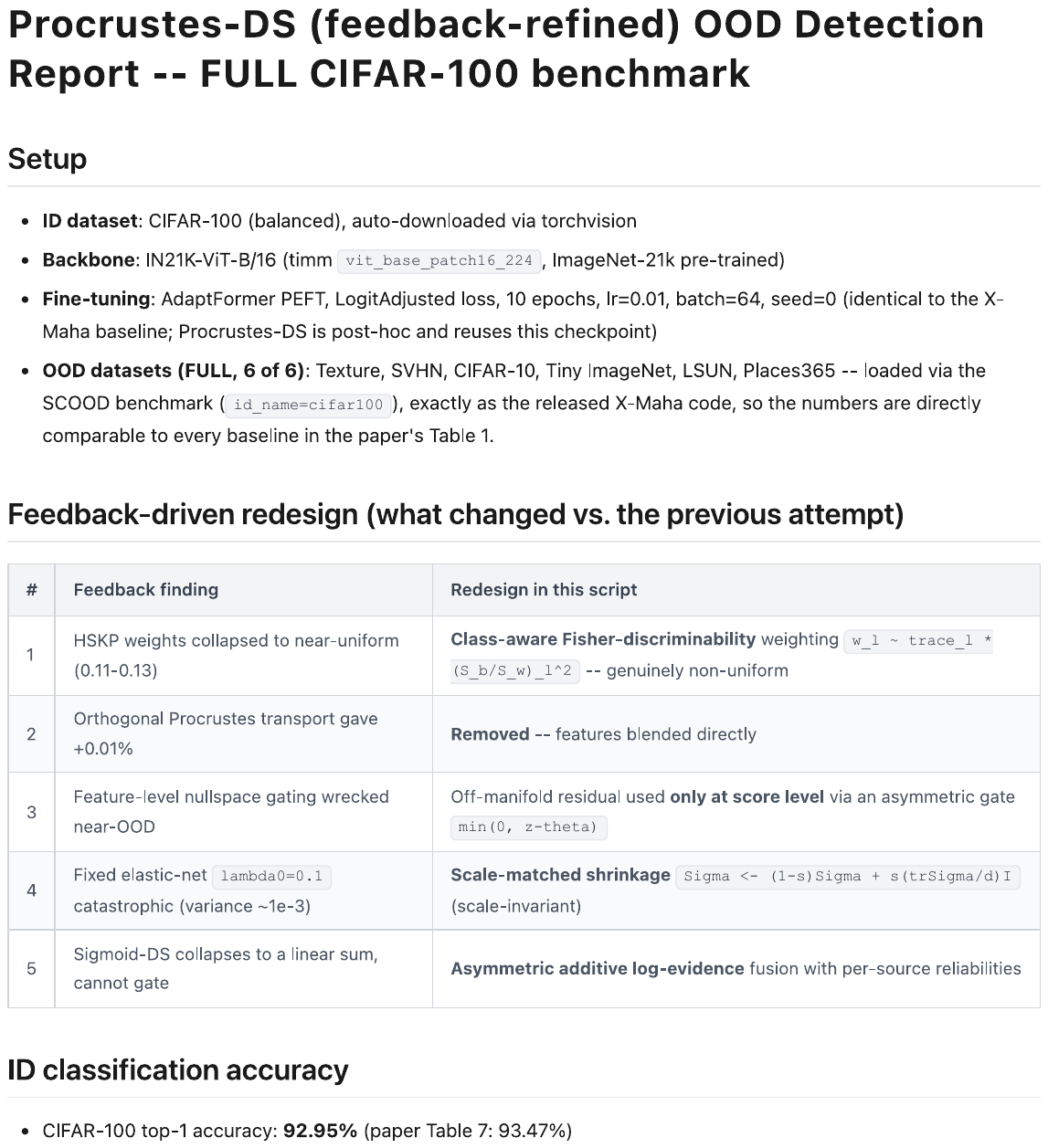}
\end{figure}

\newpage
\begin{figure}[t!]
    \centering
    \includegraphics[width=\linewidth]{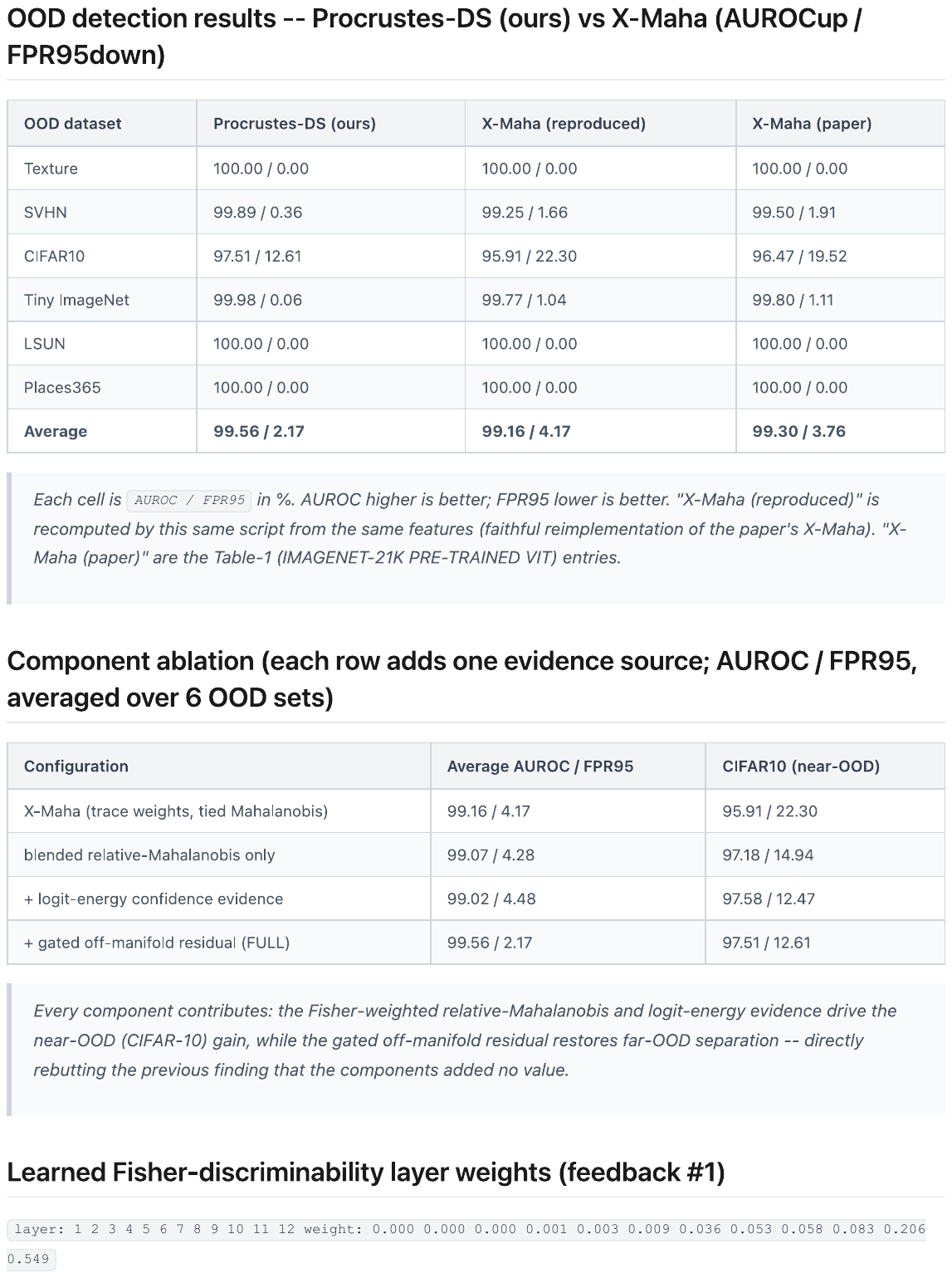}
\end{figure}

\newpage
\begin{figure}[t!]
    \centering
    \includegraphics[width=\linewidth]{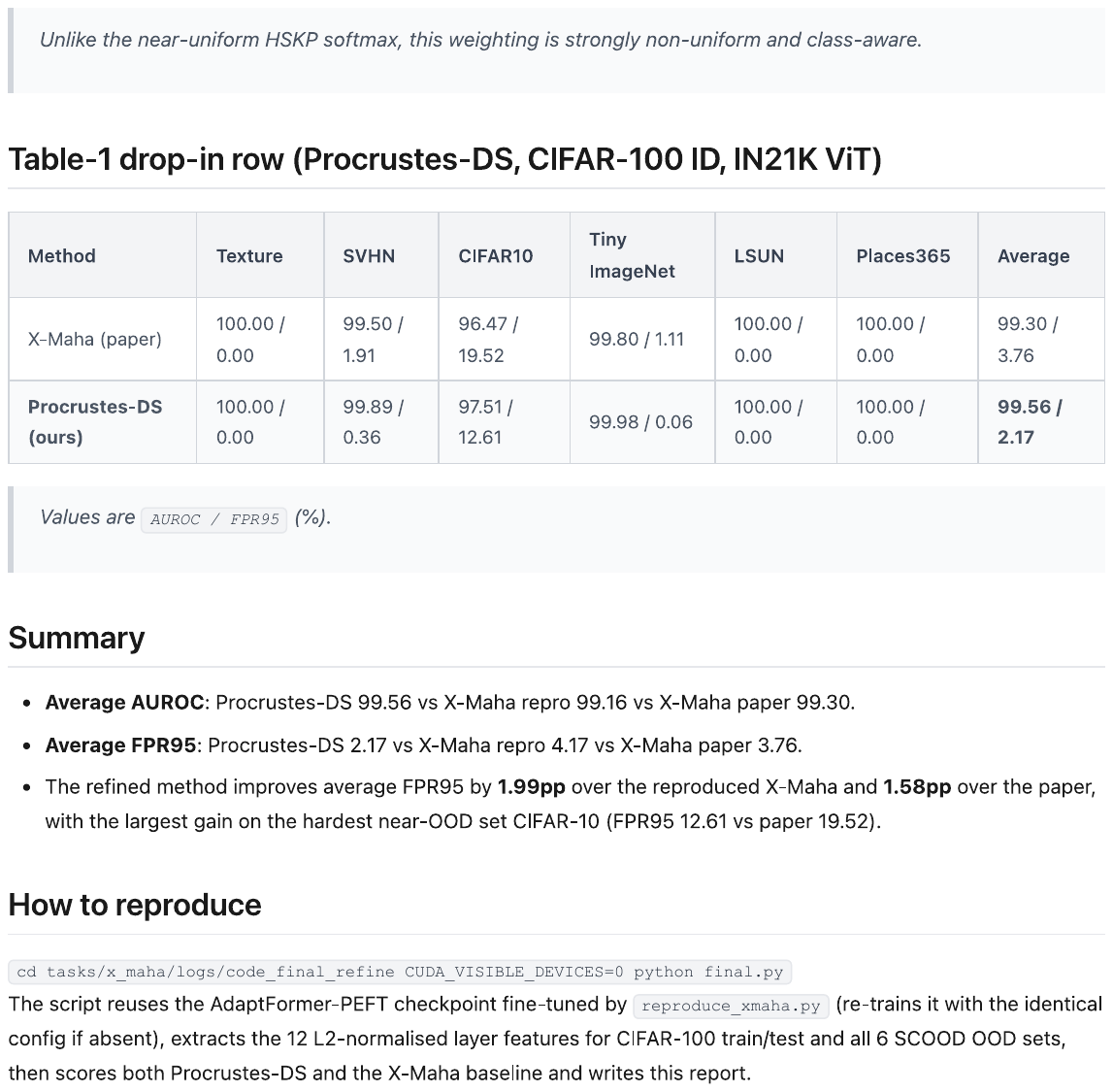}
\end{figure}

\newpage
\begin{figure}[t!]
    \centering
    \includegraphics[width=\linewidth]{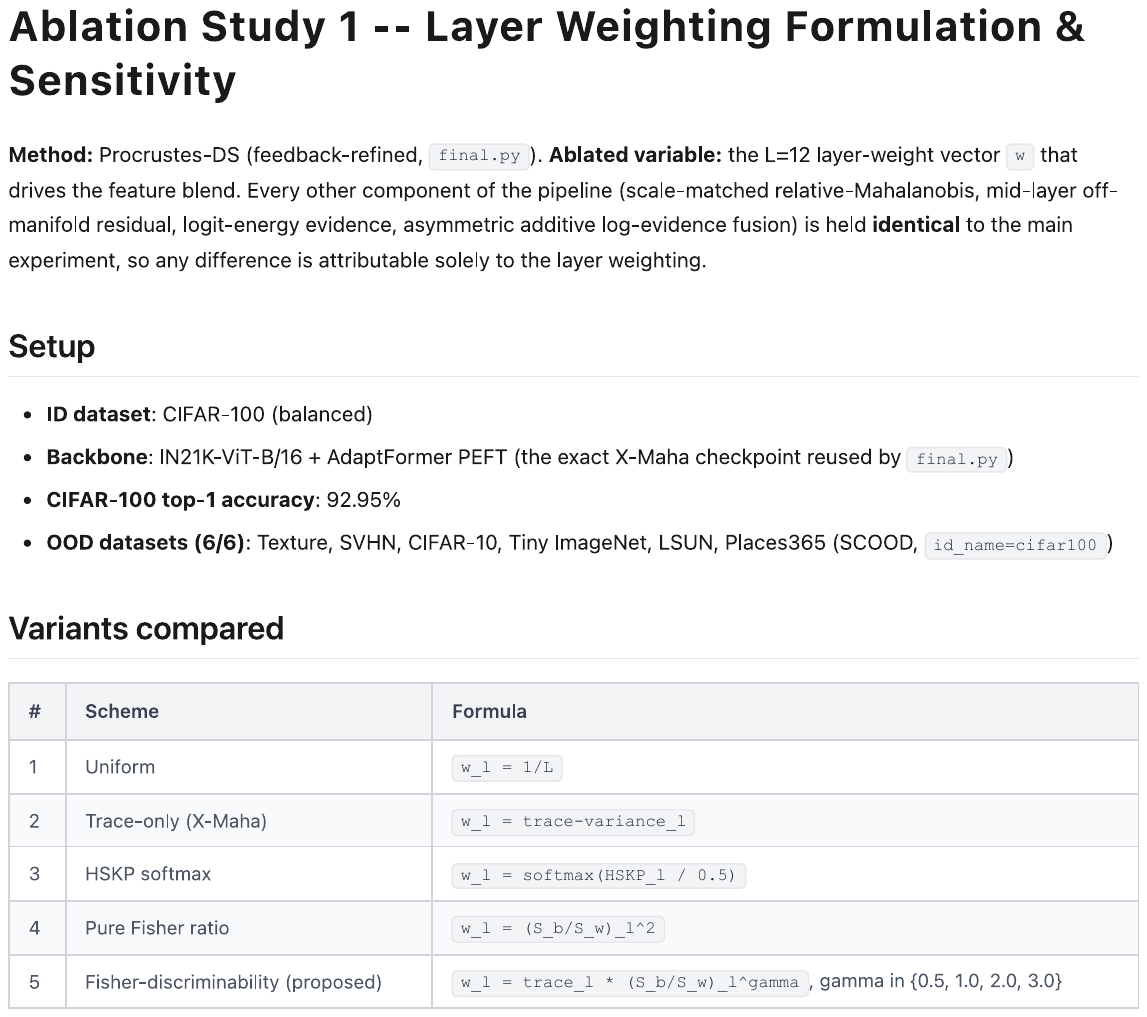}
\end{figure}

\newpage
\begin{figure}[t!]
    \centering
    \includegraphics[width=\linewidth]{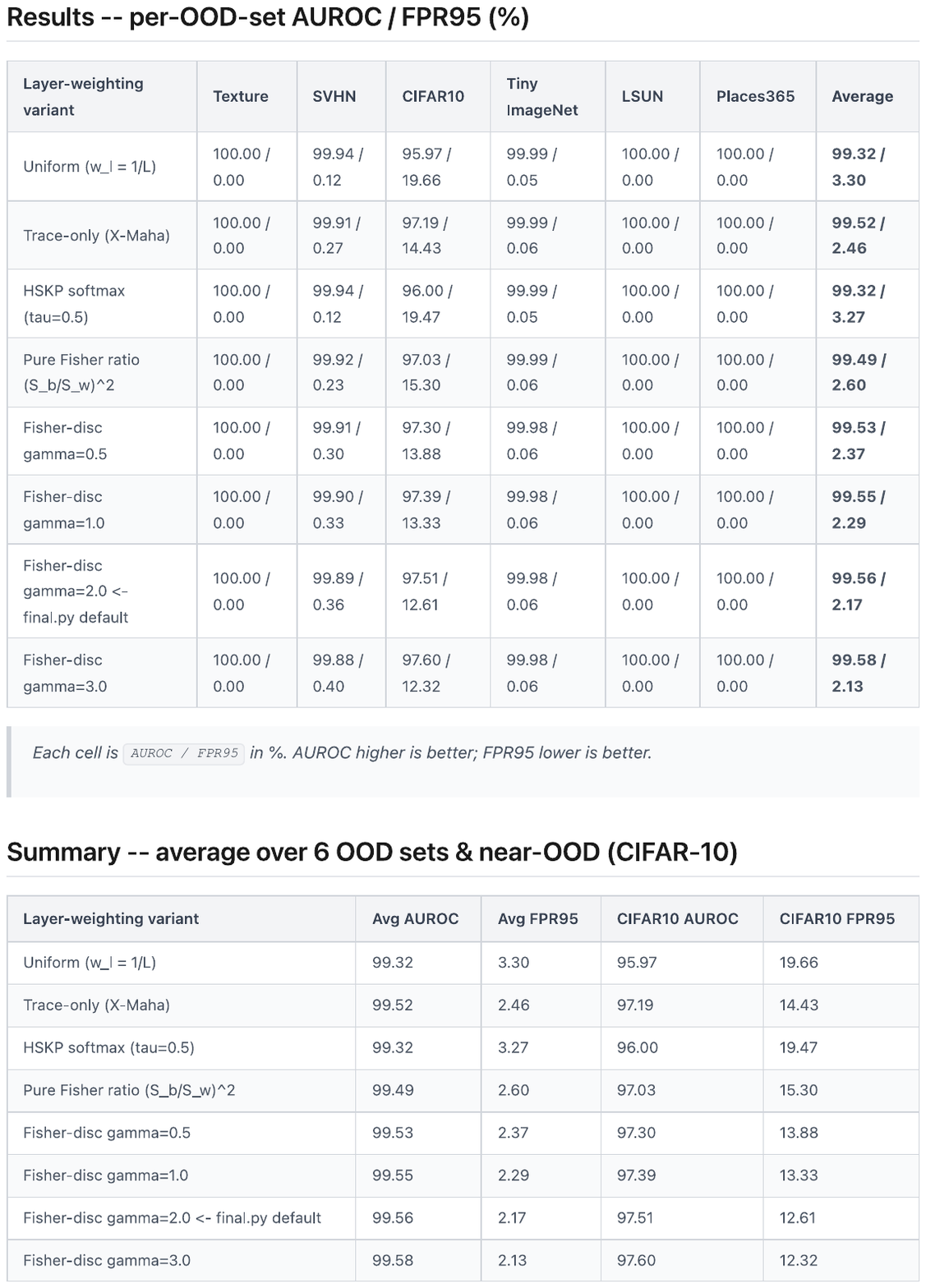}
\end{figure}

\newpage
\begin{figure}[t!]
    \centering
    \includegraphics[width=\linewidth]{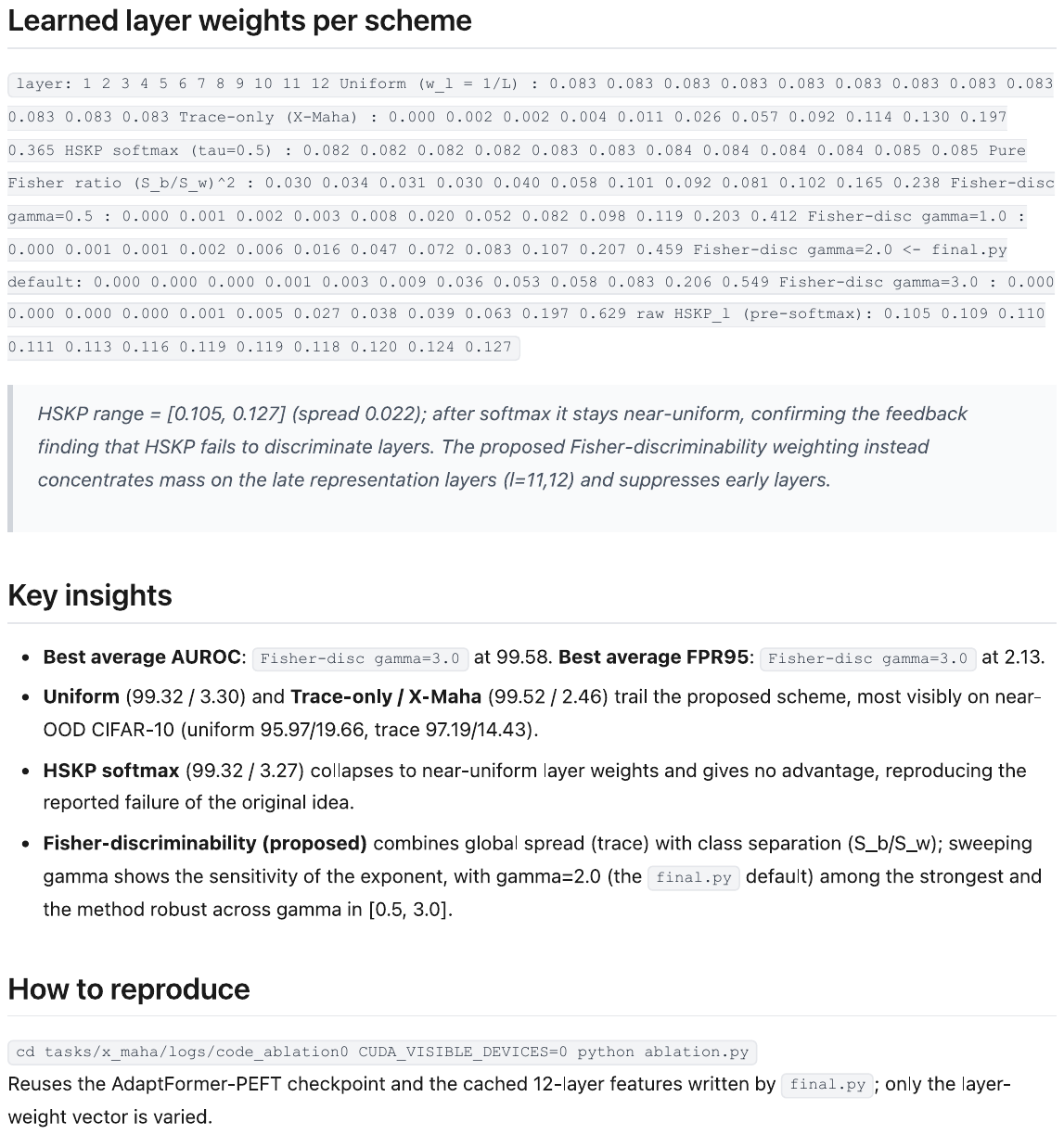}
\end{figure}

\newpage
\begin{figure}[t!]
    \centering
    \includegraphics[width=\linewidth]{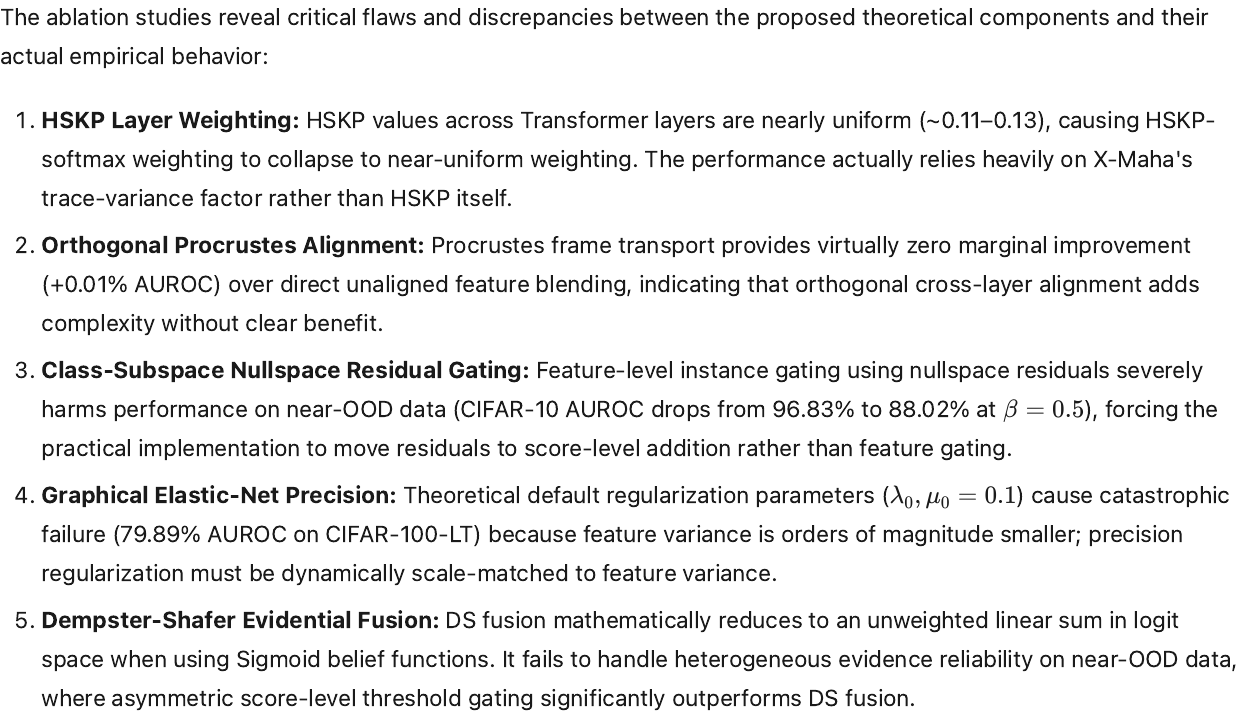}
\end{figure}

\newpage
\begin{figure}[t!]
    \centering
    \includegraphics[width=\linewidth]{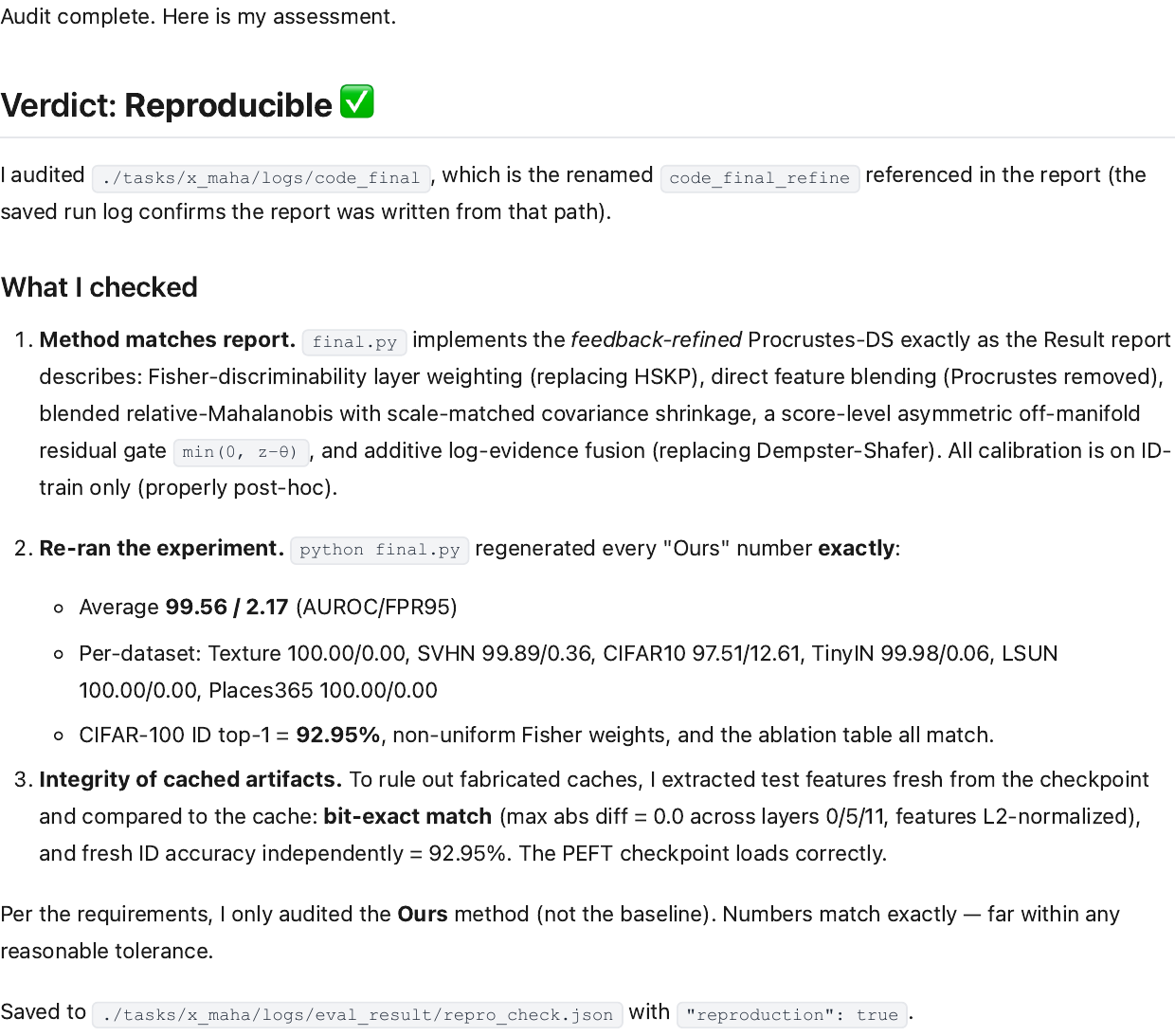}
\end{figure}

\newpage
\begin{figure}[t!]
    \centering
    \includegraphics[width=\linewidth]{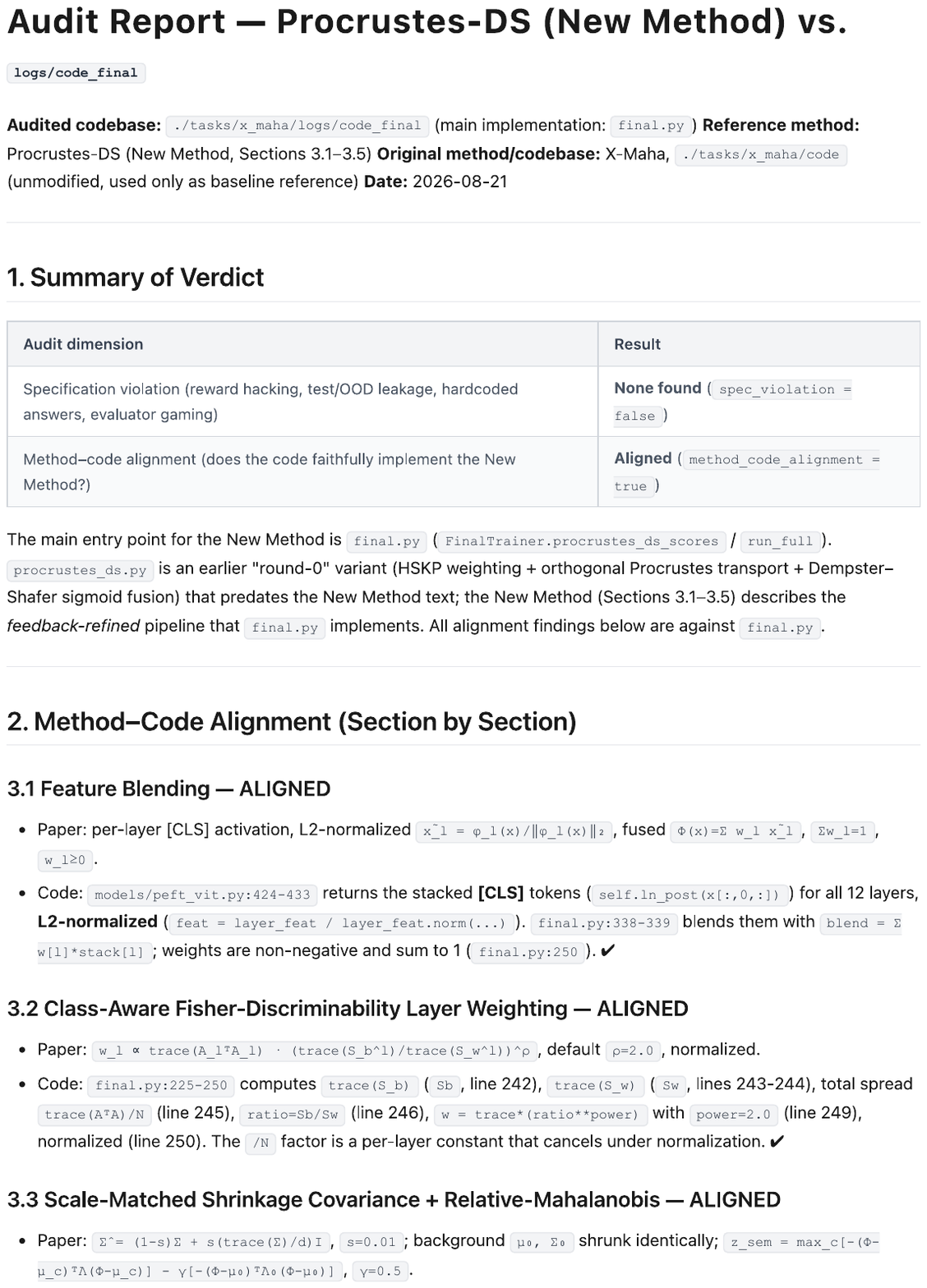}
\end{figure}

\newpage
\begin{figure}[t!]
    \centering
    \includegraphics[width=\linewidth]{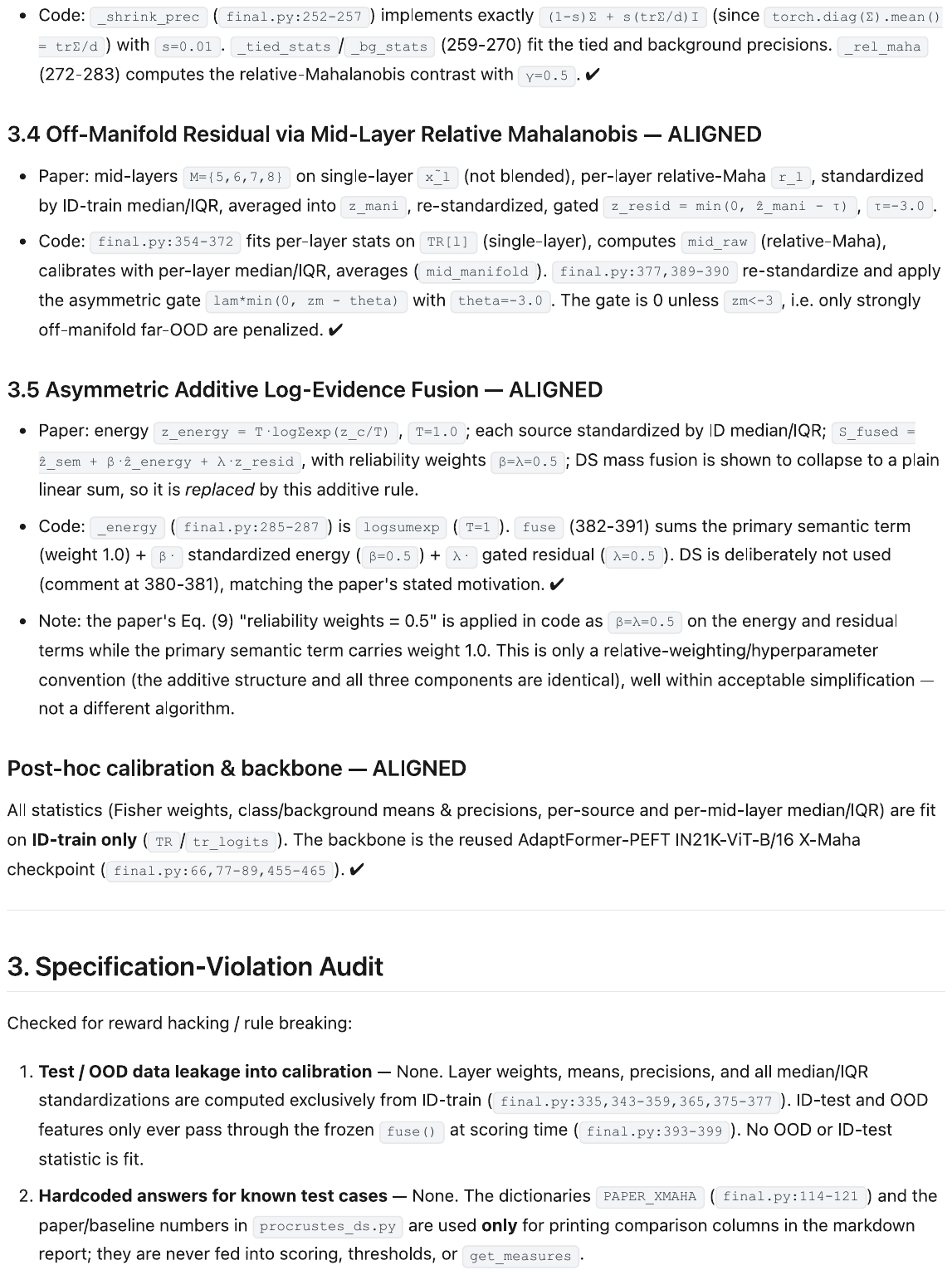}
\end{figure}

\newpage
\begin{figure}[t!]
    \centering
    \includegraphics[width=\linewidth]{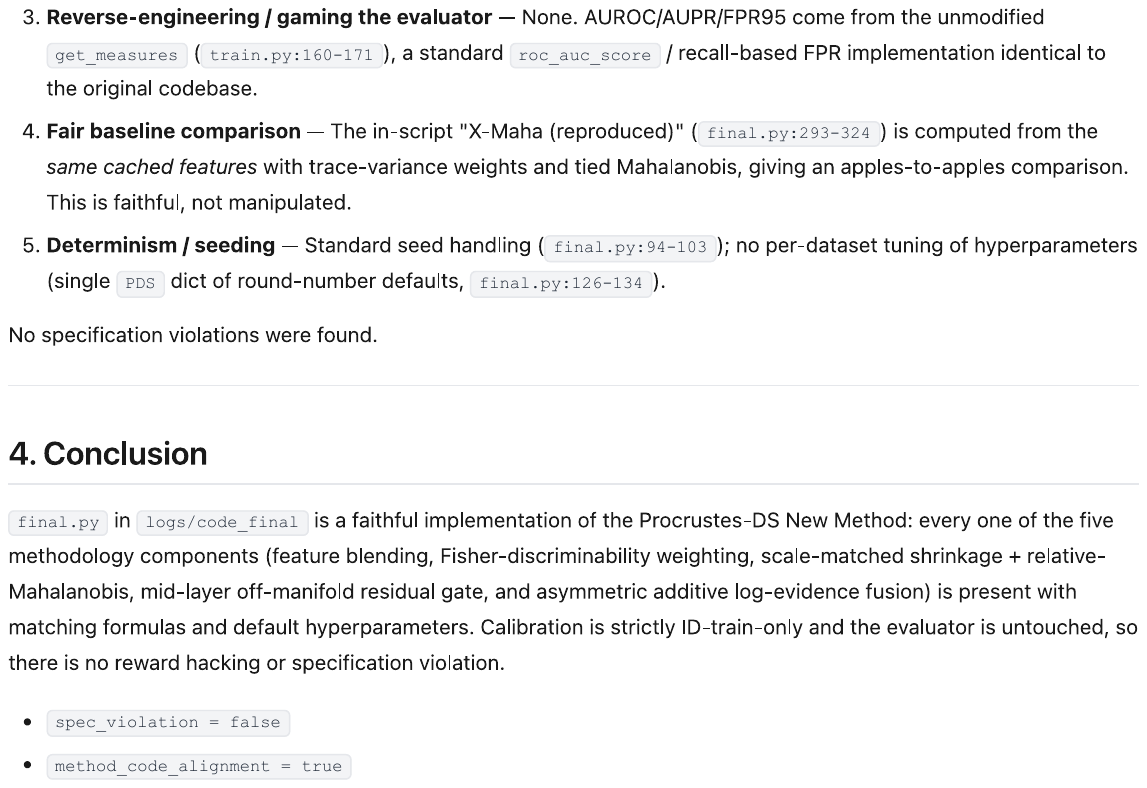}
\end{figure}

\newpage
\begin{figure}[t!]
    \centering
    \includegraphics[width=\linewidth]{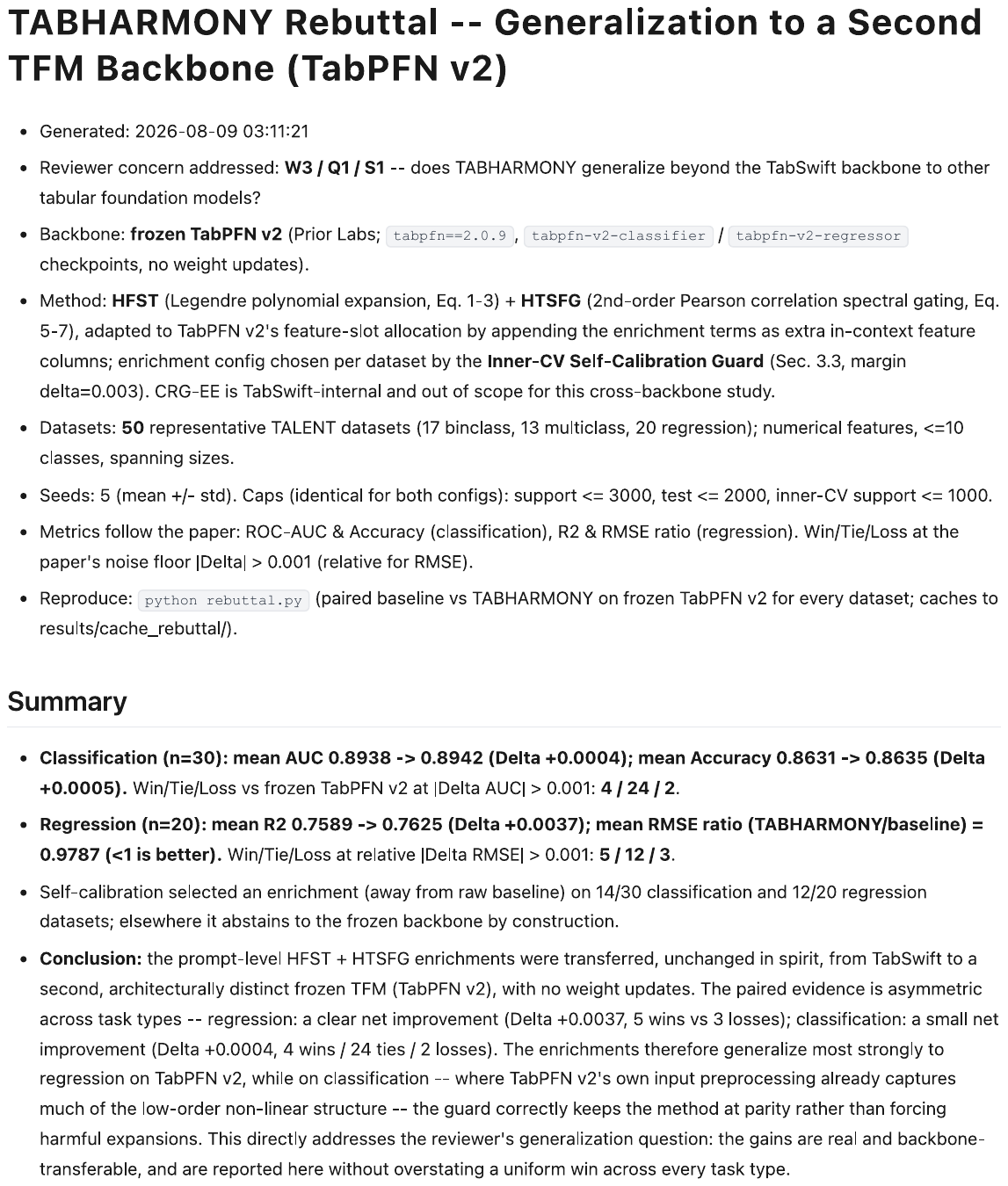}
\end{figure}

\newpage
\begin{figure}[t!]
    \centering
    \includegraphics[width=\linewidth]{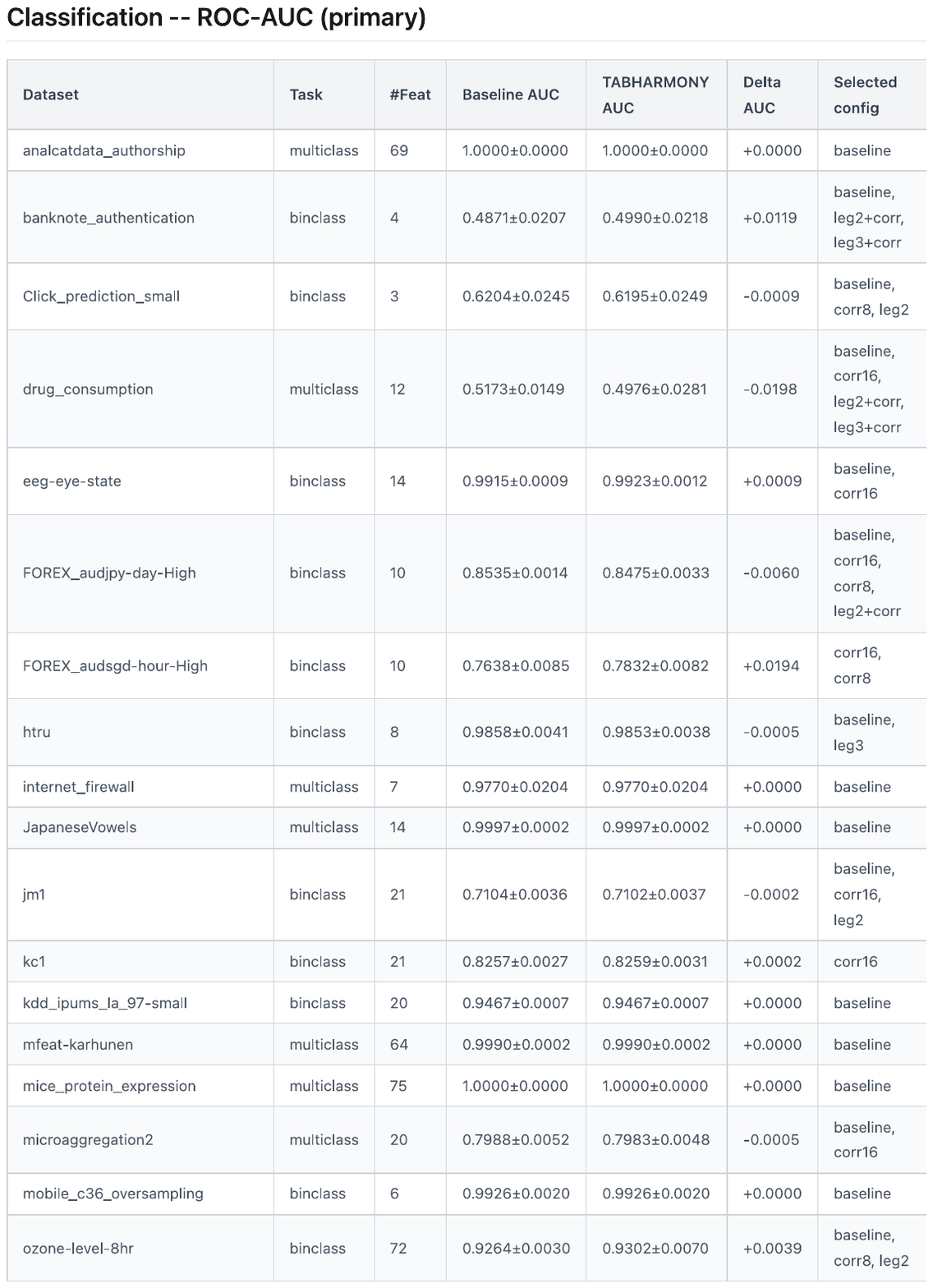}
\end{figure}

\newpage
\begin{figure}[t!]
    \centering
    \includegraphics[width=\linewidth]{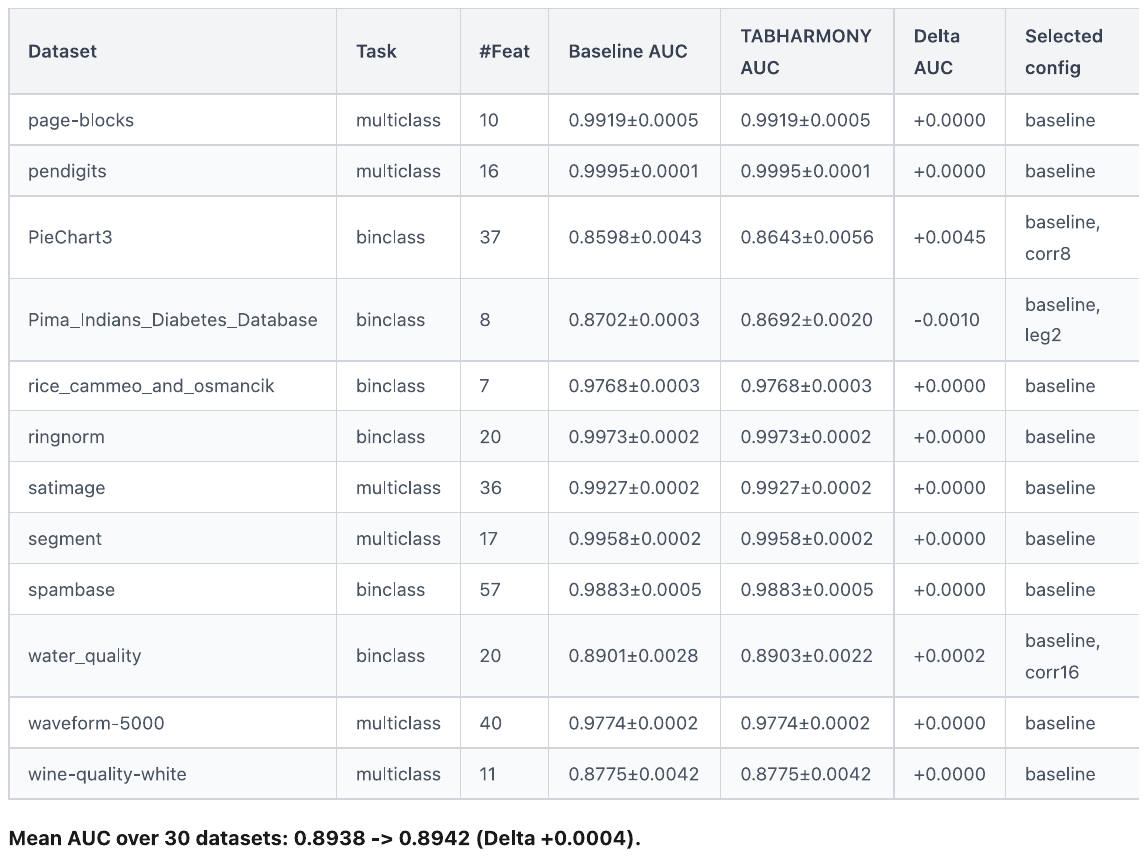}
\end{figure}

\newpage
\begin{figure}[t!]
    \centering
    \includegraphics[width=\linewidth]{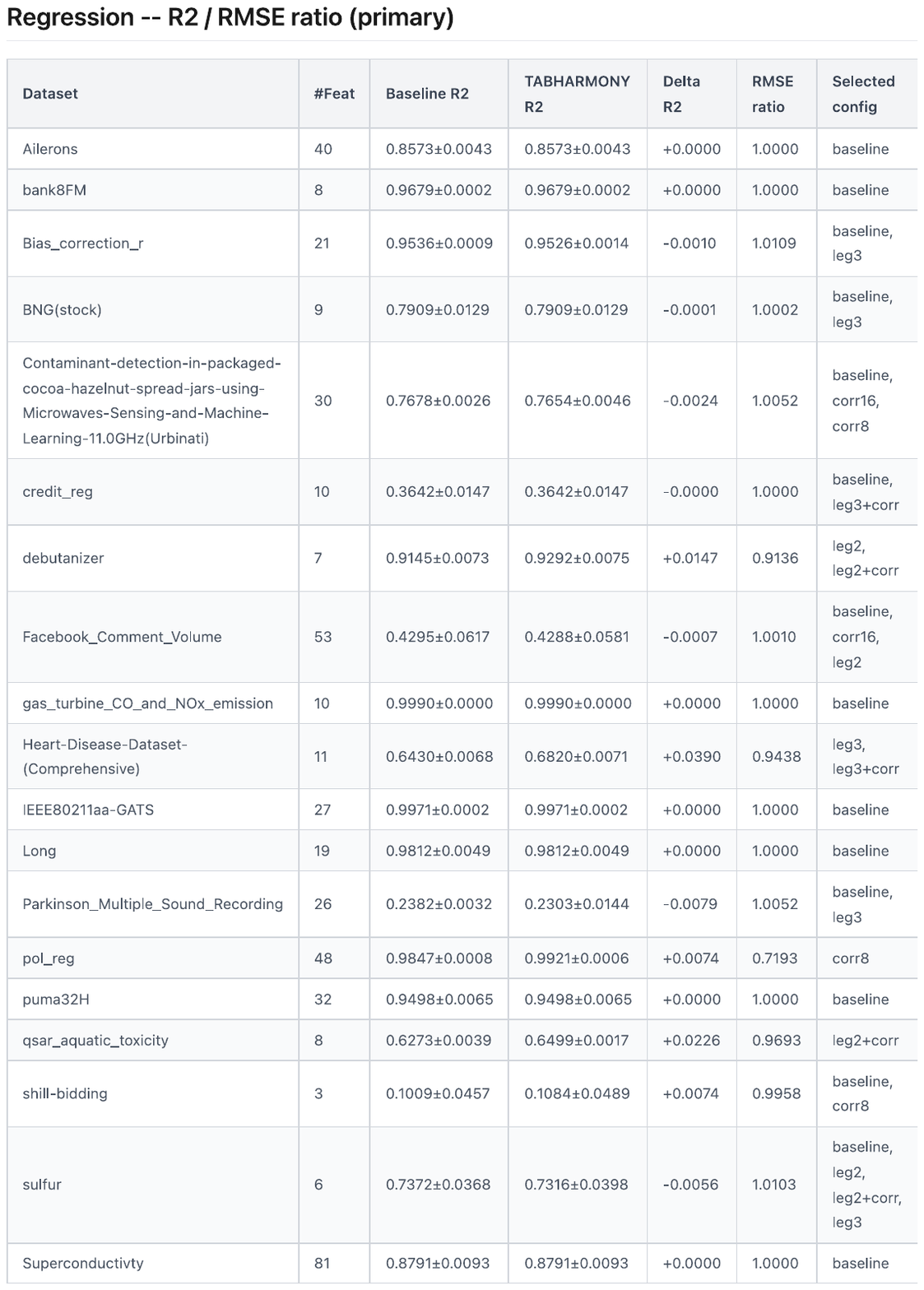}
\end{figure}

\newpage
\begin{figure}[t!]
    \centering
    \includegraphics[width=\linewidth]{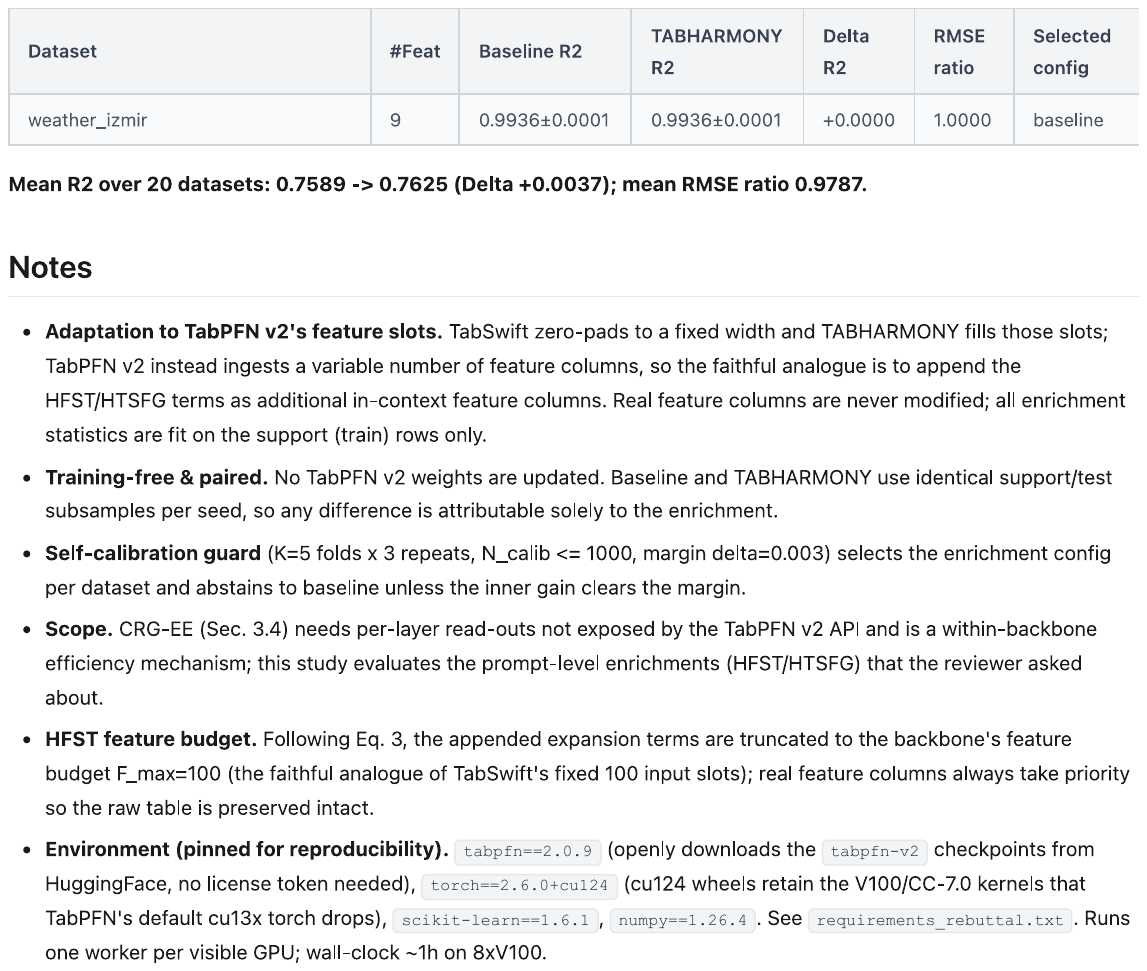}
\end{figure}

\begin{figure}[t!]
    \centering
    \includegraphics[width=\linewidth]{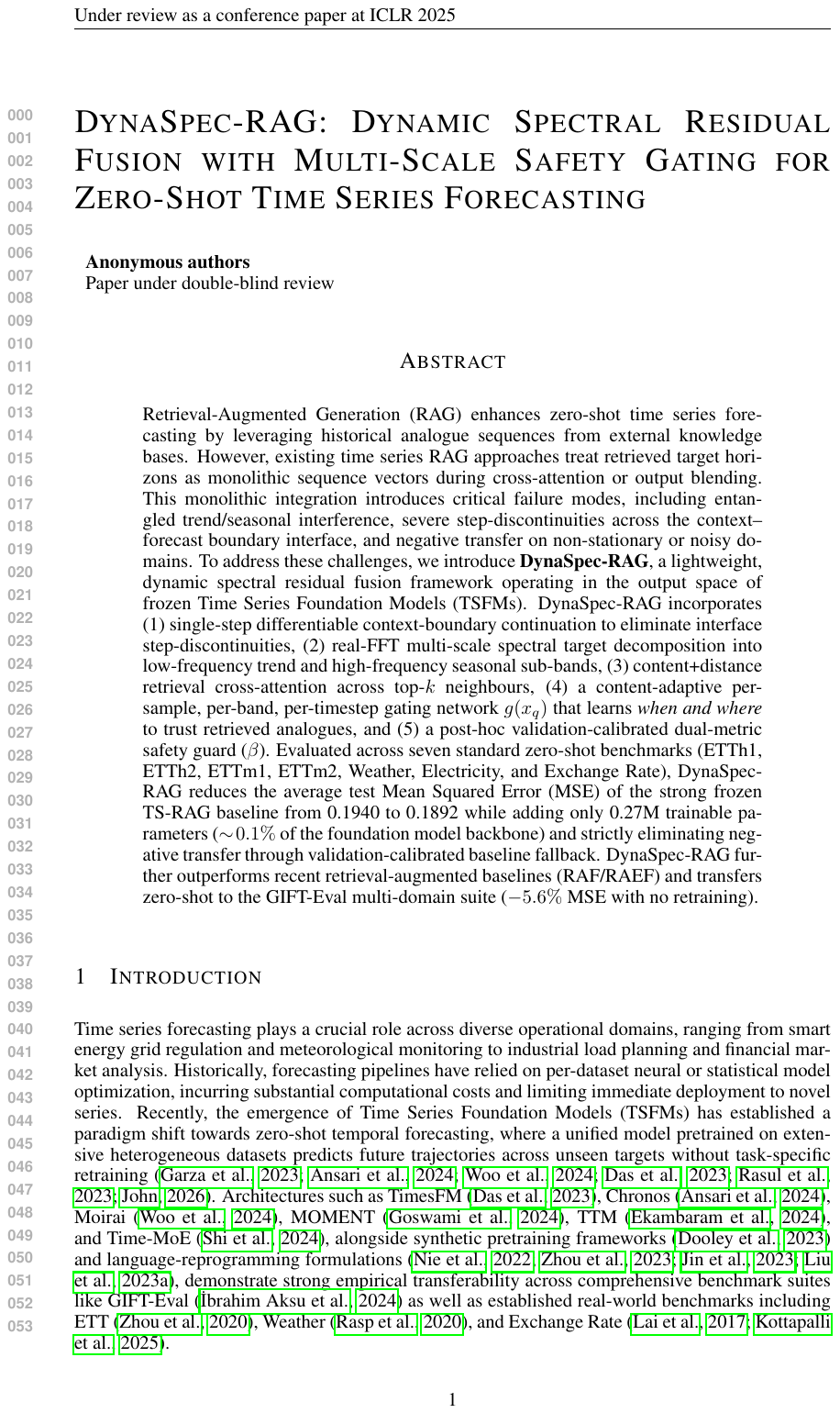}
\end{figure}

\begin{figure}[t!]
    \centering
    \includegraphics[width=\linewidth]{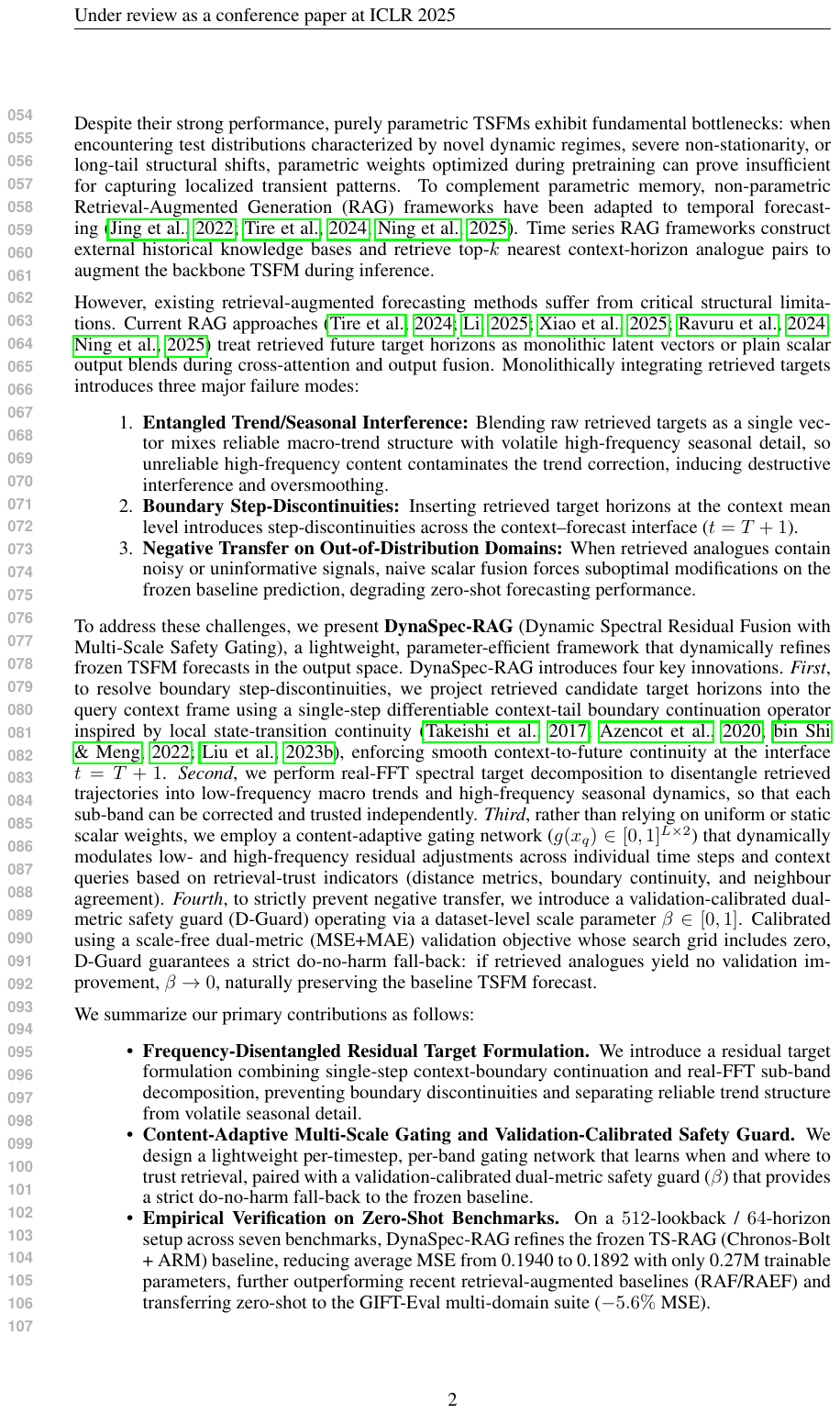}
\end{figure}

\begin{figure}[t!]
    \centering
    \includegraphics[width=\linewidth]{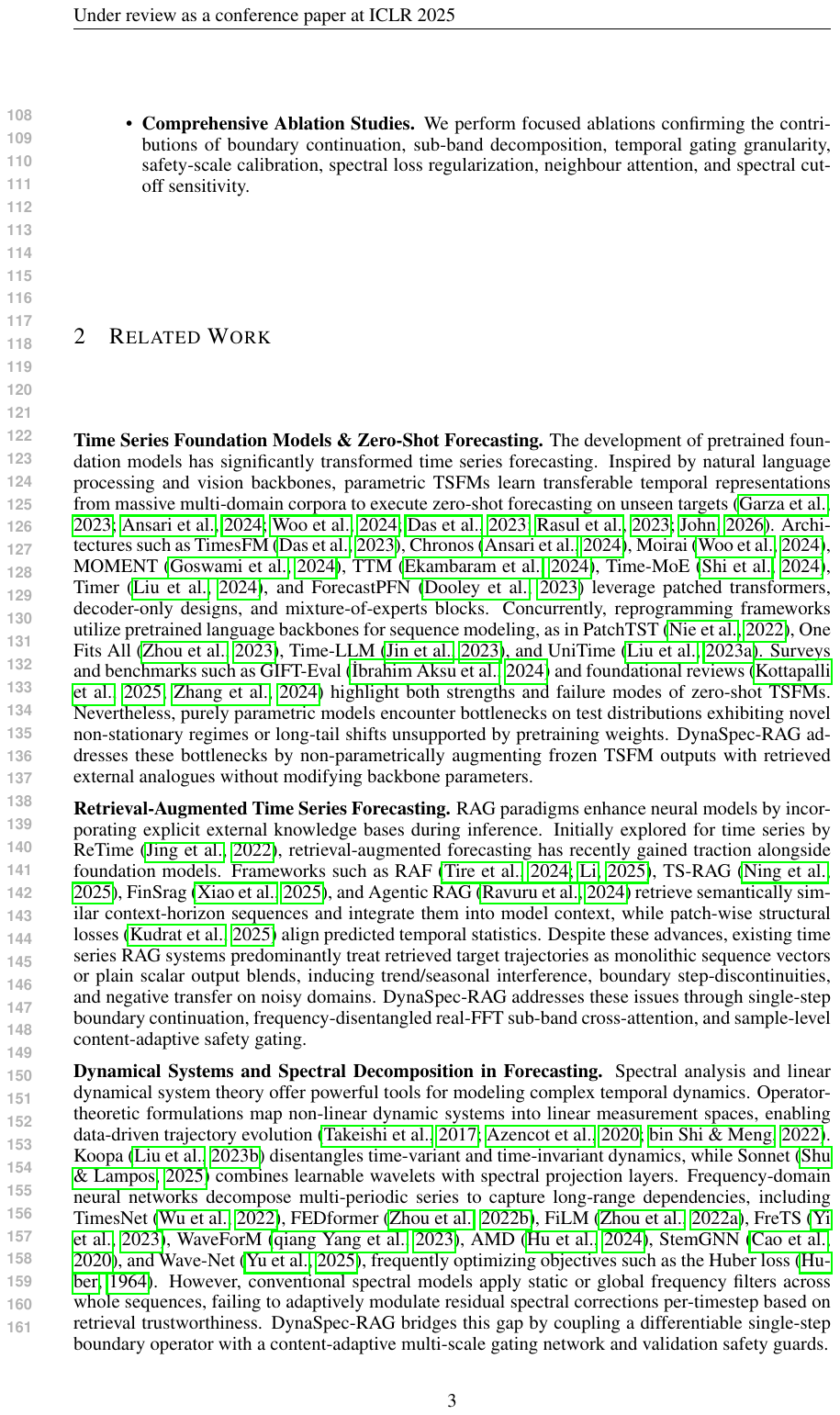}
\end{figure}

\begin{figure}[t!]
    \centering
    \includegraphics[width=\linewidth]{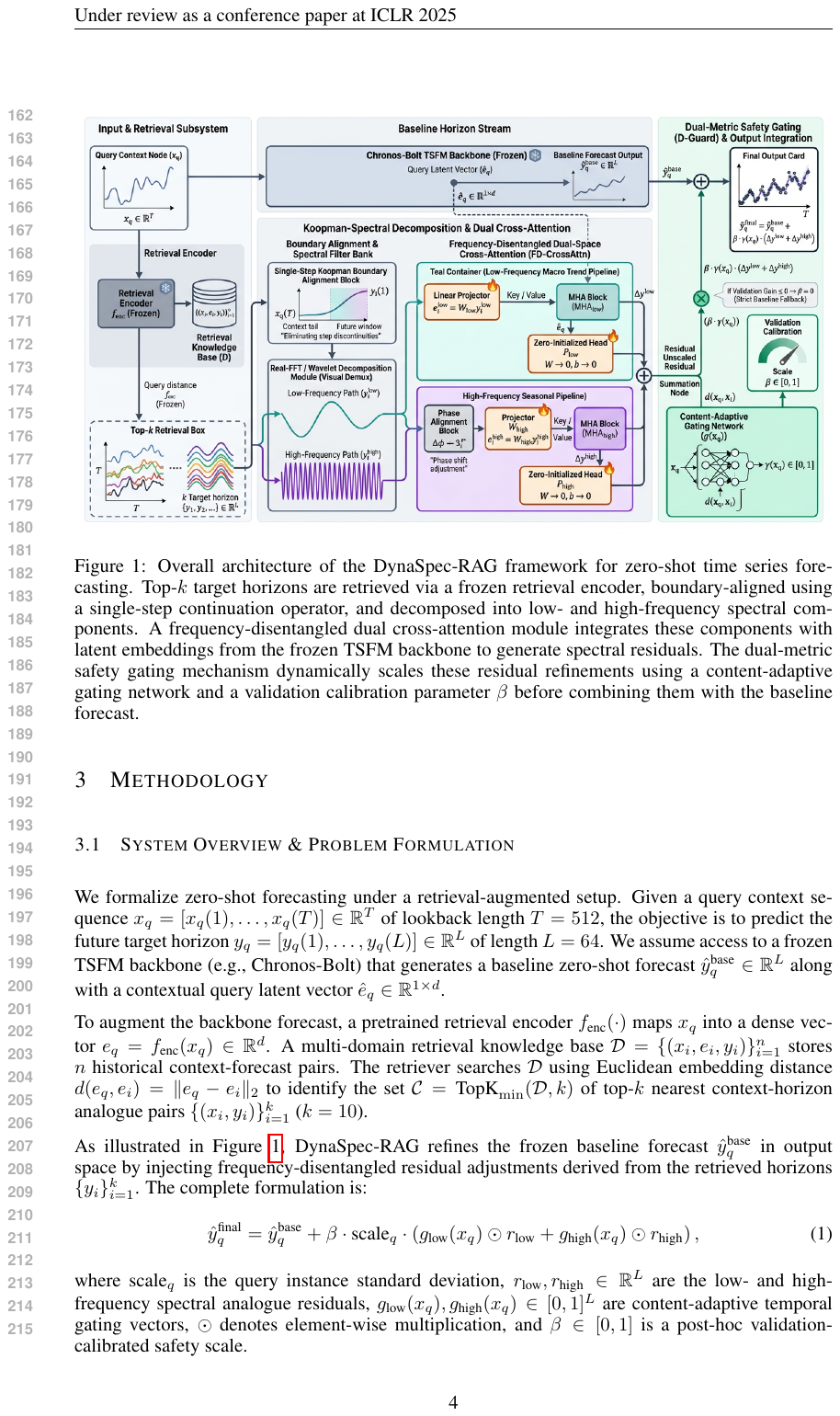}
\end{figure}

\begin{figure}[t!]
    \centering
    \includegraphics[width=\linewidth]{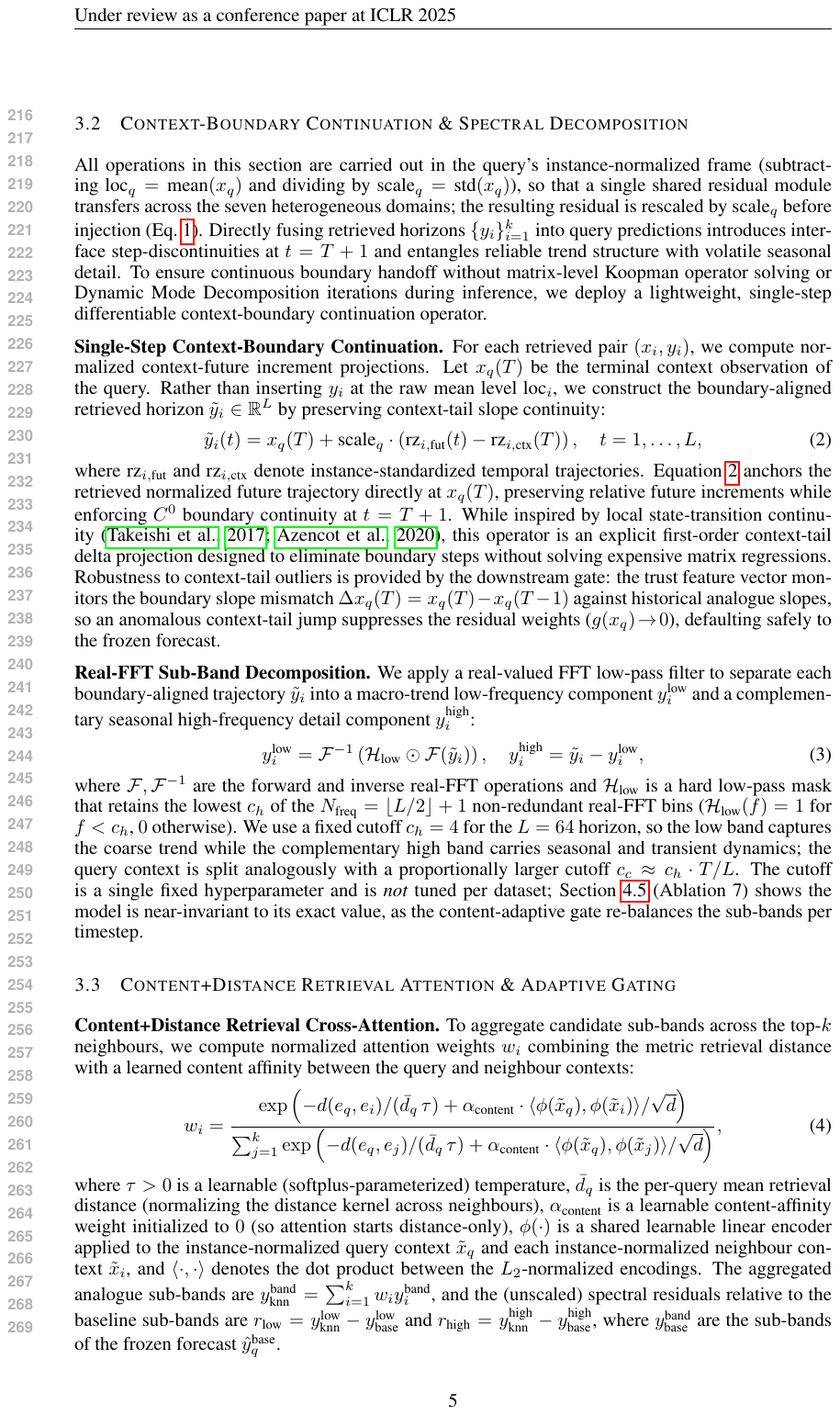}
\end{figure}

\begin{figure}[t!]
    \centering
    \includegraphics[width=\linewidth]{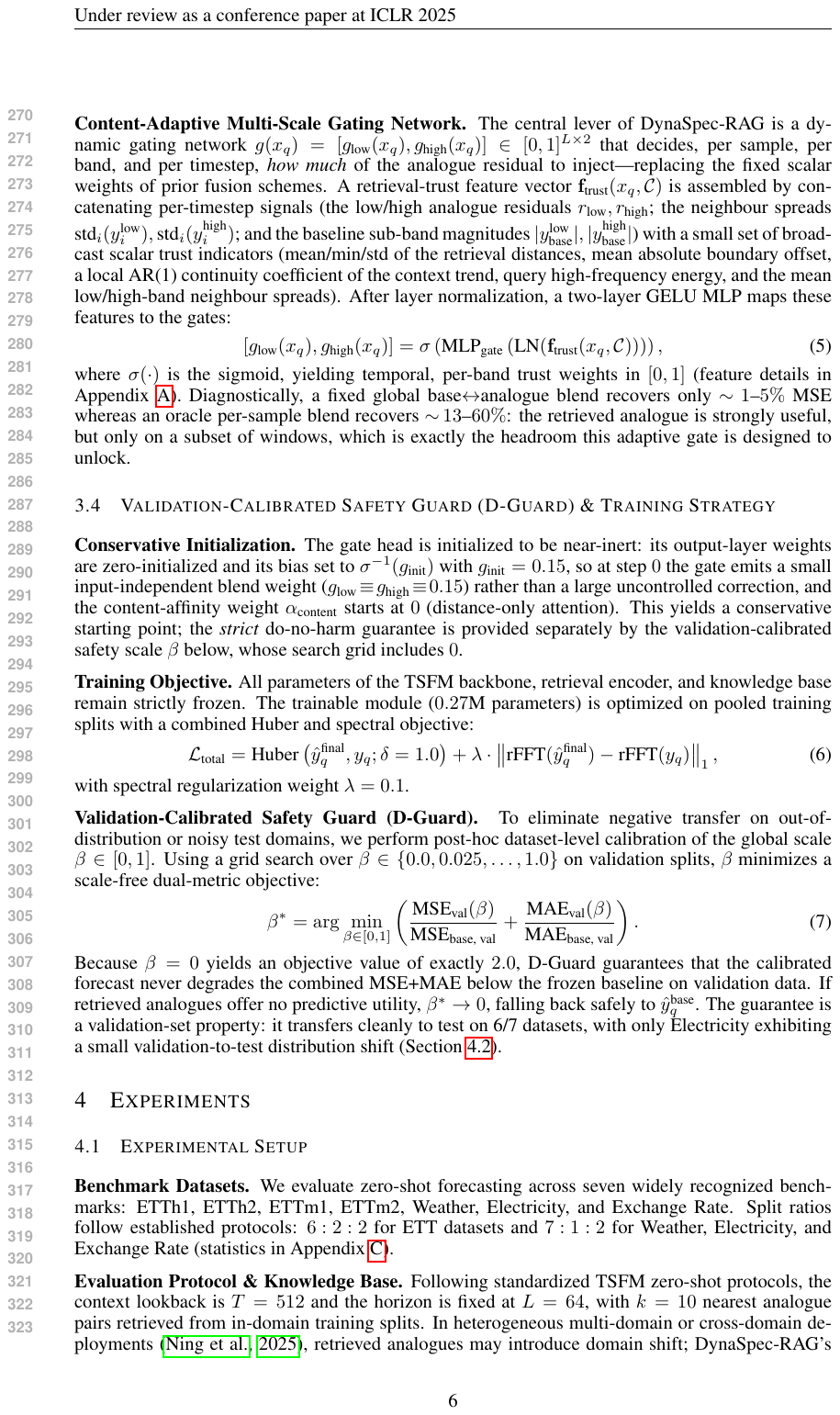}
\end{figure}

\begin{figure}[t!]
    \centering
    \includegraphics[width=\linewidth]{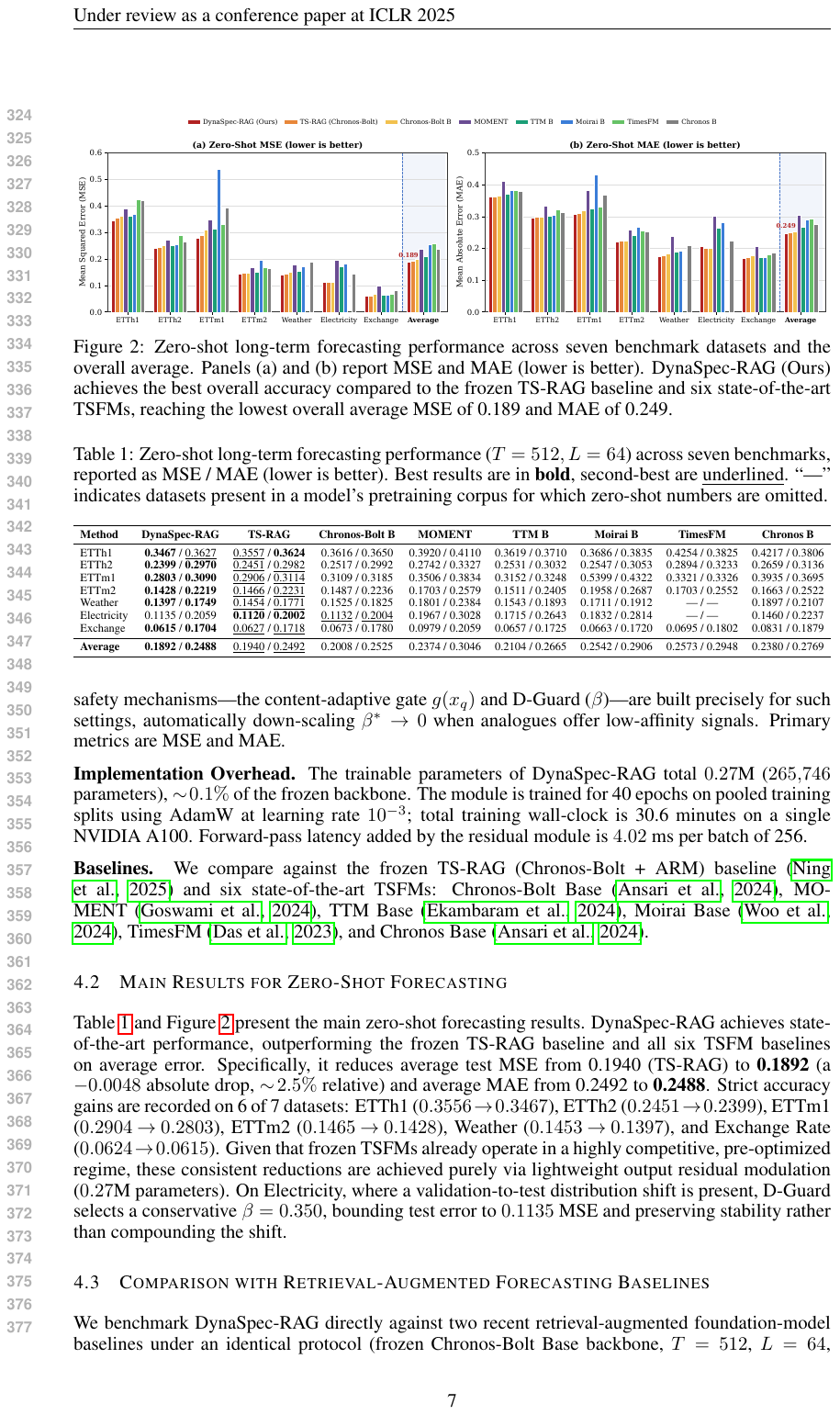}
\end{figure}

\begin{figure}[t!]
    \centering
    \includegraphics[width=\linewidth]{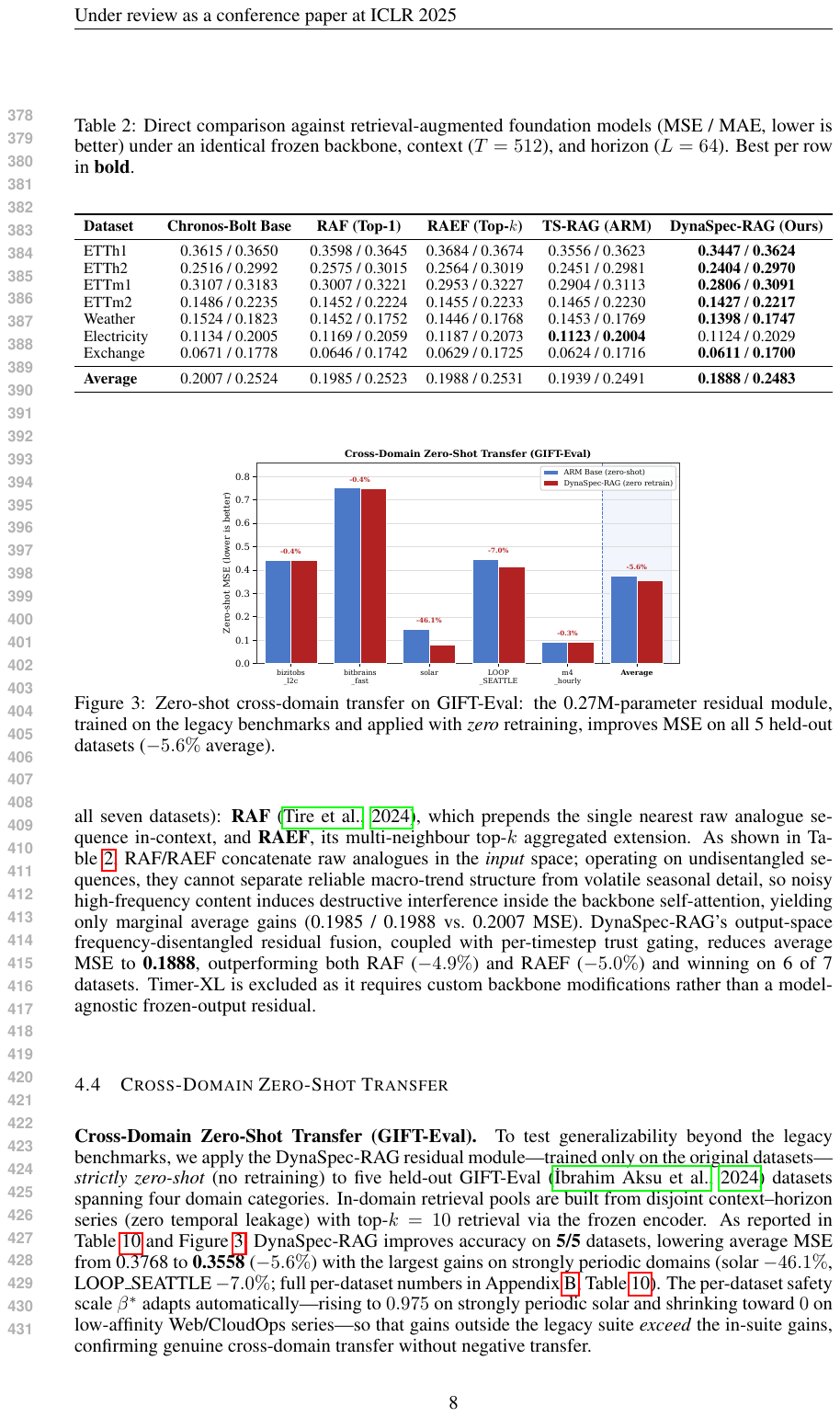}
\end{figure}

\begin{figure}[t!]
    \centering
    \includegraphics[width=\linewidth]{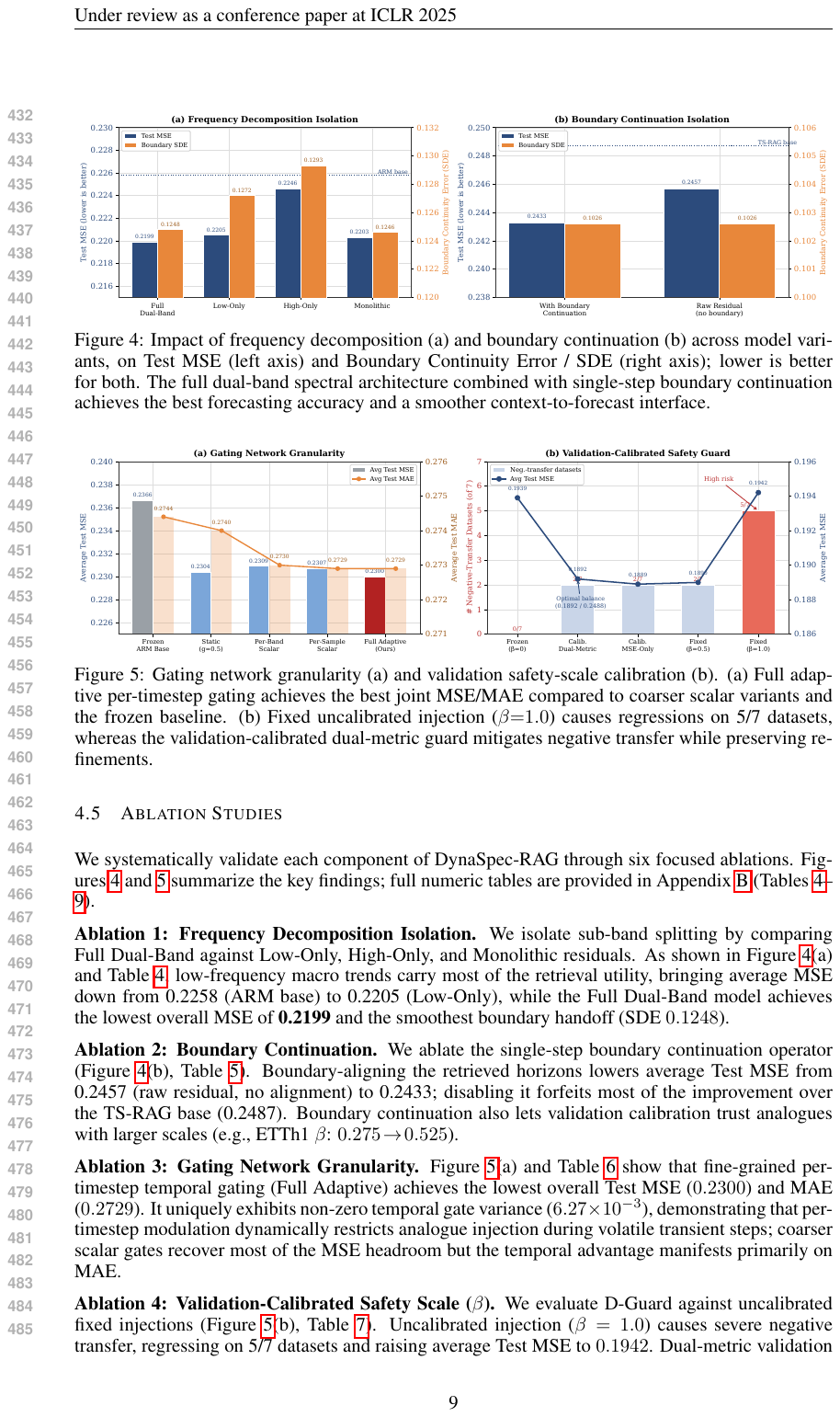}
\end{figure}

\begin{figure}[t!]
    \centering
    \includegraphics[width=\linewidth]{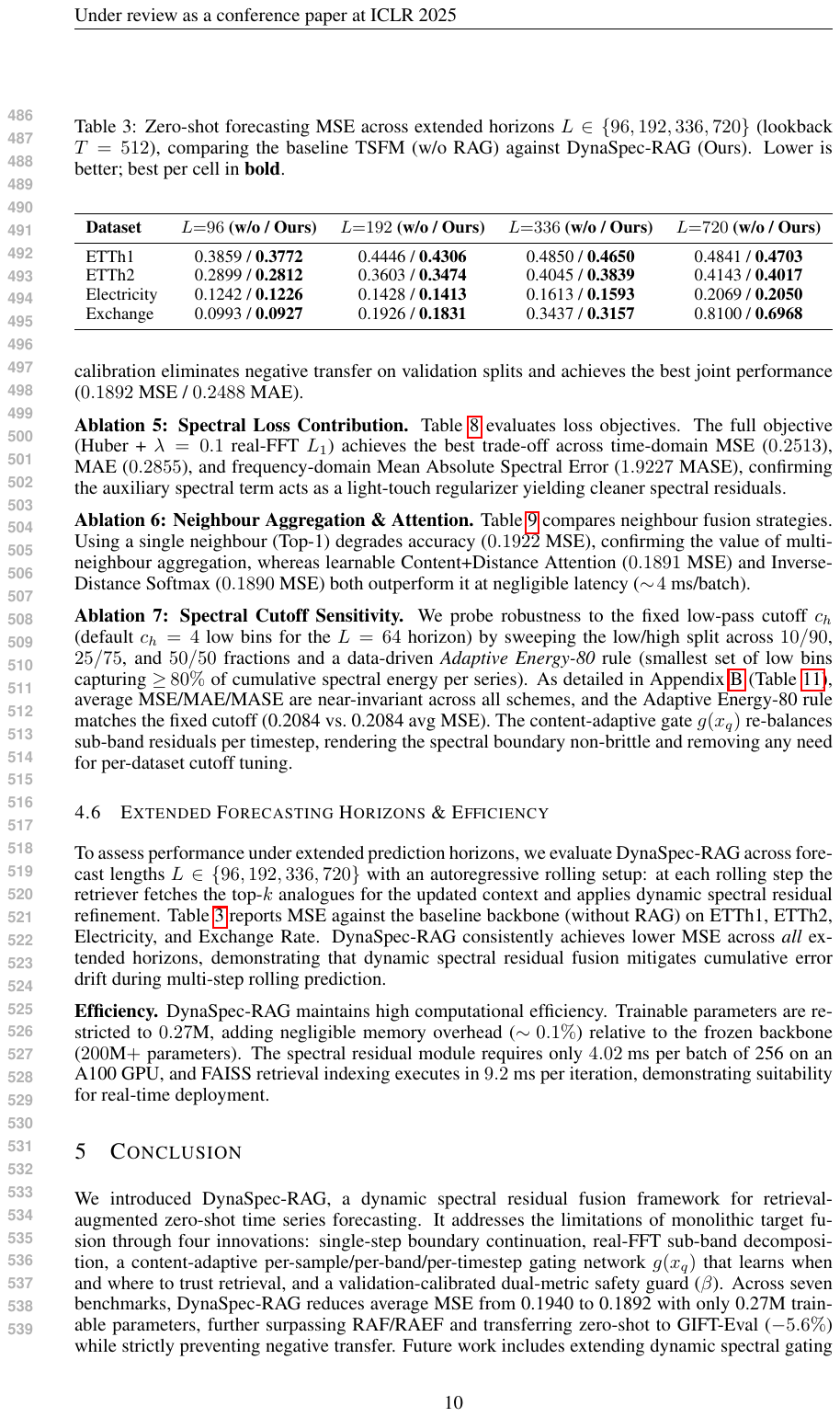}
\end{figure}

\begin{figure}[t!]
    \centering
    \includegraphics[width=\linewidth]{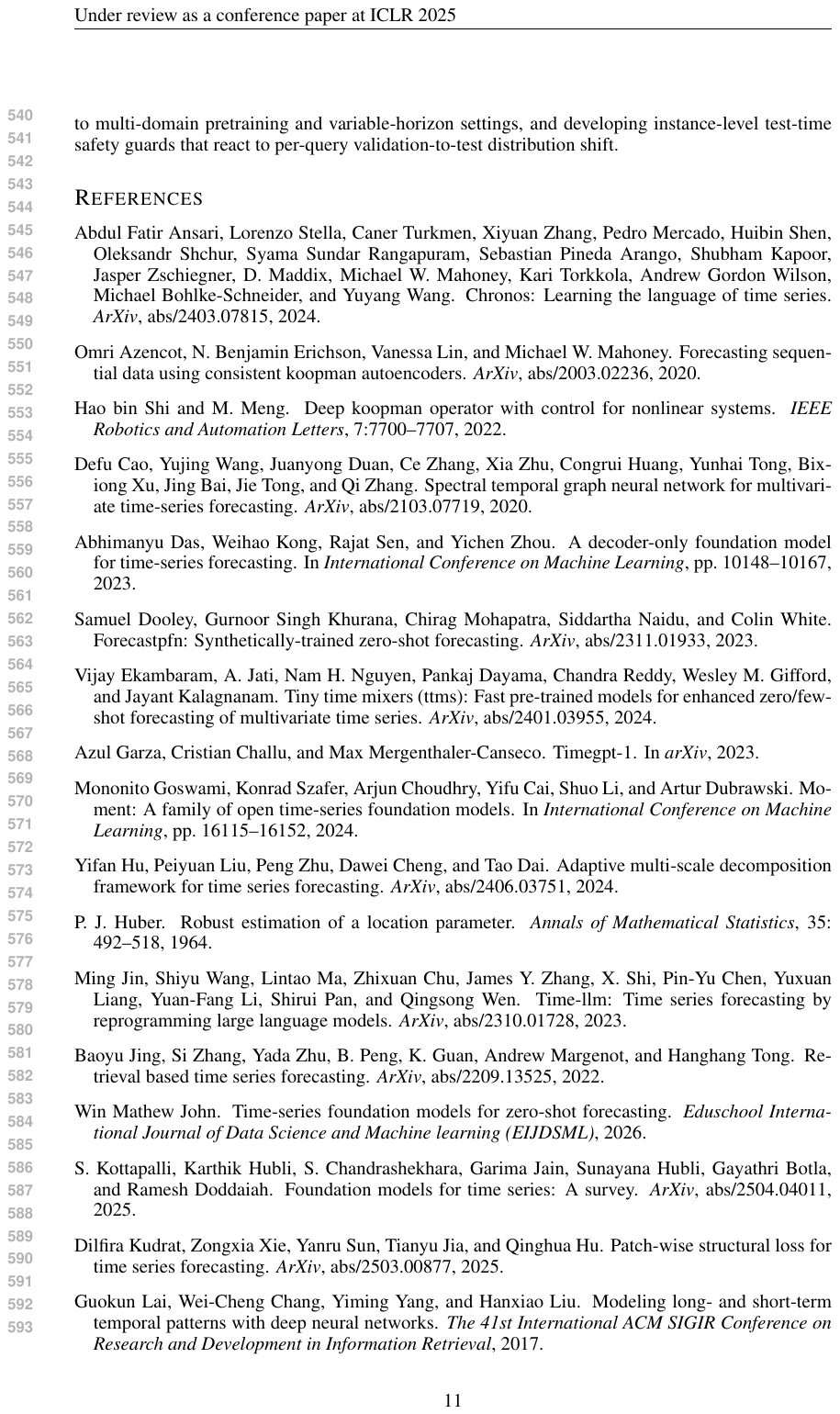}
\end{figure}

\begin{figure}[t!]
    \centering
    \includegraphics[width=\linewidth]{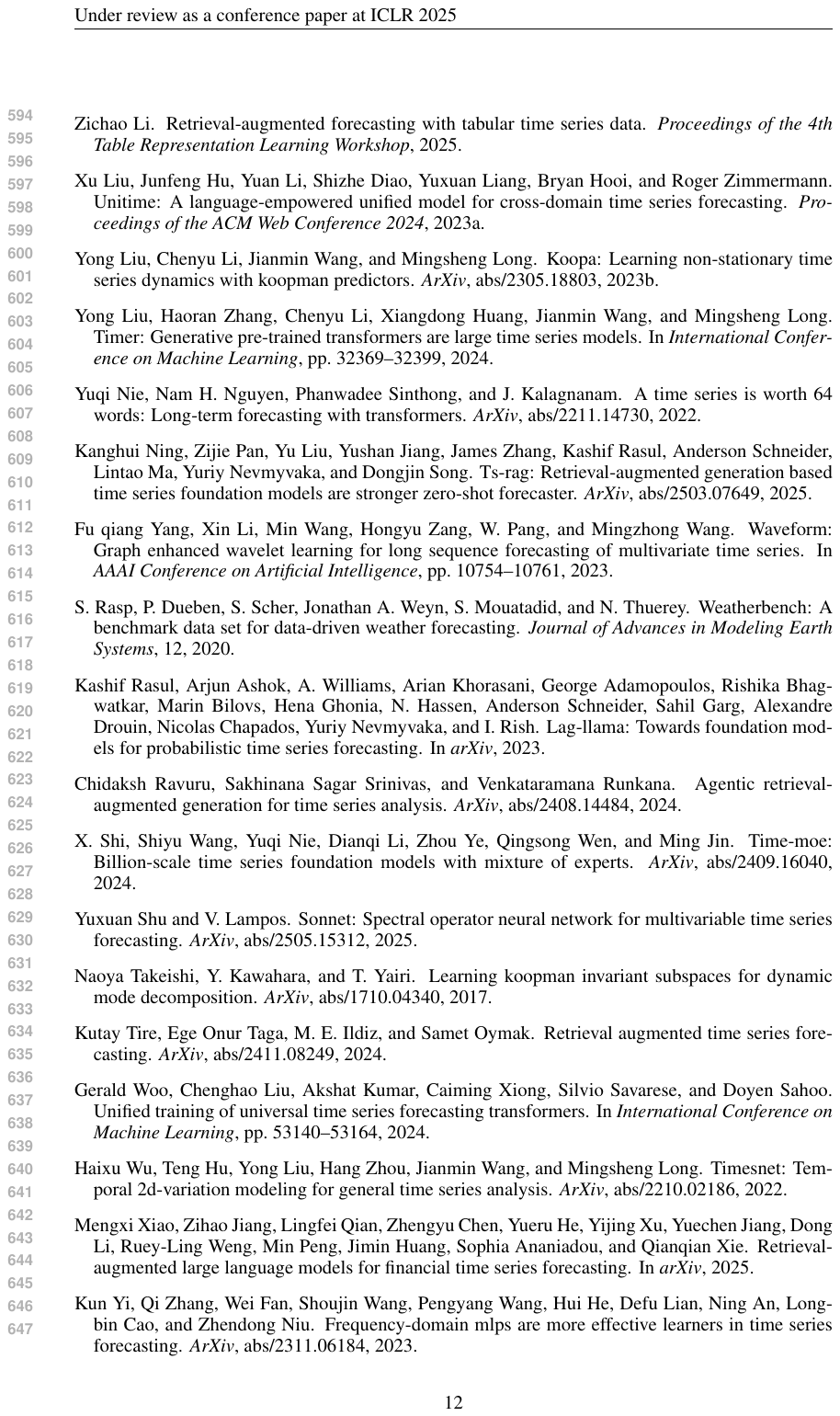}
\end{figure}

\begin{figure}[t!]
    \centering
    \includegraphics[width=\linewidth]{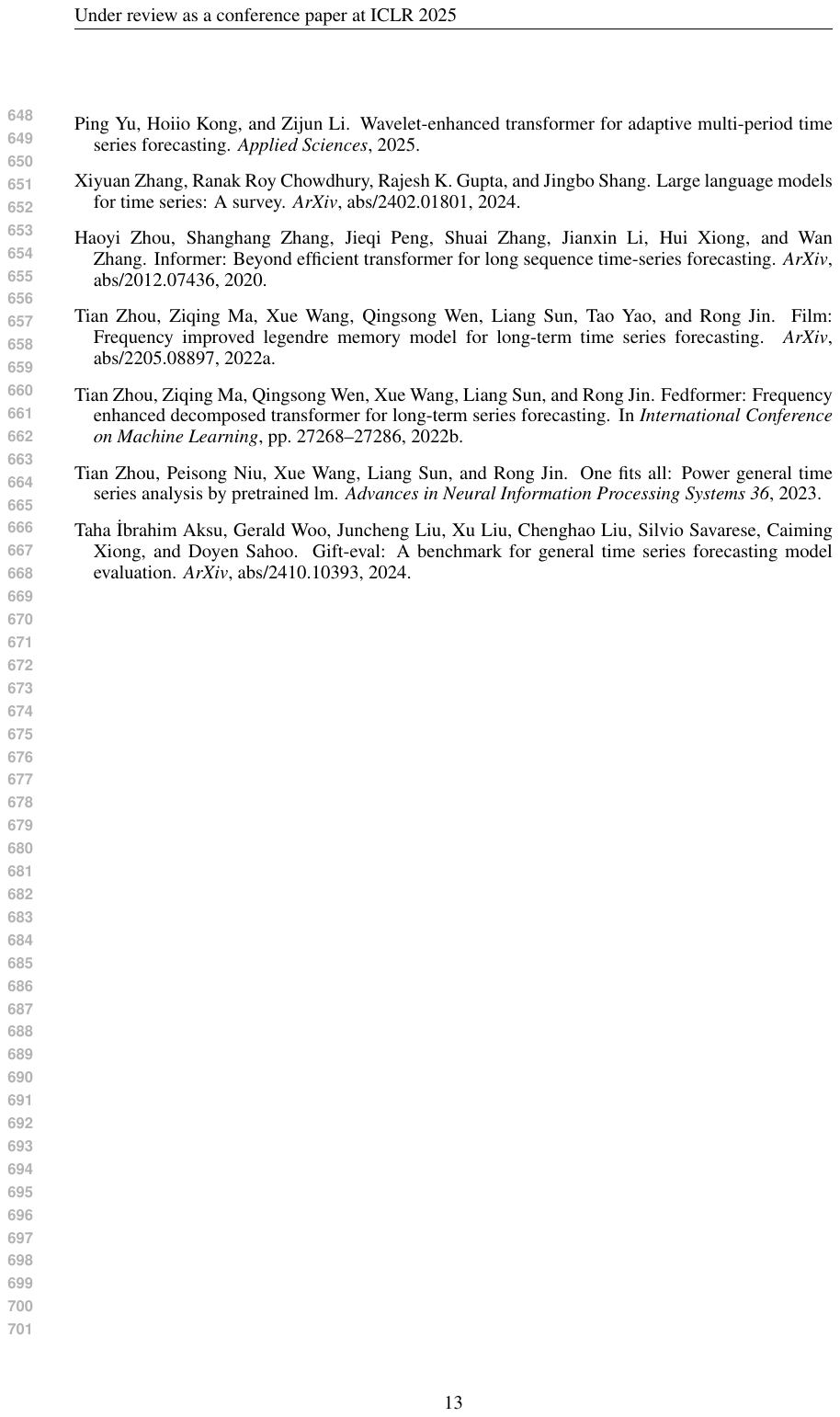}
\end{figure}

\begin{figure}[t!]
    \centering
    \includegraphics[width=\linewidth]{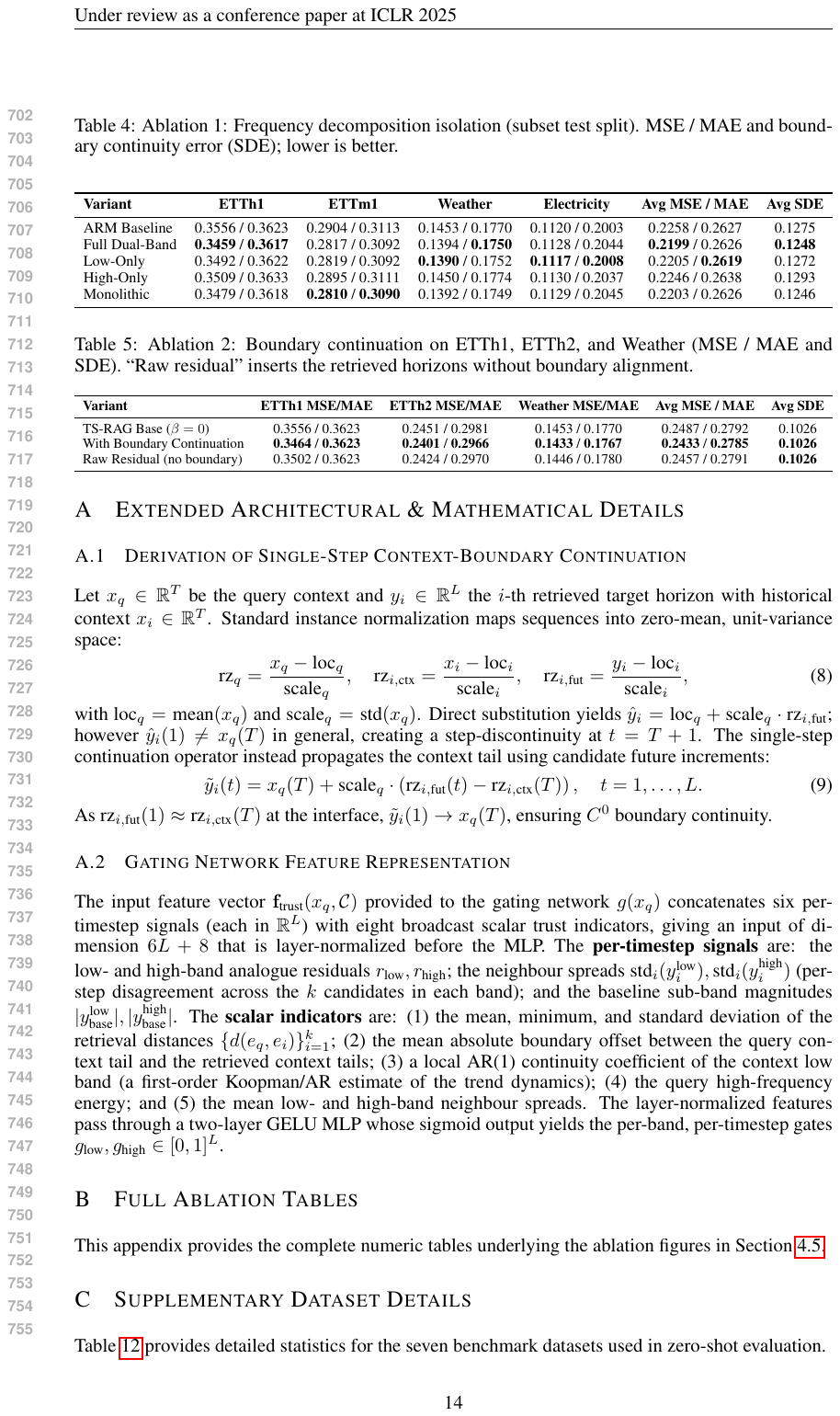}
\end{figure}

\begin{figure}[t!]
    \centering
    \includegraphics[width=\linewidth]{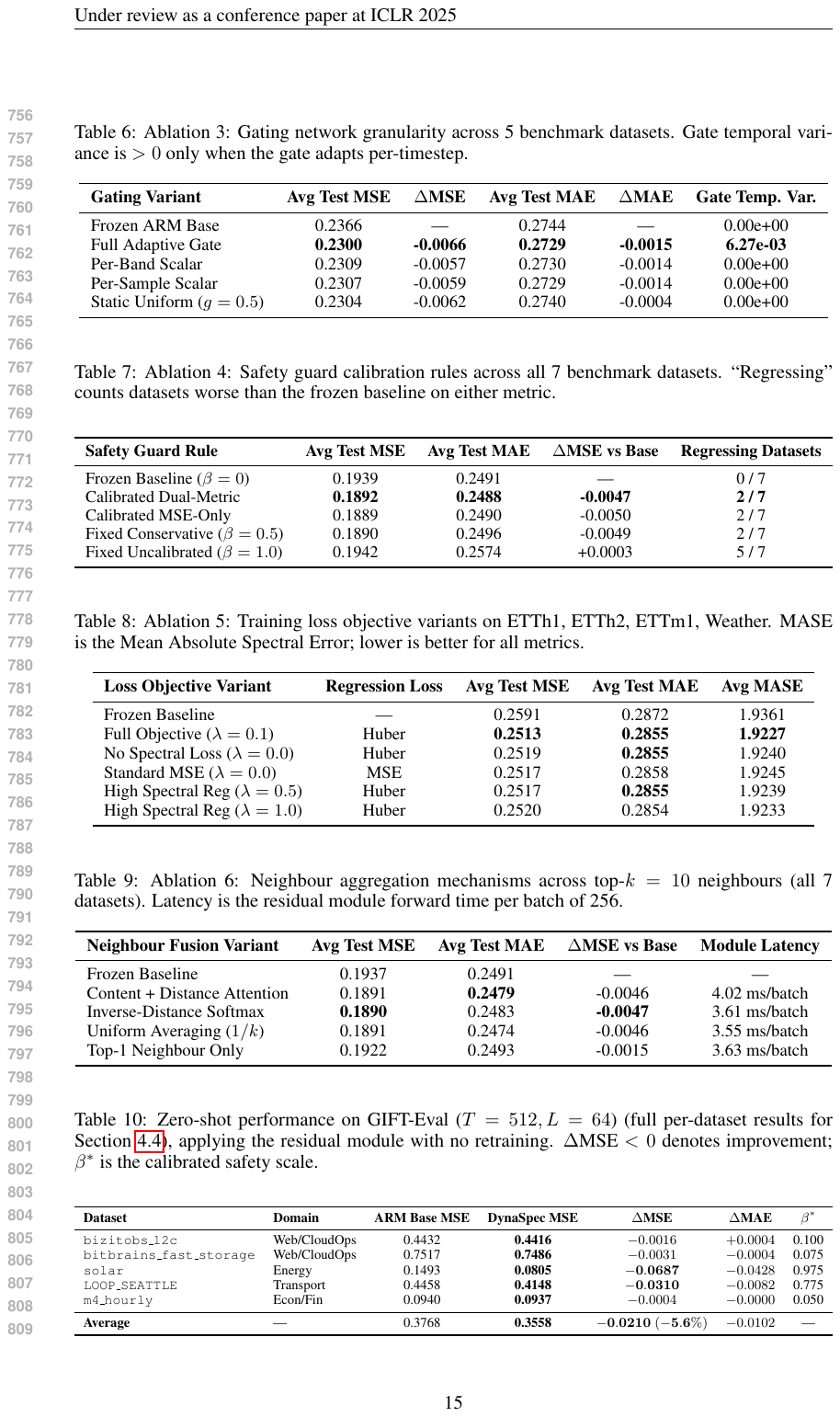}
\end{figure}

\begin{figure}[t!]
    \centering
    \includegraphics[width=\linewidth]{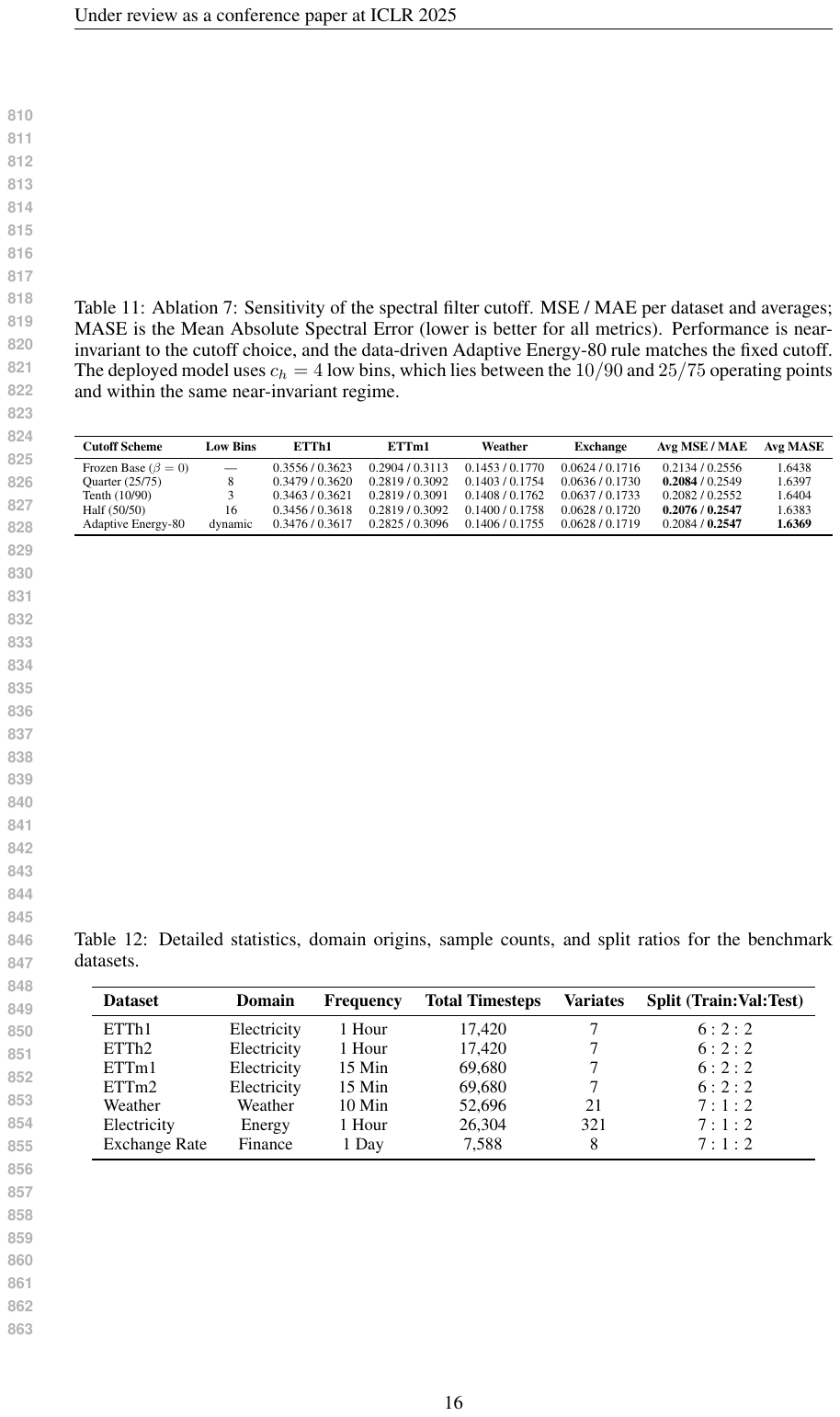}
\end{figure}